\documentclass[]{fairmeta}

\title{\titletext}

\abstract{\abstracttext}

\author[1,2,3,4]{Dylan R.\ Ashley}
\author[1]{Ga{\"{e}}l Le Lan}
\author[1]{Changsheng Zhao}
\author[1]{Naina Dhingra}
\author[1]{Zhipeng Cai}
\author[1]{Ernie Chang}
\author[5]{Mingchen Zhuge}
\author[1]{Yangyang Shi}
\author[1]{Vikas Chandra}
\author[2,3,4,5]{J{\"{u}}rgen Schmidhuber}

\affiliation[1]{Meta Platforms, Inc., Menlo Park, United States of America}
\affiliation[2]{The Swiss AI Lab IDSIA (USI-SUPSI), Lugano, Switzerland}
\affiliation[3]{Universit{\`{a}} della Svizzera italiana, Lugano, Switzerland}
\affiliation[4]{Scuola universitaria professionale della Svizzera italiana, Lugano, Switzerland}
\affiliation[5]{Center of Excellence for Generative AI, KAUST, Thuwal, Saudi Arabia}

\correspondence{Dylan R.\ Ashley at \email{dylan.ashley@usi.ch}}

\usepackage{adjustbox}
\usepackage{amsmath}
\usepackage{amssymb}
\usepackage{amsthm}
\usepackage{booktabs}
\usepackage{caption}
\usepackage{fancyvrb}
\usepackage{float}
\usepackage{framed}
\usepackage{fvextra}
\usepackage{graphicx}
\usepackage{inconsolata}
\usepackage{listings}
\usepackage{mathtools}
\usepackage{microtype}
\usepackage{rotating}
\usepackage{subcaption}
\usepackage{titletoc}
\usepackage{wrapfig}
\usepackage{xcolor}
\usepackage{xurl}

\crefname{algorithm}{Algorithm}{Algorithms}
\Crefname{algorithm}{Algorithm}{Algorithms}
\crefname{appendix}{Appendix}{Appendices}
\Crefname{appendix}{Appendix}{Appendices}
\crefname{chapter}{Chapter}{Chapters}
\Crefname{chapter}{Chapter}{Chapters}
\crefname{corollary}{Corollary}{Corollaries}
\Crefname{corollary}{Corollary}{Corollaries}
\crefname{definition}{Definition}{Definitions}
\Crefname{definition}{Definition}{Definitions}
\crefname{equation}{Equation}{Equations}
\Crefname{equation}{Equation}{Equations}
\crefname{example}{Example}{Examples}
\Crefname{example}{Example}{Examples}
\crefname{figure}{Figure}{Figures}
\Crefname{figure}{Figure}{Figures}
\crefname{lemma}{Lemma}{Lemmas}
\Crefname{lemma}{Lemma}{Lemmas}
\crefname{line}{Line}{Lines}
\Crefname{line}{Line}{Lines}
\crefname{listing}{Listing}{Listings}
\Crefname{listing}{Listing}{Listings}
\crefname{page}{Page}{Pages}
\Crefname{page}{Page}{Pages}
\crefname{part}{Part}{Parts}
\Crefname{part}{Part}{Parts}
\crefname{promptbox}{Prompt}{Prompts}
\Crefname{promptbox}{Prompt}{Prompts}
\crefname{proposition}{Proposition}{Propositions}
\Crefname{proposition}{Proposition}{Propositions}
\crefname{remark}{Remark}{Remarks}
\Crefname{remark}{Remark}{Remarks}
\crefname{section}{Section}{Sections}
\Crefname{section}{Section}{Sections}
\crefname{subsection}{Subsection}{Subsections}
\Crefname{subsection}{Subsection}{Subsections}
\crefname{subsubsection}{Subsubsection}{Subsubsections}
\Crefname{subsubsection}{Subsubsection}{Subsubsections}
\crefname{table}{Table}{Tables}
\Crefname{table}{Table}{Tables}
\crefname{theorem}{Theorem}{Theorems}
\Crefname{theorem}{Theorem}{Theorems}

\theoremstyle{plain}

\theoremstyle{definition}

\theoremstyle{remark}

\newfloat{promptbox}{tbp}{lop}
\floatstyle{plain}
\restylefloat{promptbox}
\floatname{promptbox}{Prompt}
\newcommand{\promptfile}[1]{%
\fbox{\begin{minipage}{.98\linewidth}
{\tt\fontsize{9}{11}\selectfont
\VerbatimInput[breaklines=true,breaksymbolleft={},breaksymbolright={}]{#1}}
\end{minipage}}
}

\definecolor{greatcolor}{RGB}{76,114,176}
\definecolor{okcolor}{RGB}{85,168,104}
\definecolor{badcolor}{RGB}{196,78,82}
\newcommand{\modegreat}{{\color{greatcolor} great}}
\newcommand{\modeok}{{\color{okcolor} ok}}
\newcommand{\modebad}{{\color{badcolor} bad}}

\newcommand{\abstracttext}{
Large language models (LLMs) face a fundamental trade-off between computational efficiency (e.g., number of parameters) and output quality, especially when deployed on computationally limited devices such as phones or laptops.
One way to address this challenge is by following the example of humans and have models ask for help when they believe they are incapable of solving a problem on their own; we can overcome this trade-off by allowing smaller models to respond to queries when they believe they can provide good responses, and deferring to larger models when they do not believe they can.
To this end, in this paper, we investigate the viability of Predict-Answer/Act (PA) and Reason-Predict-Reason-Answer/Act (RPRA) paradigms where models predict---prior to responding---how an LLM judge would score their output.
We evaluate three approaches: zero-shot prediction, prediction using an in-context report card, and supervised fine-tuning.
Our results show that larger models (particularly reasoning models) perform well when predicting generic LLM judges zero-shot, while smaller models can reliably predict such judges well after being fine-tuned or provided with an in-context report card.
Altogether, both approaches can substantially improve the prediction accuracy of smaller models, with report cards and fine-tuning achieving mean improvements of up to 55\% and 52\% across datasets, respectively.
These findings suggest that models can learn to predict their own performance limitations, paving the way for more efficient and self-aware AI systems.
}
\newcommand{\titletext}{
RPRA: Predicting an LLM-Judge for Efficient but Performant Inference
}

\begin{document}

\maketitle

\section{Introduction}\label{sec:introduction}

The rise of large language models (LLMs) has been the defining narrative of the recent AI boom.
The increasing investment in AI brought about by this has led to increasingly large resources being pooled into training LLMs.
The compounding effect of this has led to mainstream LLMs needing upward of a terabyte of GPU memory to run.
Such models are both incredibly costly to use (see, e.g., \citet{noffsinger2025cost}) and necessitate that consumer devices rely on API calls to access them---which users do, making more than 2.5 billion requests to ChatGPT a day \citep{roth2025openai}.
To address these limitations of large models, there has been growing interest in developing intentionally small models (e.g., \citet{chiang2023vicuna,metaai2024llama,projectapertus2025apertus}).

\begin{figure*}[t]
    \centering
    \includegraphics[width=\linewidth]{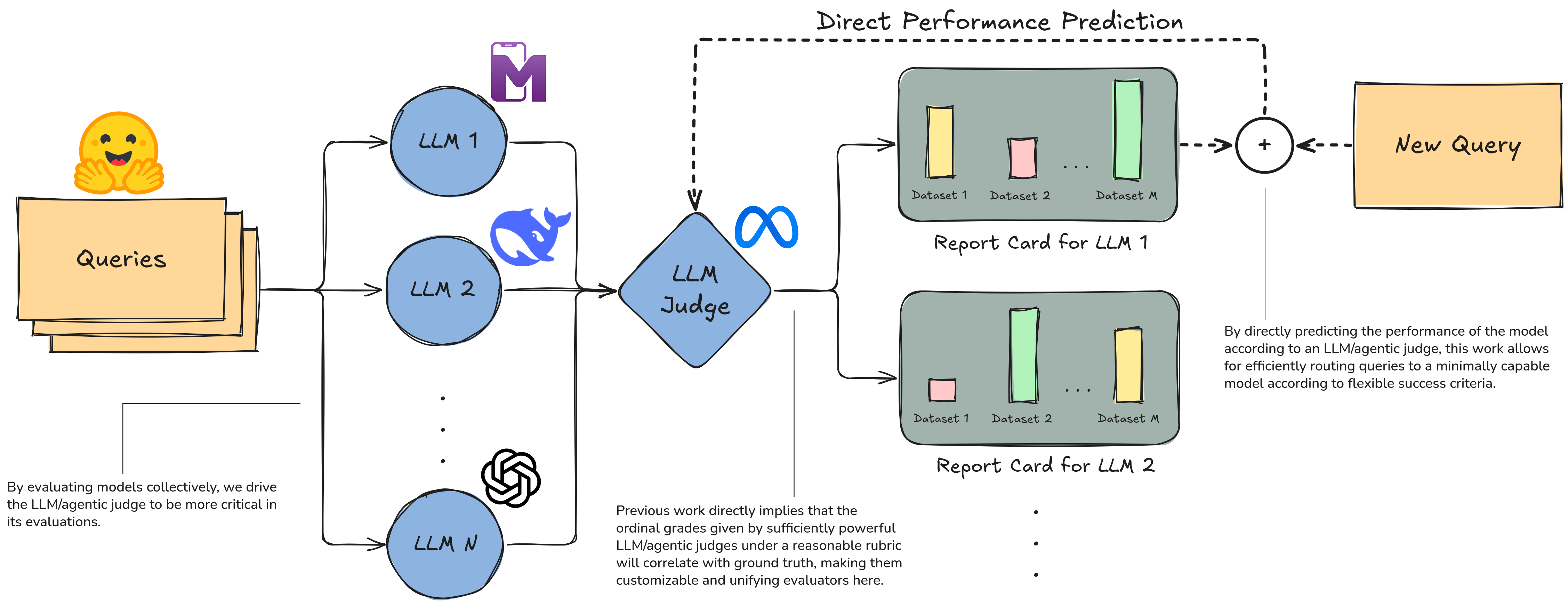}
    \caption{Overview of the report card framework. Queries are sent to multiple LLMs whose responses are jointly evaluated by an LLM/agentic judge. The judge's responses are converted into report cards that capture a model's performance across different query types. At inference time, the model predicts its own capability to answer a new query, enabling cost-efficient routing to a minimally capable model without the characteristic overconfidence of direct self-assessment.}
    \label{fig:nestor_flowchart}
\end{figure*}

Small models, such as the recent MobileLLM~\citep{liu2024mobilellm}, demonstrate comparable performance to much bigger models on many datasets.
However, they tend to demonstrate a much larger drop in their performance on other tasks compared with their bigger siblings~\citep{pecher2024comparing}.
This is particularly concerning as this performance drop is not uniform, leading to famously erratic responses such as models incorrectly responding to "how many Rs are there in strawberry?"\citep{blakesergin2024strawberrry}.
As such, it would be ideal to predict the quality of a response pre-hoc, allowing us to reduce the likelihood of an erratic response without needing to restrict ourselves to larger models.

Modern LLMs produce a wide variety of responses---from code and mathematical proofs to open-ended text and multi-step reasoning chains---making it difficult to define a single universal quality metric.
To evaluate such diverse outputs without heavily constraining the kinds of responses that can be given, one needs to employ something like an LLM-based (or agentic) judge~\citep{zheng2023judging,zhuge2024agent}.
Such judges are costly as generating output tokens is dramatically more expensive than processing input tokens~\citep{shi2024keep,li2025survey}.
This cost is both an increase in the literal cost and a delay to the user (as responses are often streamed to the user).
Additionally, as our results show, LLM judges exhibit a subpar evaluation of responses when doing so in isolation (e.g., compare \cref{fig:zero_shot_and_contextual_prediction_accuracy_with_finetuned_med_qa,fig:zero_shot_and_contextual_prediction_accuracy_with_finetuned_longfact,fig:zero_shot_and_contextual_prediction_accuracy_with_finetuned_mmlu_pro,fig:zero_shot_and_contextual_prediction_accuracy_with_finetuned_aime_2024,fig:zero_shot_and_contextual_prediction_accuracy_with_finetuned_scicode} with \cref{fig:independent_evaluation_scoreboard_med_qa,fig:independent_evaluation_scoreboard_longfact,fig:independent_evaluation_scoreboard_aime_2024,fig:independent_evaluation_scoreboard_mmlu_pro,fig:independent_evaluation_scoreboard_scicode} in Appendix~\ref{app:independent_evaluation_results}), making fair evaluation using them even more expensive.

We formalize this pre-hoc prediction of an LLM/agentic judge as two paradigms: Predict-Answer/Act (PA), where a model predicts how a judge would score its response before generating it, and Reason-Predict-Reason-Answer/Act (RPRA), which extends PA by having the model reason before predicting and again before responding.
This work investigates the viability of both paradigms across three approaches: tabula rasa (zero-shot), in-context learning, and fine-tuning.
In practice, a PA/RPRA model could directly respond to queries it is confident it can answer well, and trigger routing to a larger model otherwise.
LLM/agentic judges offer several advantages over traditional evaluation metrics: they provide reliable evaluations that correlate strongly with human evaluations~\citep{zheng2023judging,wang2024self,liu2023g,fu2024gptscore,zhuge2024agent}, support flexible evaluation criteria, and can be easily adapted to new alignment problems by modifying their prompts~\citep{zhuge2024agent}.
They also align strongly with ground truth values~\citep{zhuge2024agent}, allowing them to serve as effective unifying proxies
across different types of queries.

We evaluate our approaches across models of varying sizes: MobileLLM 0.9B, Llama 3.1 8B, Llama 3.2 1B and 3B, Llama 3.3 70B, GPT OSS 20B and 120B, DeepSeek Distilled Qwen 14B and 32B, DeepSeek Distilled 70B, and Llama 4 Scout.
Our findings reveal that while large models---particularly reasoning models---show a good ability to predict judge scores, smaller models typically suffer from miscalibration, being either overconfident or underconfident in their self-assessments.
This miscalibration reflects a well-documented tendency of LLMs toward overconfidence~\citep{huang2025confqa,ren2023self}.

To address these calibration issues, we propose two distinct approaches.
\textbf{Our first approach} provides models with a \textit{report card}: a detailed performance summary based on the model's historical performance across multiple datasets.
This method requires no additional training and can be applied to any model, including closed-weight systems (e.g., ChatGPT~\citep{openai2022introducing}).
We generate these report cards by evaluating models across diverse datasets (see \cref{fig:nestor_flowchart}) and using the modal judge ratings to build performance descriptions.

\textbf{Our second approach} fine-tunes models specifically for performance prediction, enabling the PA paradigm without requiring report cards.
While this requires additional training, it offers greater inference efficiency by eliminating the need to process report card tokens.
We construct training data using the hindsight trick~\citep{andrychowicz2017hindsight}, relabeling examples with judge scores, then apply supervised fine-tuning~\citep{ouyang2022training} to several model variants including MobileLLM 0.9B, Llama 3.1 8B, and Llama 3.2 1B and 3B.

Both approaches successfully improve smaller models' ability to predict their own performance, with effectiveness varying significantly across datasets.
Interestingly, models demonstrate stronger self-awareness on more challenging queries, suggesting that difficulty may serve as a useful signal for performance prediction.
These findings open promising directions for developing more reliable and self-aware small language models.

This paper presents a feasibility study for the PA and RPRA paradigms.
Our primary contributions can thus be summarized as that \textbf{(1)}~we propose the PA and RPRA paradigms where models estimate pre-hoc how their response to a query would be scored by an LLM or agentic judge (thereby allowing flexible evaluations) and show that large LLMs demonstrate a fair ability to do so tabula rasa; \textbf{(2)}~we propose using a report card system that summarizes the mode of the performance of a model in a number of datasets and demonstrate that it can dramatically improve the accuracy of many small models in the aforementioned prediction task; and \textbf{(3)}~we propose bypassing the inference cost of the report card system by fine-tuning models using the hindsight trick, demonstrating that this leads to strong prediction performance in many small models.
Notably, no complex architectures or training procedures were necessary for the above; the data needed to produce report cards is readily available and fine-tuning is often offered by service providers for even closed models.

\section{Preliminaries}\label{sec:preliminaries}

\subsection{Large Language Models (LLMs)}\label{sec:large_language_models}

LLMs are a class of (typically autoregressive) extremely large pretrained sequential models (in the billions of parameters), predominantly based on the transformer architecture~\citep{vaswani2017attention} (see also the earlier Unnormalized Linear Transformer~\citep{schmidhuber1992learning,schlag2021linear}).
They are trained on large text corpora, learning a probability distribution over a sequence of tokens which can be later used for text completion (i.e., text generation).
As such, these models excel at understanding, generating, and processing human language.
Mathematically, an LLM generates a sequence of tokens $y_1, \dots, y_L$ for a given input prompt $x$ by modeling the conditional probability
\begin{equation}
    P(y_1, \dots, y_L | x; \theta) = \prod_{t=1}^L P(y_t | y_{<t}, x; \theta),
\end{equation}
where $y_{<t}$ denotes previously generated tokens and $\theta$ represents the model's learned parameters.
The transformer architecture (the backbone of most LLMs) models a sequence of input embeddings $X = (x_1, \dots, x_n)$ using stacked layers of self-attention and feed-forward blocks.
Each self-attention block computes new representations $Z$ as
\begin{equation}
    Z = \mathrm{softmax}\left(\frac{QK^\top}{\sqrt{d_k}}\right)V,
\end{equation}
where $Q = XW_Q$, $K = XW_K$, $V = XW_V$ are learned linear projections of the input, and $d_k$ is the key dimension.
Positional encodings are added to $X$ to retain order information.
This architecture is exceptionally parallelizable and has an extraordinary ability for modeling long-range dependencies~\citep{yun2020transformers}.
Research has demonstrated that LLMs, particularly when augmented with an external read-write memory, are Turing complete, capable of simulating any algorithm~\citep{schuurmans2023memory}.
The behavior of an LLM is heavily guided by \textit{prompts}, with carefully crafted, context-rich instructions being essential for achieving high-quality and task-aligned outputs \citep{faang2024advanced}.
Although their pre-training imbues them with broad knowledge, to further align their output to human preferences and intents, LLMs are frequently fine-tuned (see, e.g., \citet{guo2025deepseek,muldrew2024active}).
This alignment helps to make models more helpful, harmless, and honest.

\subsection{Agentic Systems/Agentic AI}\label{sec:agentic_systems}

Building upon the capabilities of individual LLMs, \textit{agentic systems} (sometimes called Agentic AI) are modular AI architectures that orchestrate one or more LLM invocations through arbitrary control flow, often integrating external tool calls \citep{du2024improving,schick2023toolformer,zeng2023socratic,zhuge2023mindstorms}.
These systems differ from standalone LLMs by their ability to engage in multi-step reasoning, planning, and acting to solve complex tasks autonomously or semi-autonomously \citep{zhuge2023mindstorms,zhuge2024agent}.
This multi-step approach enables decomposition of complex problems into manageable subtasks and allows for iterative refinement based on intermediate results---capabilities that single-pass LLM inference cannot achieve \citep{zhuge2024agent}.
The internal workflow of an agentic system can be conceptualized as a directed graph $G=(V,E)$, where each node $v \in V$ represents an LLM call, tool execution, or decision point, and the edges $(u,v) \in E$ denote dependencies between actions \citep{zhuge2024agent}.
Tool use represents a fundamental form of agentic AI, enabling systems to dynamically interact with environments and access specialized capabilities such as code interpreters and search engines \citep{schick2023toolformer,qin2024toolllm}.
This approach extends LLM capabilities beyond their training limitations through external resource integration.

A prominent application of agentic systems is automated evaluation through \textit{LLM-as-a-Judge} frameworks, which utilize LLMs to assess text generation quality and model behavior \citep{li2024llms}.
These frameworks can be extended into \textit{Agent-as-a-Judge} systems that leverage full agentic capabilities to provide richer evaluation feedback, including assessment of tool use and multi-step reasoning processes \citep{zhuge2024agent}.

\section{Experimental Setup}\label{sec:experimental_setup}

\begin{table*}[t]
    \centering
    \begin{tabular}{lcccc}
        \textbf{Shorthand} & \textbf{Full Name} & \textbf{\# Parameter} & \textbf{Reasoning} & \textbf{Reference} \\
        \hline \vspace{-0.75em}\\
        M09B & MobileLLM 0.9B & 0.9 Billion & No & \citet{liu2024mobilellm} \\
        L318B & Llama 3.1 8B Instruct & 8.03 Billion & No & \citet{grattafiori2024llama} \\
        L321B & Llama 3.2 1B Instruct & 1.24 Billion & No & \citet{metaai2024llama} \\
        L323B & Llama 3.2 3B Instruct & 3.21 Billion & No & \citet{metaai2024llama} \\
        L3370B & Llama 3.3 70B Instruct & 70.6 Billion & No & \citet{metaai2024llama} \\
        L416E & Llama 4 Scout 17B 16E Instruct & 109 Billion & No & \citet{metaai2025llama} \\
        DSQ14B & DeepSeek R1 Distilled Qwen 14B & 14.8 Billion & Yes & \citet{guo2025deepseek} \\
        DSQ32B & DeepSeek R1 Distilled Qwen 32B & 32.8 Billion & Yes & \citet{guo2025deepseek} \\
        DSL70B & DeepSeek R1 Distilled Llama 70B & 70.6 Billion & Yes & \citet{guo2025deepseek} \\
        GPT20B & GPT OSS 20B & 21.5 Billion & Yes & \citet{openai2025introducing} \\
        GPT120B & GPT OSS 120B & 120 Billion & Yes & \citet{openai2025introducing} \\
    \end{tabular}
    \caption{Models considered for our experiments. The parameter count (in billions) is taken from the HuggingFace Safetensor values to ensure equal treatment between models.}
    \label{tab:models}
\end{table*}

\begin{wrapfigure}[18]{r}{0.48\textwidth}
    \centering
    \vspace{-0.7em}
    \includegraphics[width=\linewidth]{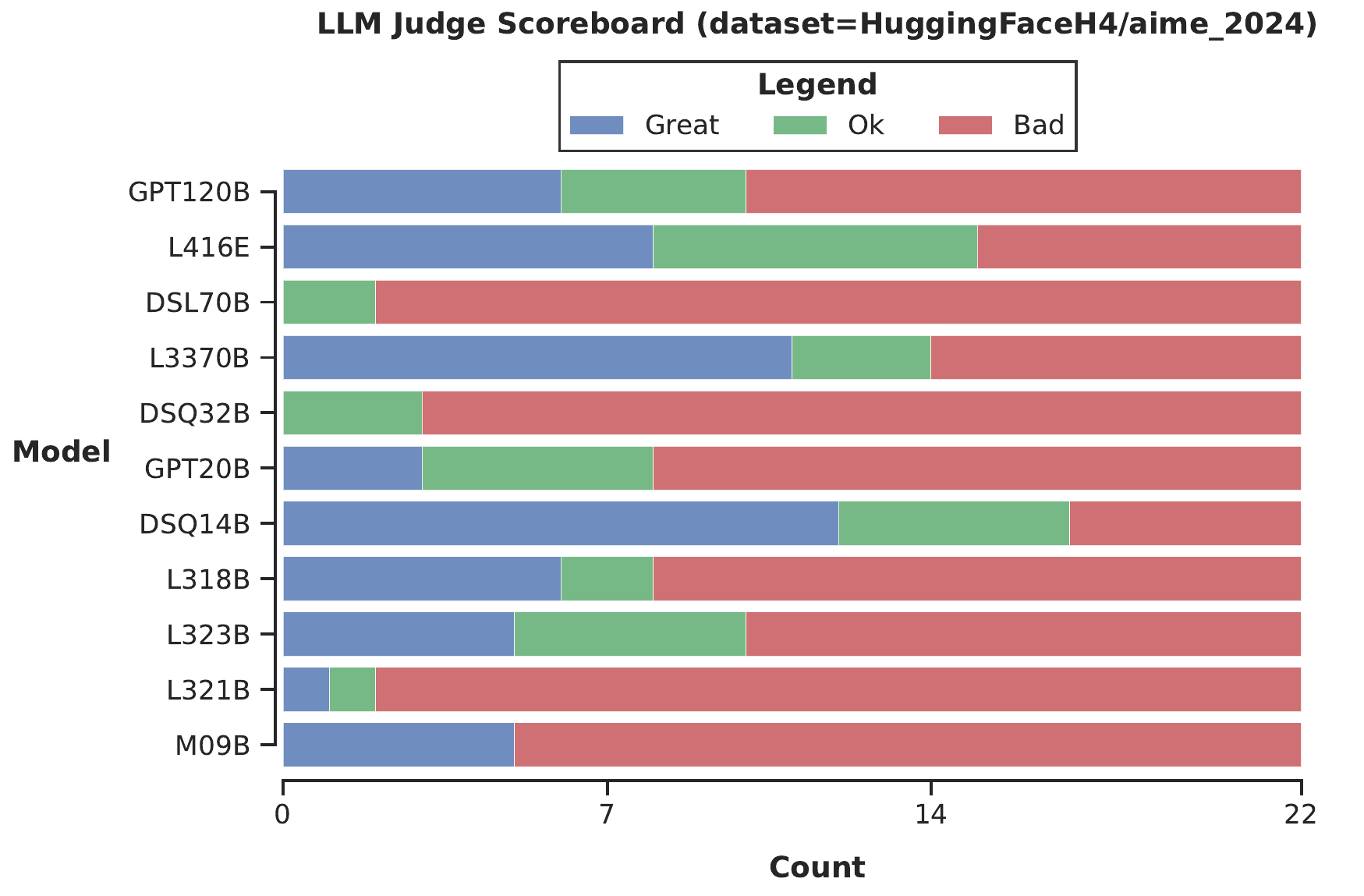}
    \caption{The distribution of the LLM judge scores for each of the models on the AIME 2024 dataset. Note that the poor performance of some reasoning models here was due to us limiting reasoning models to producing no more than 24576 tokens.}
    \label{fig:scoreboard_aime_2024}
\end{wrapfigure}

We experiment with five (5) datasets that are commonly used in the literature: MedQA, which contains queries asking for a medical diagnosis~\citep{jin2020disease}; LongFact, which contains factual trivia~\citep{wei2024long}; AIME 2024, which contains competition-level maths problems~\citep{huggingfaceh42025aime}; SciCode, which contains requests for producing code for scientific purposes~\citep{tian2024scicode}; and MMLU-Pro, which contains general undergraduate-level examination questions~\citep{wang2024mmlu}.
Altogether, these datasets cover a broad range of application cases for LLMs.

In total, we experimented with the eleven (11) models shown in \cref{tab:models}.
We use Llama 3.3 70B as our judge, instructing it to evaluate all the responses of the models to the query (we also provided any relevant answers included in the dataset to the judge) according to a predefined rubric.
Llama 3.3 70B remains a well-studied model with robust performance, making it well-suited to use as a judge here.
We include an ablation using GPT OSS 120B as a judge in Appendix~\ref{app:alternate_judge_results}, which confirms that our findings generalize across evaluators.
The rubric used is shown in \cref{ppt:rubric} alongside all the other prompts used in Appendix~\ref{app:prompt_templates}.
To ground evaluations more, we had the judge evaluate simultaneously the responses of all the models for one query simultaneously.
We performed an ablation experiment that confirms the independent evaluations led to less diversity between the models in Appendix~\ref{app:independent_evaluation_results}.
In addition to providing a grade for each response of either ``great,'' ``ok,'' or ``bad,'' the judge was instructed to provide a rationale for why it gave each response the grade it gave them.
We selected Llama 3.3 70B as our ad-hoc observations found that it was able to effectively follow the instructions given and generally gave good rationales for its grading.
In cases where the response given by the judge could not be parsed, one request was made to correct it.
Cases where the corrected response could not be parsed were negligible.

\begin{figure}[t]
    \centering
    \begin{minipage}{0.48\textwidth}
        \centering
        \vspace{1em}

        \includegraphics[width=\linewidth]{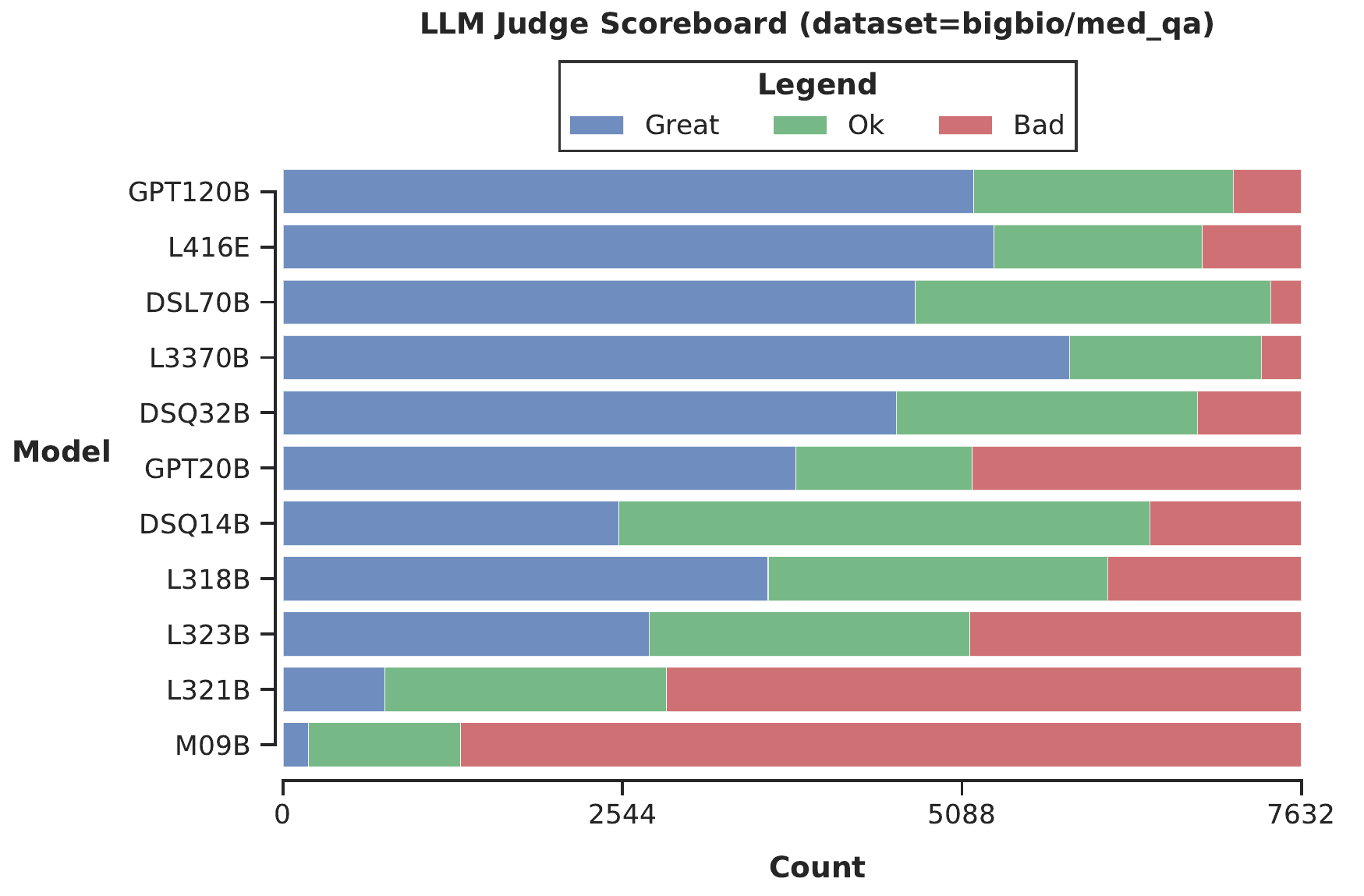}
        \caption{The distribution of the LLM judge scores for each of the models on the MedQA dataset. Note the wide spread of performance here across the different models.}
        \label{fig:scoreboard_med_qa}

        \vspace{2em}

        \includegraphics[width=\linewidth]{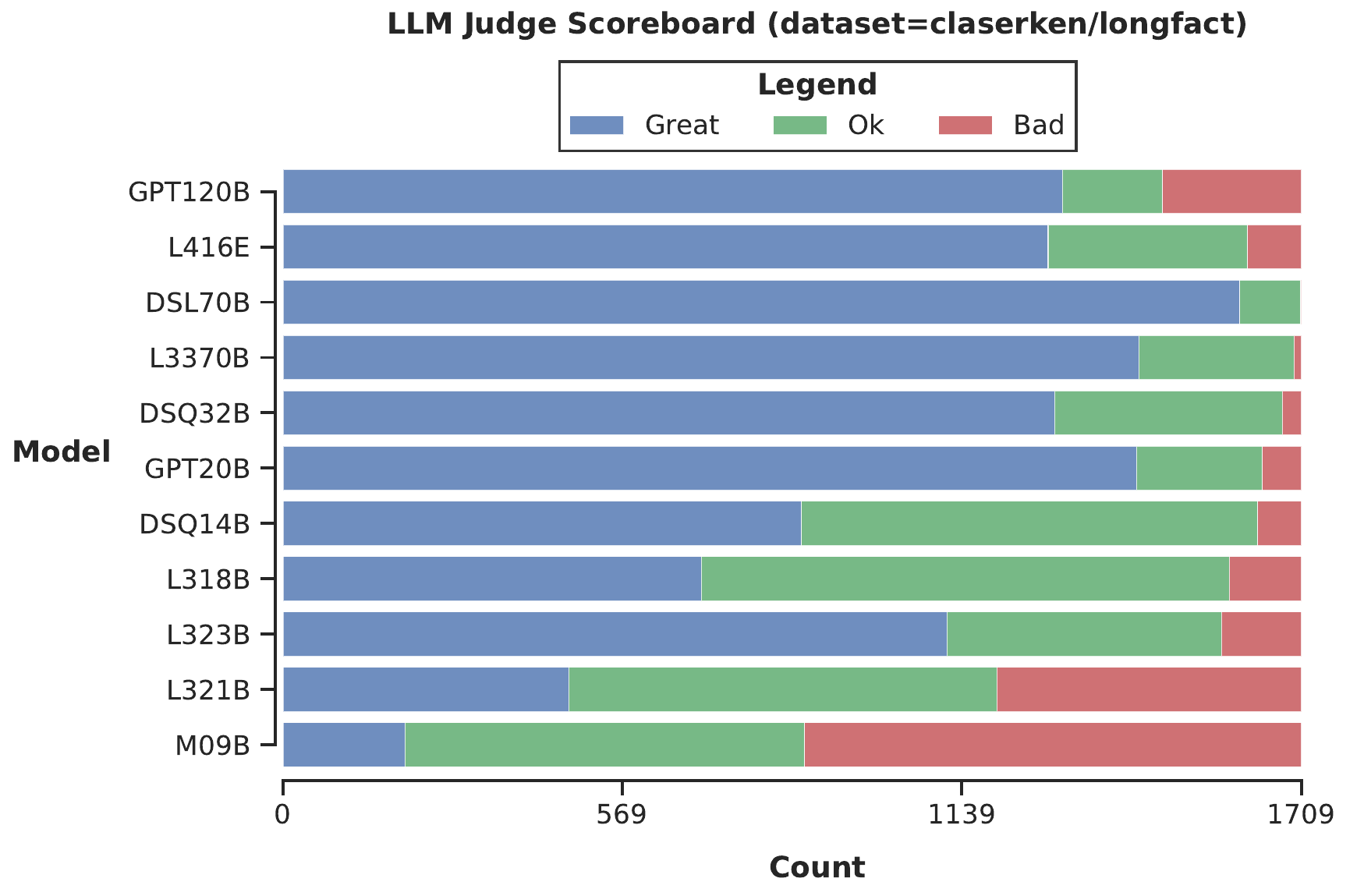}
        \caption{The distribution of the LLM judge scores for each of the models on the LongFact dataset. Much of this information is in-distribution, so most LLMs can do well here.}
        \label{fig:scoreboard_longfact}
    \end{minipage}
    \hfill
    \begin{minipage}{0.48\textwidth}
        \centering

        \includegraphics[width=\linewidth]{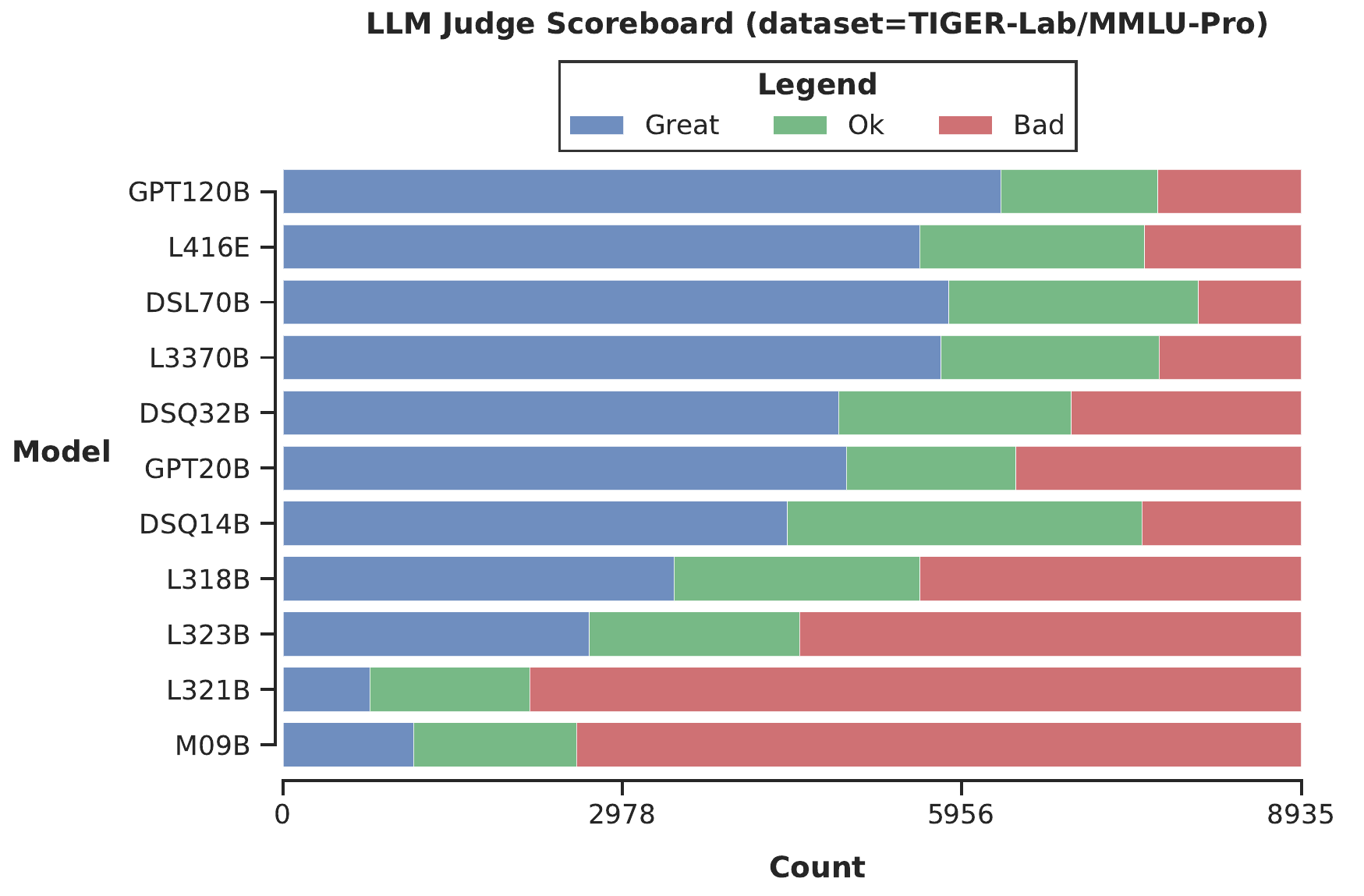}
        \caption{The distribution of the LLM judge scores for each of the models on the MMLU-Pro dataset. We provide a per-category breakdown in \cref{fig:mmlu_pro_dual_heatmap} in \cref{sec:mmlu_pro_category_results}.}
        \label{fig:scoreboard_mmlu_pro}

        \vspace{2em}

        \includegraphics[width=\linewidth]{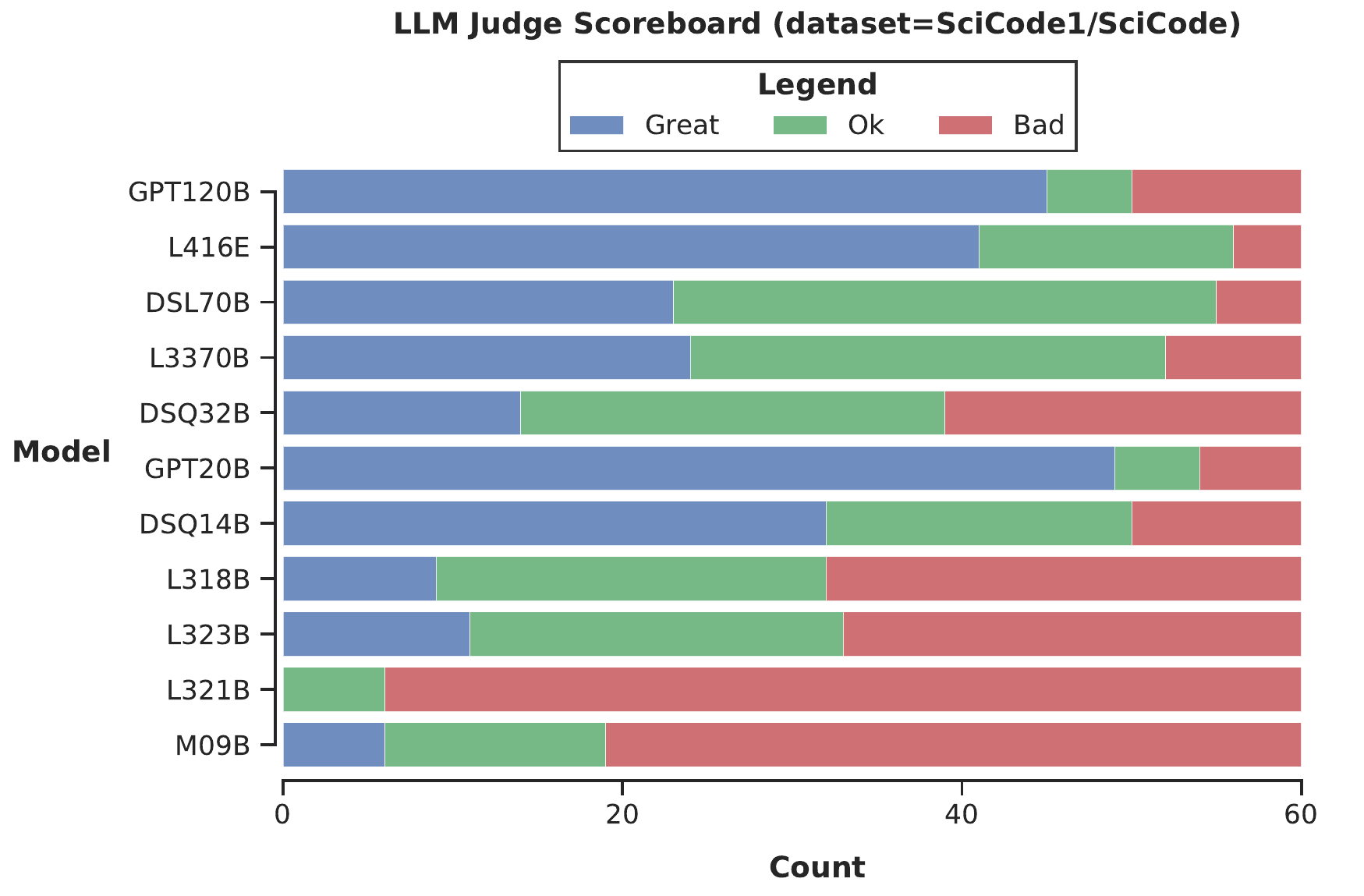}
        \caption{The distribution of the LLM judge scores for each of the models on the SciCode dataset. This dataset involves coding so most models struggle here.}
        \label{fig:scoreboard_scicode}
    \end{minipage}
\end{figure}

The distribution of scores given by the LLM judge for each of the models on each of the datasets is shown in \cref{fig:scoreboard_med_qa,fig:scoreboard_longfact,fig:scoreboard_aime_2024,fig:scoreboard_scicode,fig:scoreboard_mmlu_pro}.
These results roughly conform to expectations, with most models doing \textbf{(1)}~well with basic factual information, \textbf{(2)}~poorly with mathematic questions, \textbf{(3)}~larger and reasoning models doing better than smaller and non-reasoning models, and \textbf{(4)}~newer models doing better for their size than older models.

\section{Zero-Shot and In Context Prediction}\label{sec:zero_shot_and_in_context_prediction}

LLMs have demonstrated strong generalization abilities across a broad range of tasks \citep{brown2020language,wei2022emergent}, making it reasonable to think that a model might be able to predict their performance pre-hoc if the judge is provided with a generic rubric.
We experiment with the model's given in \cref{tab:models} ability to directly predict how an LLM judge would score their response without any context apart from the query (see \cref{ppt:zero_shot_score_prediction_template} in Appendix~\ref{app:prompt_templates} for the exact query given to the model here).
The context for all the models was limited to one round of interactions, i.e., one query.
The prediction performance of the models in the zero-shot case is shown alongside our other results in \cref{fig:zero_shot_and_contextual_prediction_accuracy_with_finetuned_med_qa,fig:zero_shot_and_contextual_prediction_accuracy_with_finetuned_longfact,fig:zero_shot_and_contextual_prediction_accuracy_with_finetuned_mmlu_pro,fig:zero_shot_and_contextual_prediction_accuracy_with_finetuned_aime_2024,fig:zero_shot_and_contextual_prediction_accuracy_with_finetuned_scicode}.
We provide a summary of these results in \cref{tab:contextual_summary} in Appendix~\ref{app:summary_table_for_contextual_results}.

\begin{figure*}[p]
    \centering
    \begin{minipage}[t]{0.48\textwidth}
        \centering

        \includegraphics[width=\linewidth]{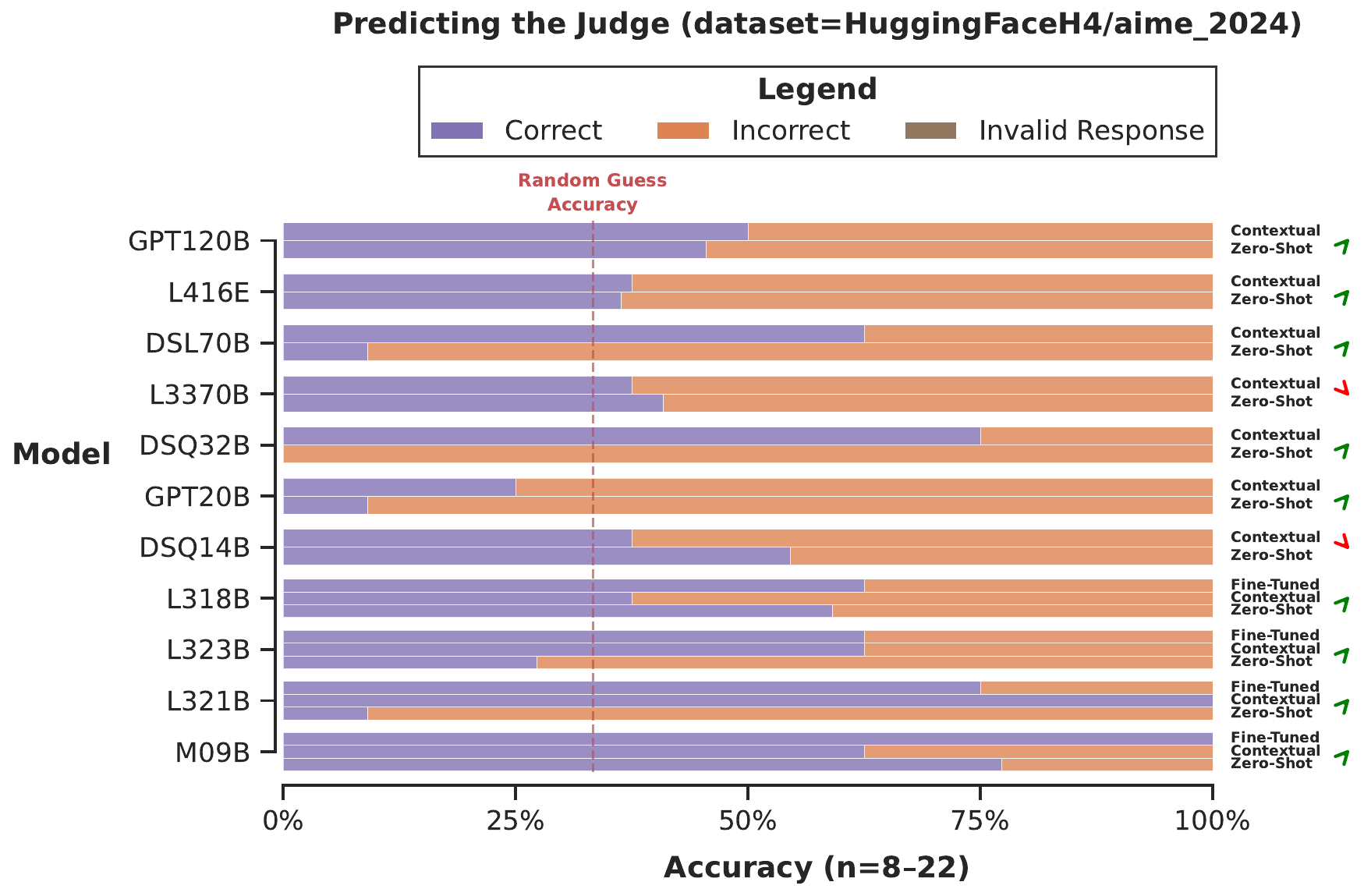}
        \caption{The zero-shot and contextual prediction accuracy of all the models on the AIME 2024 dataset (a green arrow indicates improvement over zero-shot). Note the dramatic improvement for smallest models here.}
        \label{fig:zero_shot_and_contextual_prediction_accuracy_with_finetuned_aime_2024}

        \vspace{1.375em}

        \includegraphics[width=\linewidth]{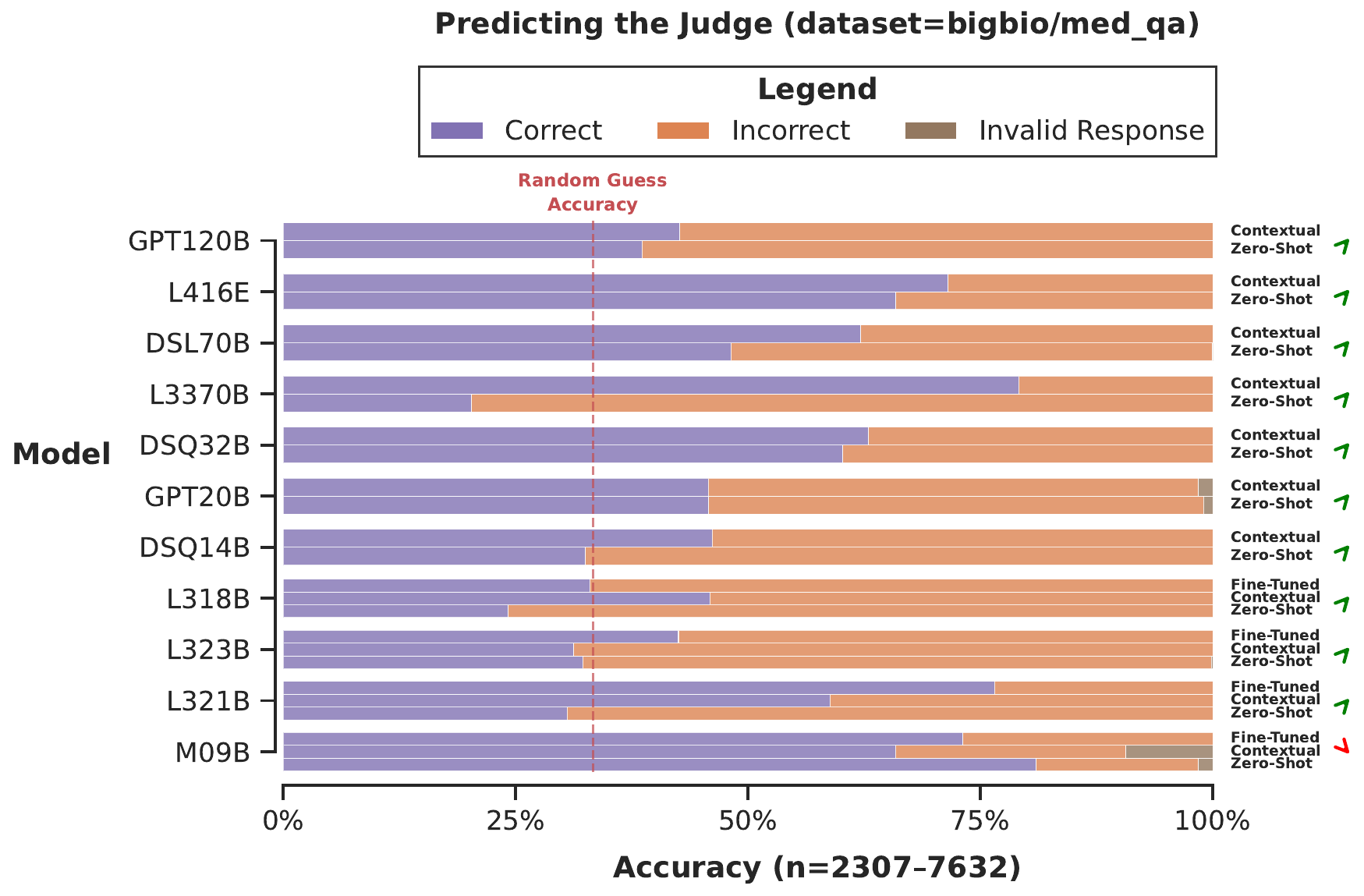}
        \caption{The zero-shot and contextual prediction accuracy of all the models on the MedQA dataset (a green arrow indicates improvement over zero-shot). Note how dramatic the improvement offered to even big non-reasoning models is (i.e., L3370B here).}
        \label{fig:zero_shot_and_contextual_prediction_accuracy_with_finetuned_med_qa}

        \vspace{1.375em}

        \includegraphics[width=\linewidth]{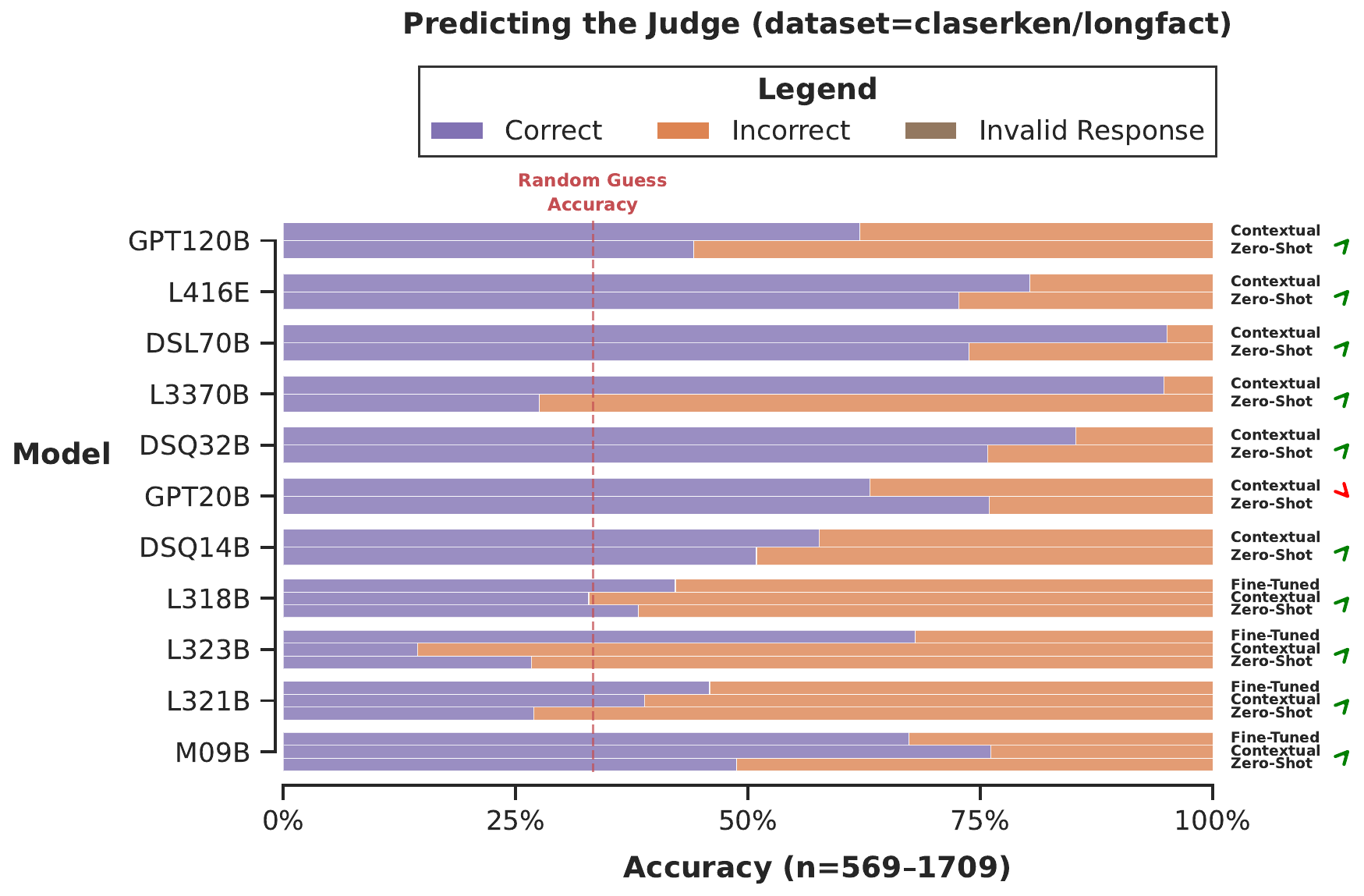}
        \caption{The zero-shot and contextual prediction accuracy of all the models on the LongFact dataset (a green arrow indicates improvement over zero-shot). Note that, even on a dataset where all the models do well, their predictions are better in the contextual/fine-tuning setting.}
        \label{fig:zero_shot_and_contextual_prediction_accuracy_with_finetuned_longfact}

        \vfill
    \end{minipage}
    \hfill
    \begin{minipage}[t]{0.48\textwidth}
        \centering

        \includegraphics[width=\linewidth]{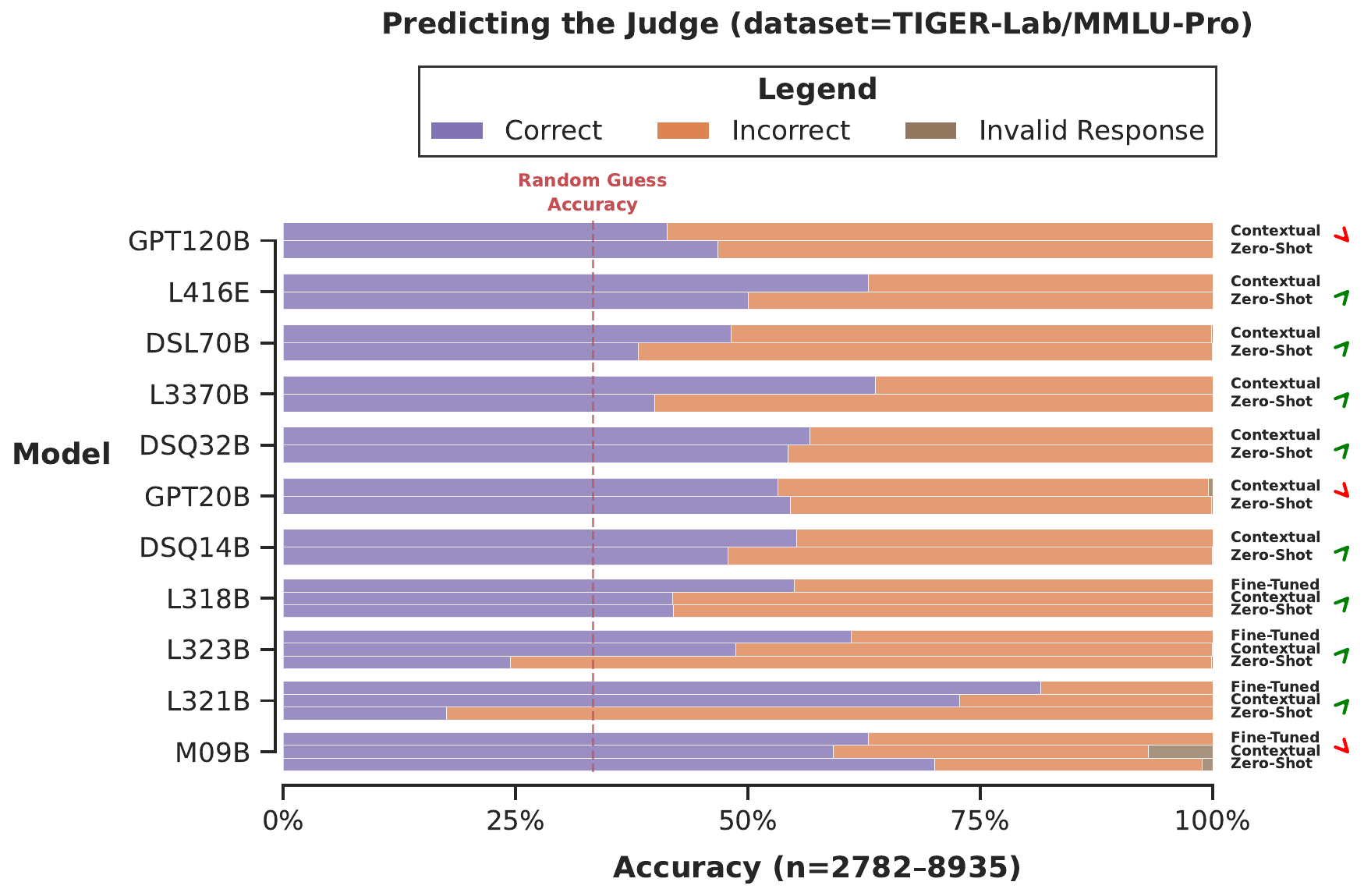}
        \caption{The zero-shot and contextual prediction accuracy of all the models on the MMLU-Pro dataset (a green arrow indicates improvement over zero-shot). We provide a per-category breakdown of the zero-shot results in \cref{fig:mmlu_pro_dual_heatmap} in Section~\ref{sec:mmlu_pro_category_results}.}
        \label{fig:zero_shot_and_contextual_prediction_accuracy_with_finetuned_mmlu_pro}

                    \vspace{1.375em}

        \includegraphics[width=\linewidth]{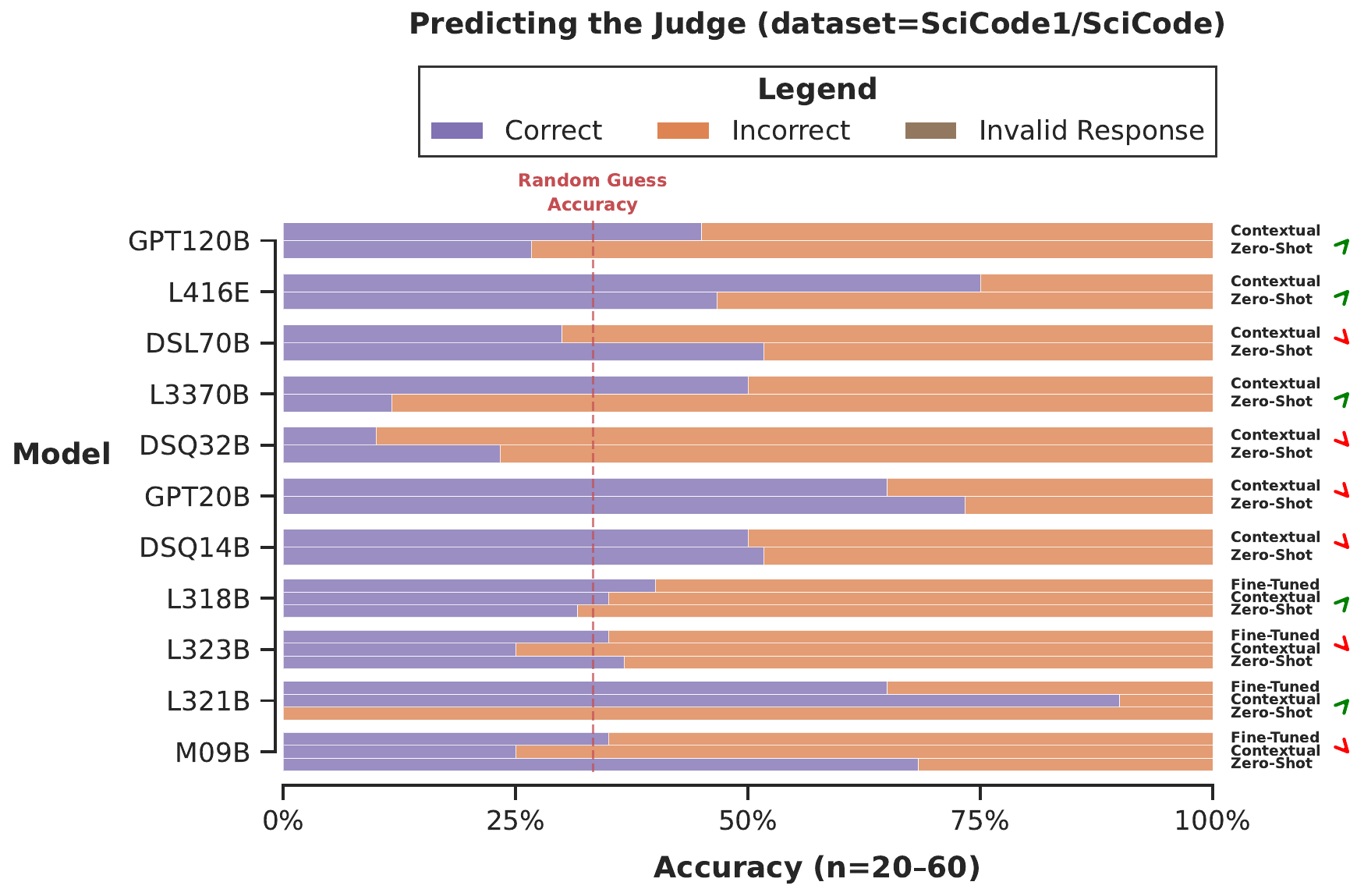}
        \caption{The zero-shot and contextual prediction accuracy of all the models on the SciCode dataset (a green arrow indicates improvement over zero-shot). This is the only dataset where the contextual/fine-tuning approach frequently didn't help. We hypothesize that the judge was not sophisticated enough to evaluate models on this dataset (see, e.g., \citet{zhuge2024agent}).}
        \label{fig:zero_shot_and_contextual_prediction_accuracy_with_finetuned_scicode}

                    \vspace{3.25em}

        \setlength{\tabcolsep}{2pt}
        \begin{tabular}{lcccc}
            \textbf{Dataset} & \textbf{M09B} & \textbf{L321B} & \textbf{L323B} & \textbf{L318B} \\
            \hline  \vspace{-0.75em}\\
            MedQA & -.09 & +.46 & +.10 & +.09 \\
            LongFact & +.19 & +.19 & +.41 & +.04 \\
            AIME 2024 & +.23 & +.66 & +.35 & +.03 \\
            MMLU-Pro & -.07 & +.64 & +.37 & +.13 \\
            SciCode & -.33 & +.65 & -.02 & +.08  \vspace{0.25em}\\
            \hline \vspace{-0.75em}\\
            \textit{Mean} & -.02 & \textbf{+.52} & +.24 & +.07 \\
        \end{tabular}
        \captionof{table}{Summary of the improvement in the prediction accuracy for the fine-tuned models as compared with the zero-shot setting. For a summary of the contextual setting results, see \cref{tab:contextual_summary} in Appendix~\ref{app:summary_table_for_contextual_results}.}
        \label{tab:fine_tuning_summary}

        \vfill
    \end{minipage}
\end{figure*}

\textbf{Zero-shot Results.}\enspace
The key observation to take from the zero-shot results in \cref{fig:zero_shot_and_contextual_prediction_accuracy_with_finetuned_med_qa,fig:zero_shot_and_contextual_prediction_accuracy_with_finetuned_longfact,fig:zero_shot_and_contextual_prediction_accuracy_with_finetuned_mmlu_pro,fig:zero_shot_and_contextual_prediction_accuracy_with_finetuned_aime_2024,fig:zero_shot_and_contextual_prediction_accuracy_with_finetuned_scicode} is that \textbf{(1)}~the prediction performance of the models varied wildly based on the dataset used, \textbf{(2)}~reasoning models exhibited an often better ability to estimate their own performance, and \textbf{(3)}~smaller models typically performed at or worse than a random guess accuracy.

Based on the large variation observed in model and prediction performance by dataset, it seems reasonable that a promising approach would be to determine the kind of query and predict based on that.
To that end, we produce a ``report card'' for each model based on their performance on the datasets (see \cref{fig:scoreboard_med_qa,fig:scoreboard_longfact,fig:scoreboard_aime_2024,fig:scoreboard_scicode,fig:scoreboard_mmlu_pro}).
This report card simply provides the mode of the scores for that model on that dataset.
These mode scores are given in Appendix~\ref{app:mode_scores} with the report card structure shown in \cref{ppt:long_feedback_template} in Appendix~\ref{app:prompt_templates} (we provide an ablation on the report card structure in Appendix~\ref{app:short_feedback_template_results}).

\textbf{Contextual Results.}\enspace
The results of the report card-based, or contextual prediction performance are given alongside the previous results in \cref{fig:zero_shot_and_contextual_prediction_accuracy_with_finetuned_aime_2024,fig:zero_shot_and_contextual_prediction_accuracy_with_finetuned_longfact,fig:zero_shot_and_contextual_prediction_accuracy_with_finetuned_med_qa,fig:zero_shot_and_contextual_prediction_accuracy_with_finetuned_mmlu_pro,fig:zero_shot_and_contextual_prediction_accuracy_with_finetuned_scicode}.
The effect of the report card is particularly good on the smaller and non-reasoning models, enabling them to work in a PA paradigm.
This suggests that the inclusion of the report card was effective with the small models on allowing them to categorize the query and match it to their skill sets, validating the hypothesis that even a small model can work in a PA paradigm with the right information.
We further validate the report card approach with the ablation given in Appendix~\ref{app:mischievous_rubric_results}, which confirms that the approach is reasonably robust to even major variations in the rubric.

\subsection{MMLU-Pro Category Results}\label{sec:mmlu_pro_category_results}

\begin{wrapfigure}[20]{r}{0.48\textwidth}
    \centering
    \vspace{-1.7em}
    \includegraphics[width=\linewidth]{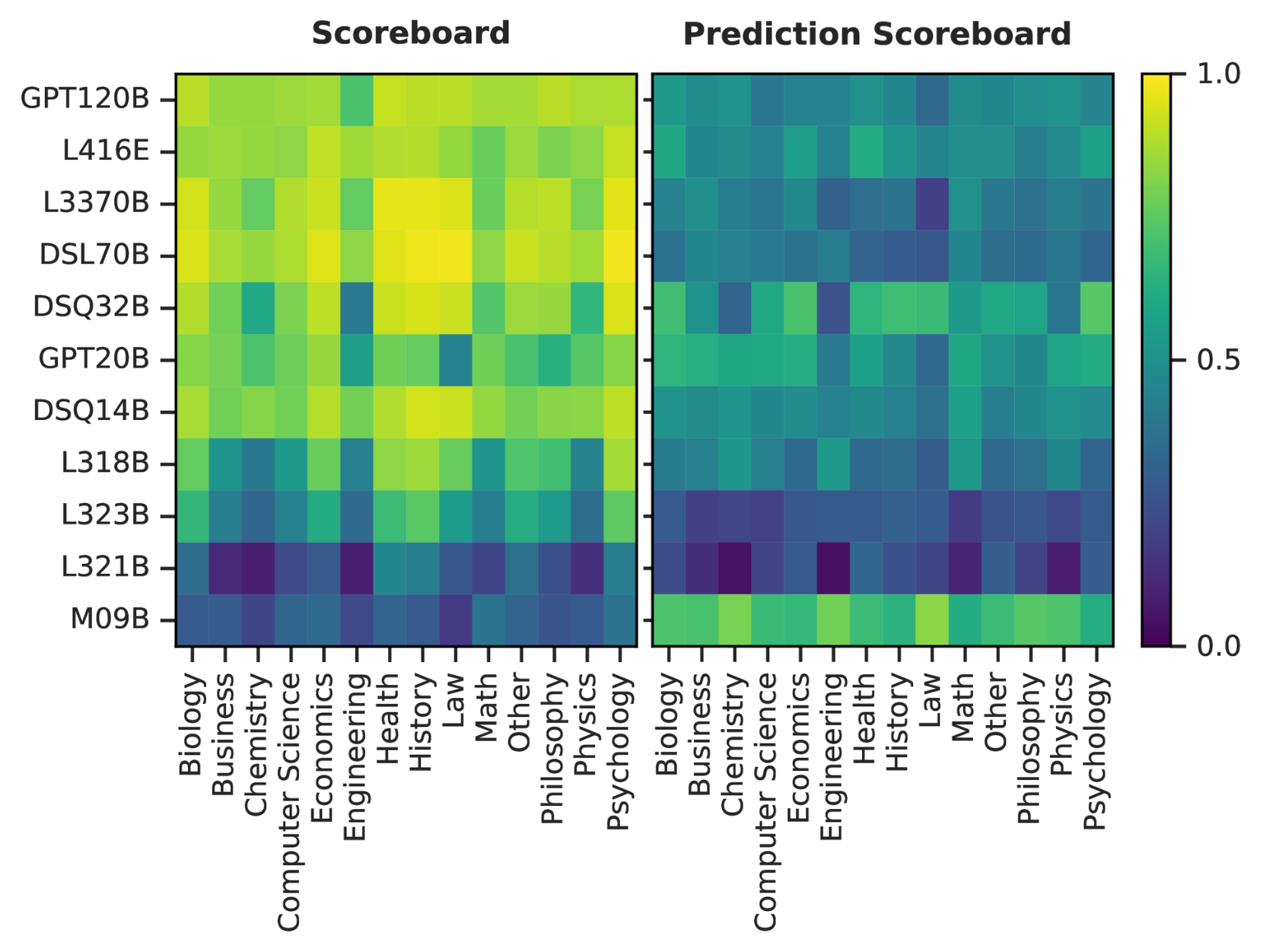}
    \caption{The probability of a model receiving a grade of ``great'' or ``ok'' (left) and correctly predicting its score zero-shot (right) as a function of the query's category on the MMLU-Pro dataset.}
    \label{fig:mmlu_pro_dual_heatmap}
\end{wrapfigure}

One common argument against LLM judges is that they might be unreliable.
While this has been refuted (see, e.g., \citet{zheng2023judging}), this does not preclude the possibility that the particular judge used here could be highly stochastic.
To test this, we split the queries in the MMLU-Pro dataset up by category and observed the performance score distribution of the models and their predictions as a function of the category.
\cref{fig:mmlu_pro_dual_heatmap} shows the probability that the judge gives a score of "great" or "ok" for each model on the MMLU-Pro dataset as a function of the category.
The conclusion that can be drawn from \cref{fig:mmlu_pro_dual_heatmap} is that the distribution of scores given by the judge is less affected by the category than by the model, implying a comparatively lower variance in the judging.

\cref{fig:mmlu_pro_dual_heatmap} also shows the probability that the model correctly predicts its score zero-shot.
Again, the variance is more visible across models than categories, suggesting that the categorization ability of the models has not been deeply affected by the variance of the judge (note that we would expect roughly equal performance on a dataset given the homogeneity of the datasets in the kind of queries they ask and the broad generalization of the models in certain kinds of queries).

\section{Fine-Tuning Approach}\label{sec:fine_tuning_approach}

The biggest drawback of using report cards to allow a model to estimate its performance is that it necessitates a model processing the large number of tokens needed for a comprehensive report card.
While the cost of producing input tokens is relatively inexpensive compared to output tokens~\citep{shi2024keep,li2025survey}, we nevertheless conducted an ablation where we use shorter report cards (see Appendix~\ref{app:short_feedback_template_results}), but this resulted in a notable drop in prediction accuracy.
To get the best of both worlds, in this section we experimented with fine-tuning versions of the small models to perform these predictions without a report card.

To accomplish the above, we produced a dataset of zero-shot prompts and the respective zero-shot model evaluations collected in \cref{sec:zero_shot_and_in_context_prediction}.
We then applied the hindsight trick (see, e.g., \citet{andrychowicz2017hindsight}) to relabel the zero-shot predictions as though they were accurate.
Afterwards, we applied supervised fine-tuning~\citep{ouyang2022training} using this data to MobileLLM 0.9B, Llama 3.1 8B, and Llama 3.2 1B and 3B.
We refer to these models fine-tuned to predict as MP09B, LP318B, LP321B, and LP323B, respectively.
Note that none of the base models are reasoning models (so these models enable the PA rather than RPRA paradigm) and the respective output for the prediction should be one token.

The results of this fine-tuning approach are presented alongside our other results in \cref{fig:zero_shot_and_contextual_prediction_accuracy_with_finetuned_aime_2024,fig:zero_shot_and_contextual_prediction_accuracy_with_finetuned_longfact,fig:zero_shot_and_contextual_prediction_accuracy_with_finetuned_med_qa,fig:zero_shot_and_contextual_prediction_accuracy_with_finetuned_mmlu_pro,fig:zero_shot_and_contextual_prediction_accuracy_with_finetuned_scicode}, with an arrow annotation indicating if either the contextual or fine-tuning approach improved the prediction performance (a green arrow denotes an improvement).
We provide a summary of these results in \cref{tab:fine_tuning_summary}.
The prediction quality of the fine-tuned models is clearly at or above the performance of even the report card-based predictions.
However, as with the report cards, the prediction quality is heavily dependent on the dataset and is much stronger in the smaller models (where the distribution of judge evaluations is more concentrated).

Altogether, this suggests that models fine-tuned to predict can enable the PA paradigm with prediction performance at or above the endowment given by a report card.

\section{Related Work}\label{sec:related_work}

We present the most relevant related work below, with an additional supplementary discussion of other (less-)related work in Appendix~\ref{app:less_related_works}.

\textbf{LLM and Agentic Judges.}\enspace
The advent of Large Language Models (LLMs) has catalyzed their adoption as automated evaluators, particularly through the \textit{LLM-as-a-Judge} paradigm \citep{zheng2023judging,liu2023g,fu2024gptscore}.
This approach leverages powerful foundation models to assess text quality, dialogue performance, and agent behavior, offering a scalable and cost-effective alternative to human evaluation.
Studies demonstrate that single-LLM judges can achieve high correlation (Spearman coefficients of $\approx 0.8$) with aggregated human preferences across diverse tasks \citep{zheng2023judging}.

However, this paradigm faces big limitations.
Single-judge systems are prone to systematic biases, including preferences for specific output lengths, styles, or verbosity \citep{dubois2024length}.
They also exhibit vulnerabilities to adversarial attacks and may fail to detect nuanced factual inaccuracies or complex reasoning errors \citep{eiras2025know,li2025llms}.
Recent work has further highlighted position bias \citep{wang2024large} and limited reasoning depth in complex evaluation scenarios.

To address these challenges, the research community has progressed to \textit{Agent-as-a-Judge} and multi-agent evaluation frameworks.
These systems employ multiple LLM agents that collaborate, debate, or assume specialized roles (e.g., Grader, Critic, Defender) to produce more robust and reliable assessments \citep{chen2025multi}.
Notable implementations include \textsc{MAJ-EVAL} \citep{chen2025multi}, featuring a multi-agent jury; \textsc{AgentsCourt} \citep{he2024agentscourt} for competitive agent environments; and \textsc{FinCon} \citep{yu2024fincon} for multi-agent financial decision making.
Recent advancements also explore \textit{debate-based evaluation} \citep{chan2023chateval} and \textit{iterative refinement} processes \citep{liang2024encouraging} to enhance judgment quality.

Nevertheless, multi-agent judges introduce new complexities, such as potential collusion when agents share similar model backbones, emergent group biases, and increased computational costs \citep{chan2023chateval,liang2024encouraging}.
The search for optimal agent architectures and interaction protocols remains an active area of research (e.g., see the work by \citet{qin2024toolllm} or \citet{wu2023autogen}).
\textbf{Our work diverges} from these directions by not aiming to improve the judging mechanism itself.
Instead, we focus on \textit{preemptively predicting the scores} that an LLM or agentic judge would assign to a model's response \textit{before} its generation.
This approach enables more efficient system design, rapid prototyping, and optimal resource allocation by shifting the focus from \textit{being} the evaluator to \textit{anticipating} the evaluation outcome.

\textbf{Predicting Model Performance.}\enspace
Another line of research explores models' abilities to self-assess or predict aspects of their own performance, particularly regarding confidence estimation and hallucination mitigation.
\textsc{ConfQA}, for example, introduces a fine-tuning strategy that explicitly trains LLMs to express uncertainty when lacking confidence in factual statements, significantly reducing hallucination rates through dampening prompts and factual calibration \citep{huang2025confqa}.
Similarly, \citet{ren2023self} proposed using self-evaluation scores for selective generation, where LLMs decide when to abstain from generating to boost accuracy by reformulating open-ended tasks into token-level predictions.

\textsc{DeepConf} contributes to efficient reasoning by filtering low-confidence reasoning traces and enabling early termination based on local confidence signals, thereby optimizing computational overhead \citep{fu2025deep}.

\textbf{Our work diverges} significantly from these internal confidence mechanisms.
Rather than focusing on self-assessment or selective generation, we aim to predict the scores assigned by \textit{external} LLM or agentic judges to a model's response \textit{before} generation occurs.
This involves learning to anticipate evaluator judgments directly, encompassing both zero-shot and contextual predictions based on historical performance data.
This pre-hoc prediction enables resource optimization without requiring real-time introspection or response generation.

\section{Conclusion}\label{sec:conclusion}

In this work, we examined the viability of PA and RPRA paradigms where language models predict the quality of their own responses as judged by an external agentic (LLM-based) evaluator.
Our results show that while larger models tend to be better calibrated in their self-assessments, smaller models often exhibit overconfidence or underconfidence.
To address this, we proposed two approaches: providing models with report cards summarizing their historical performance, and fine-tuning models specifically for performance prediction.
Both methods enable smaller models to follow the PA paradigm effectively, with the degree of improvement depending on model size and task difficulty.
Altogether, we believe that this work represents a step forward in increasing the usability of small language models by enabling effective PA and RPRA paradigms.

\section{Limitations \& Future Work}\label{sec:limitations_and_future_work}

This work focuses on establishing the feasibility of pre-hoc performance prediction rather than demonstrating the exact degree of improvement this would offer in deployment; evaluating complete routing systems that leverage these predictions to optimize cost-quality trade-offs in practice remains future work.

Our evaluation is limited to single-turn interactions, which may not generalize to diverse evaluation frameworks or multi-turn conversations.
It is entirely possible for later queries in a conversation to demand a much more capable model to answer properly, requiring more thought about routing using the kinds of predictions discussed here.
Future work will look at the applicability of the paradigms presented here in multi-turn interactions.

Another unrealized opportunity of this work is that LLM/agentic judges can integrate arbitrary alignment requirements into their evaluations.
Here, we focused on a relatively generic evaluation rubric with our judge.
While we did do some preliminary analysis here (see \cref{app:mischievous_rubric_results}), additional work would be needed to demonstrate how things like alignment can be predicted with the PA/RPRA paradigms.

Another key limitation of this work is that the report card approach requires processing many tokens to generate a prediction.
While models are typically orders of magnitude faster at processing input tokens than generating output tokens, this still introduces additional overhead.
Likewise, the fine-tuning approach necessitates either maintaining a second set of weights or sacrificing model performance.
Future work will look at integrating the predictions directly into the models so that the same prompt can be used for both prediction and response generation.
In addition, future work will also explore alternative training strategies, such as reinforcement learning from judge feedback or leveraging human-in-the-loop corrections.

\section*{Ethics Statement}

This paper presents work that enables more efficient deployment of large language models.
By enabling smaller models to recognize their own limitations, the PA and RPRA paradigms advocated for here could reduce computational costs and energy consumption associated with LLM inference.
We see no obvious negative societal consequences of this work requiring special attention.

\section*{Reproducibility Details}\label{app:reproducibility_details}

All experiments were performed on one machine with eight (8) NVIDIA H100 chips.
Replicating all the experiments run in the paper requires approximately forty-eight (48) hours on such a machine.
All of the datasets and models were taken from HuggingFace directly.
Existing splits in the datasets were merged, and then a 75/25 split was made (using a static seed) to produce a training and testing set (hence the ranges of values in the value of $n$ in several plots).
The distribution of judge scores and zero-shot results in the paper refer to the training set.
The contextual and fine-tuning results in the paper refer to the testing set.
The full source code to replicate our experiments will be made available shortly.

\section*{Acknowledgments}

The authors would like to thank the Meta Reality Labs Core AI ASL team, without whom this work would not have been possible.

\bibliography{dylan}

\bibliographystyle{iclr2026_conference}

\clearpage
\appendix

\onecolumn

\titlecontents{section}[0em]{}{\thecontentslabel\enspace}{}{\titlerule*[0.5pc]{.}\contentspage}

\section*{Appendix Table of Contents}
\startcontents[appendices]
\printcontents[appendices]{}{1}{\setcounter{tocdepth}{1}}

\clearpage

\section{Supplementary Related Works}\label{app:less_related_works}

Beyond the core areas of LLM evaluation and performance prediction discussed in the main text, several adjacent research directions provide important context for our work.

\textbf{Prompt Engineering and Optimization.}\enspace
Substantial research focuses on optimizing LLM interactions through sophisticated prompt engineering techniques.
Such prompt engineering strategies date back to the work on Learning to Think \citep{schmidhuber2015learning} (later refined~\citep{schmidhuber2018one}), wherein one network learns how to query and extract information from another network.
Evolutionary algorithms offer a gradient-free approach for prompt optimization in black-box scenarios by evolving effective instructions and demonstrations \citep{guo2024connecting,fernando2024promptbreeder}.

\textbf{Benchmarks for LLM and Agent Evaluation.}\enspace
The rapid advancement of LLMs and agentic systems has necessitated continuous development of comprehensive benchmarks.
These span from basic code generation (HumanEval \citep{chen2021evaluating}, MBPP \citep{austin2021program}) to complex software engineering tasks (SWE-Bench \citep{jimenez2024swe}, DevBench \citep{li2024devbench}).
For machine learning workflows specifically, benchmarks like DevAI \citep{zhuge2024agent}, ML-Dev-Bench \citep{padigela2025ml}, and ML-BENCH evaluate end-to-end capabilities including environment setup and API integration, posing distinct challenges for current models \citep{tang2023ml}.
Broader agent evaluation frameworks such as AgentBench \citep{liu2023agentbench} further assess LLM-as-Agent capabilities across diverse interactive environments.

\textbf{Retrieval-Augmented Generation (RAG) Systems.}\enspace
RAG systems are extensively developed to enhance LLMs' factual accuracy and contextual understanding, incorporating advanced strategies for chunking, metadata enrichment, and confidence-based retrieval triggering \citep{lewis2020retrieval}.
Recent advancements include adaptive retrieval mechanisms \citep{jeong2024adaptive} and multi-hop reasoning frameworks \citep{liu2025hoprag} that improve information integration.

\textbf{Model Efficiency and Optimization.}\enspace
Parallel research focuses on optimizing LLM inference through techniques such as speculative decoding \citep{leviathan2023fast} and quantization methods \citep{frantar2022gptq}.
While these approaches target computational efficiency during generation, our work addresses efficiency at the evaluation level by predicting scores without requiring full response generation.

\clearpage

\section{Summary Table for Contextual Results}\label{app:summary_table_for_contextual_results}

\cref{tab:contextual_summary} summarizes the improvement in prediction accuracy for the contextual setting as compared with the zero-shot setting.
For a summary of the fine-tuned model results, see \cref{tab:fine_tuning_summary}.

\begin{table}[h!]
    \centering
    \small
    \begin{tabular}{lcccccc}
        \textbf{Model} & \textbf{MedQA} & \textbf{LongFact} & \textbf{AIME 2024} & \textbf{MMLU-Pro} & \textbf{SciCode} & \textit{Mean} \\
        \hline \vspace{-0.75em}\\
        M09B & -0.10 & +0.27 & -0.15 & -0.11 & -0.43 & -0.10 \\
        L321B & +0.28 & +0.12 & +0.91 & +0.55 & +0.90 & \textbf{+0.55} \\
        L323B & -0.01 & -0.12 & +0.35 & +0.24 & -0.12 & +0.07 \\
        L318B & +0.22 & -0.05 & -0.22 & -0.00 & +0.03 & -0.00 \\
        DSQ14B & +0.14 & +0.07 & -0.17 & +0.07 & -0.02 & +0.02 \\
        GPT20B & +0.00 & -0.13 & +0.16 & -0.01 & -0.08 & -0.01 \\
        DSQ32B & +0.03 & +0.09 & +0.75 & +0.02 & -0.13 & +0.15 \\
        L3370B & +0.59 & +0.67 & -0.03 & +0.24 & +0.38 & +0.37 \\
        DSL70B & +0.14 & +0.21 & +0.53 & +0.10 & -0.22 & +0.15 \\
        L416E & +0.06 & +0.08 & +0.01 & +0.13 & +0.28 & +0.11 \\
        GPT120B & +0.04 & +0.18 & +0.05 & -0.05 & +0.18 & +0.08 \\
    \end{tabular}
    \caption{Summary of the improvement in the prediction accuracy for the contextual setting as compared with the zero-shot setting.}
    \label{tab:contextual_summary}
\end{table}

\clearpage

\section{Prompts and Prompt Templates}\label{app:prompt_templates}

\cref{ppt:responder_system_prompt,ppt:evaluator_system_prompt,ppt:zero_shot_score_prediction_template,ppt:contextual_score_prediction_template,ppt:long_feedback_template,ppt:short_feedback_template,ppt:contextual_other_model_score_prediction_template,ppt:score_prediction_correction_prompt,ppt:combined_evaluation_template,ppt:independent_evaluation_template,ppt:rubric,ppt:json_correction_template} shows the prompts used for different parts of the system.
The contents enclosed in curly brackets are substituted by the relevant content while the experiment is running.

\begin{promptbox}[hp]
    \promptfile{assets/prompt_templates/responder_system_prompt.txt}
    \caption{The system prompt used by the models when asked to respond to a query or estimate how the LLM judge will score their response.}
    \label{ppt:responder_system_prompt}
\end{promptbox}

\begin{promptbox}[hp]
    \promptfile{assets/prompt_templates/evaluator_system_prompt.txt}
    \caption{The system prompt used by the LLM judge.}
    \label{ppt:evaluator_system_prompt}
\end{promptbox}

\begin{promptbox}[hp]
    \promptfile{assets/prompt_templates/zero_shot_score_prediction_template.txt}
    \caption{The prompt used to ask a model to predict the LLM judge's score for its future response zero-shot.}
    \label{ppt:zero_shot_score_prediction_template}
\end{promptbox}

\begin{promptbox}[hp]
    \promptfile{assets/prompt_templates/contextual_score_prediction_template.txt}
    \caption{The prompt used to ask a model to predict the LLM judge's score for its future response based on a report card for the model.}
    \label{ppt:contextual_score_prediction_template}
\end{promptbox}

\begin{promptbox}[hp]
    \promptfile{assets/prompt_templates/long_feedback_template.txt}
    \caption{The sub-prompt used to encode the report card for a model.}
    \label{ppt:long_feedback_template}
\end{promptbox}

\begin{promptbox}[hp]
    \promptfile{assets/prompt_templates/short_feedback_template.txt}
    \caption{A short variant of the sub-prompt in \cref{ppt:long_feedback_template} used to encode the report card for a model. See Appendix~\ref{app:short_feedback_template_results} for the ablation results using this prompt.}
    \label{ppt:short_feedback_template}
\end{promptbox}

\begin{promptbox}[hp]
    \promptfile{assets/prompt_templates/contextual_other_model_score_prediction_template.txt}
    \caption{The prompt used to ask a model to predict the LLM judge's score for an arbitrary model's future response based on a report card for the model.}
    \label{ppt:contextual_other_model_score_prediction_template}
\end{promptbox}

\begin{promptbox}[hp]
    \promptfile{assets/prompt_templates/score_prediction_correction_prompt.txt}
    \caption{The prompt used to request a model to correct its prediction if the parser was unable to read it.}
    \label{ppt:score_prediction_correction_prompt}
\end{promptbox}

\begin{promptbox}[hp]
    \promptfile{assets/prompt_templates/combined_evaluation_template.txt}
    \caption{The prompt used when asking the LLM judge to simultaneously grade the responses of all the models to a single query.}
    \label{ppt:combined_evaluation_template}
\end{promptbox}

\begin{promptbox}[hp]
    \promptfile{assets/prompt_templates/independent_evaluation_template.txt}
    \caption{A variant of the prompt in \cref{ppt:combined_evaluation_template} used when asking the LLM judge to grade the responses of a single model to a single query. See Appendix~\ref{app:independent_evaluation_results} for the ablation results using this prompt.}
    \label{ppt:independent_evaluation_template}
\end{promptbox}

\begin{promptbox}[hp]
    \promptfile{assets/prompt_templates/rubric.txt}
    \caption{The rubric used by the LLM judge to grade responses.}
    \label{ppt:rubric}
\end{promptbox}

\begin{promptbox}[hp]
    \promptfile{assets/prompt_templates/mischievous_rubric.txt}
    \caption{The mischievous rubric used by the LLM judge to grade responses for \cref{app:mischievous_rubric_results}.}
    \label{ppt:mischievous_rubric}
\end{promptbox}

\begin{promptbox}[hp]
    \promptfile{assets/prompt_templates/json_correction_template.txt}
    \caption{The prompt used to request the LLM judge to correct its evaluation if the parser was unable to read it.}
    \label{ppt:json_correction_template}
\end{promptbox}

\clearpage

\section{Mode Scores}\label{app:mode_scores}

\cref{tab:mode_scores} shows the mode scores for each of the models on each of the datasets.
Likewise, \cref{tab:mode_scores_by_category_mmlu_pro} shows the mode scores for each of the models on each of the categories in the MMLU-Pro dataset.
As would be expected from looking at \cref{fig:scoreboard_med_qa,fig:scoreboard_longfact,fig:scoreboard_aime_2024,fig:scoreboard_scicode,fig:scoreboard_mmlu_pro}, a good variation is observed between the different models and datasets, implying that the results presented here generalize well.

\begin{table}[hp]
    \centering
    \begin{tabular}{lccccc}
        Model & AIME 2024 & LongFact & MedQA & SciCode & MMLU-Pro \\
        \hline \vspace{-0.75em}\\
        M09B & \modebad & \modebad & \modebad & \modebad & \modebad \\
        L321B & \modebad & \modeok & \modebad & \modebad & \modebad \\
        L323B & \modebad & \modegreat & \modegreat & \modebad & \modebad \\
        L318B & \modebad & \modeok & \modegreat & \modebad & \modegreat \\
        GPT20B & \modebad & \modegreat & \modegreat & \modegreat & \modegreat \\
        DSQ14B & \modegreat & \modegreat & \modeok & \modegreat & \modegreat \\
        DSQ32B & \modebad & \modegreat & \modegreat & \modeok & \modegreat \\
        GPT120B & \modebad & \modegreat & \modegreat & \modegreat & \modegreat \\
        DSL70B & \modebad & \modegreat & \modegreat & \modeok & \modegreat \\
        L3370B & \modegreat & \modegreat & \modegreat & \modeok & \modegreat \\
        L416E & \modegreat & \modegreat & \modegreat & \modegreat & \modegreat \\
    \end{tabular}
    \caption{The mode scores for each model on each of the datasets.}
    \label{tab:mode_scores}
\end{table}

\begin{table}[hp]
    \centering
    \setlength{\tabcolsep}{3pt}
    \begin{tabular}{lcccccc}
        Model & Biology & Business & Chemistry & Computer Science & Economics & Engineering \\
        \hline \vspace{-0.75em}\\
        M09B & \modebad & \modebad & \modebad & \modebad & \modebad & \modebad \\
        L321B & \modebad & \modebad & \modebad & \modebad & \modebad & \modebad \\
        L323B & \modegreat & \modebad & \modebad & \modebad & \modebad & \modebad \\
        L318B & \modegreat & \modebad & \modebad & \modebad & \modegreat & \modebad \\
        GPT20B & \modegreat & \modegreat & \modegreat & \modegreat & \modegreat & \modebad \\
        DSQ14B & \modegreat & \modegreat & \modegreat & \modegreat & \modegreat & \modegreat \\
        DSQ32B & \modegreat & \modegreat & \modebad & \modegreat & \modegreat & \modebad \\
        GPT120B & \modegreat & \modegreat & \modegreat & \modegreat & \modegreat & \modegreat \\
        DSL70B & \modegreat & \modegreat & \modegreat & \modegreat & \modegreat & \modegreat \\
        L3370B & \modegreat & \modegreat & \modegreat & \modegreat & \modegreat & \modeok \\
        L416E & \modegreat & \modegreat & \modegreat & \modegreat & \modegreat & \modegreat \\
    \end{tabular}
    \\
    \vspace{1em}
    \setlength{\tabcolsep}{4pt}
    \begin{tabular}{lcccccccc}
        Graph & Health & History & Law & Math & Other & Philosophy & Physics & Psychology \\
        \hline \vspace{-0.75em}\\
        M09B & \modebad & \modebad & \modebad & \modebad & \modebad & \modebad & \modebad & \modebad \\
        L321B & \modebad & \modebad & \modebad & \modebad & \modebad & \modebad & \modebad & \modebad \\
        L323B & \modegreat & \modegreat & \modebad & \modebad & \modegreat & \modebad & \modebad & \modegreat \\
        L318B & \modegreat & \modegreat & \modegreat & \modebad & \modegreat & \modegreat & \modebad & \modegreat \\
        GPT20B & \modegreat & \modegreat & \modebad & \modegreat & \modegreat & \modegreat & \modegreat & \modegreat \\
        DSQ14B & \modegreat & \modegreat & \modeok & \modegreat & \modegreat & \modegreat & \modegreat & \modegreat \\
        DSQ32B & \modegreat & \modegreat & \modegreat & \modegreat & \modegreat & \modegreat & \modegreat & \modegreat \\
        GPT120B & \modegreat & \modegreat & \modegreat & \modegreat & \modegreat & \modegreat & \modegreat & \modegreat \\
        DSL70B & \modegreat & \modegreat & \modegreat & \modegreat & \modegreat & \modegreat & \modegreat & \modegreat \\
        L3370B & \modegreat & \modegreat & \modegreat & \modegreat & \modegreat & \modegreat & \modegreat & \modegreat \\
        L416E & \modegreat & \modegreat & \modegreat & \modegreat & \modegreat & \modegreat & \modegreat & \modegreat \\
    \end{tabular}
    \caption{The mode scores for each model on each of the categories in MMLU-Pro.}
    \label{tab:mode_scores_by_category_mmlu_pro}
\end{table}

\clearpage

\section{Independent Evaluation Results}\label{app:independent_evaluation_results}

\begin{figure}[hp]
    \centering
    \includegraphics[width=0.9\textwidth]{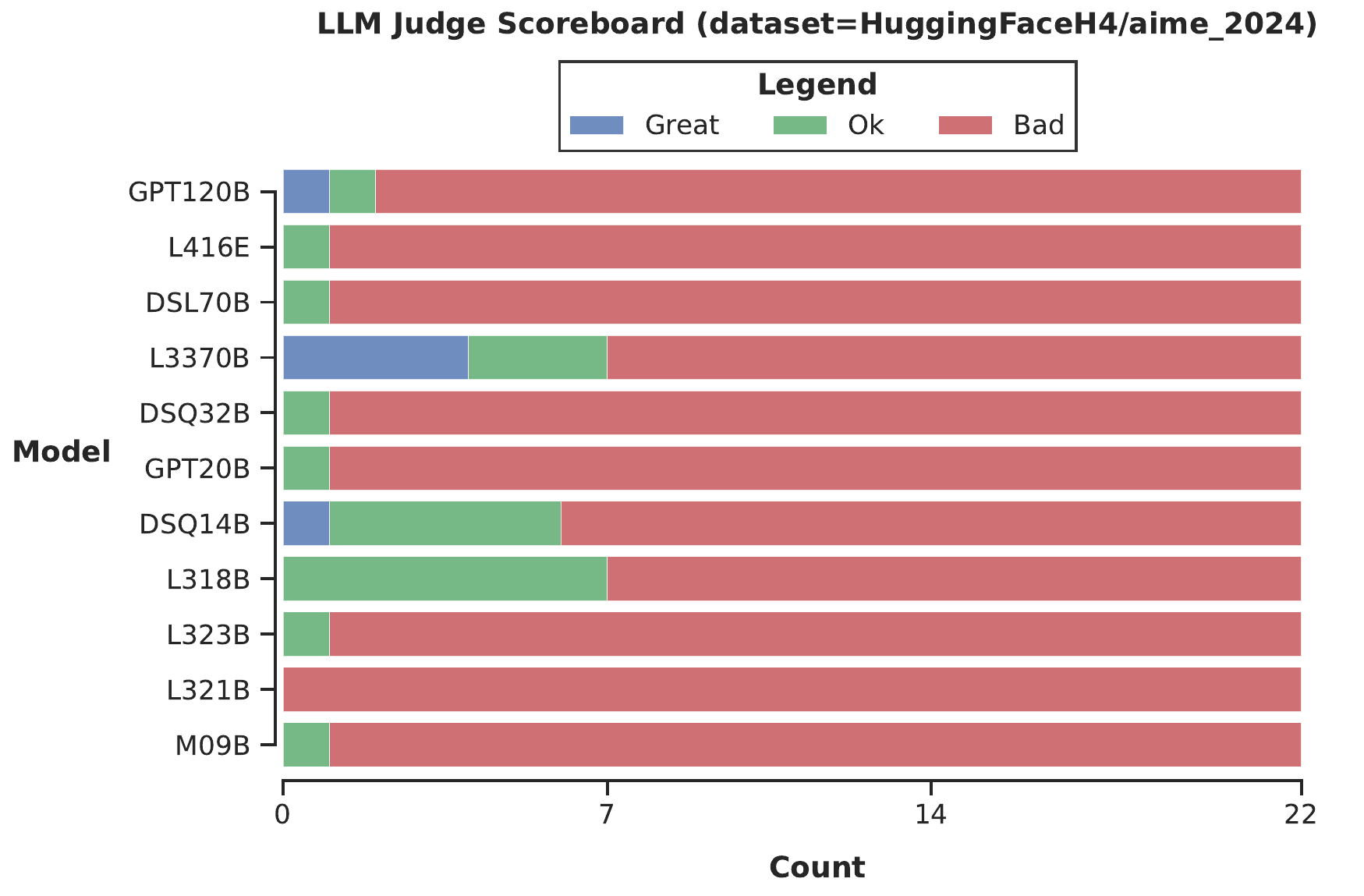}
    \caption{Distribution of LLM judge scores on AIME 2024 dataset under independent evaluation.}
    \label{fig:independent_evaluation_scoreboard_aime_2024}
\end{figure}

\begin{figure}[hp]
    \centering
    \includegraphics[width=0.9\textwidth]{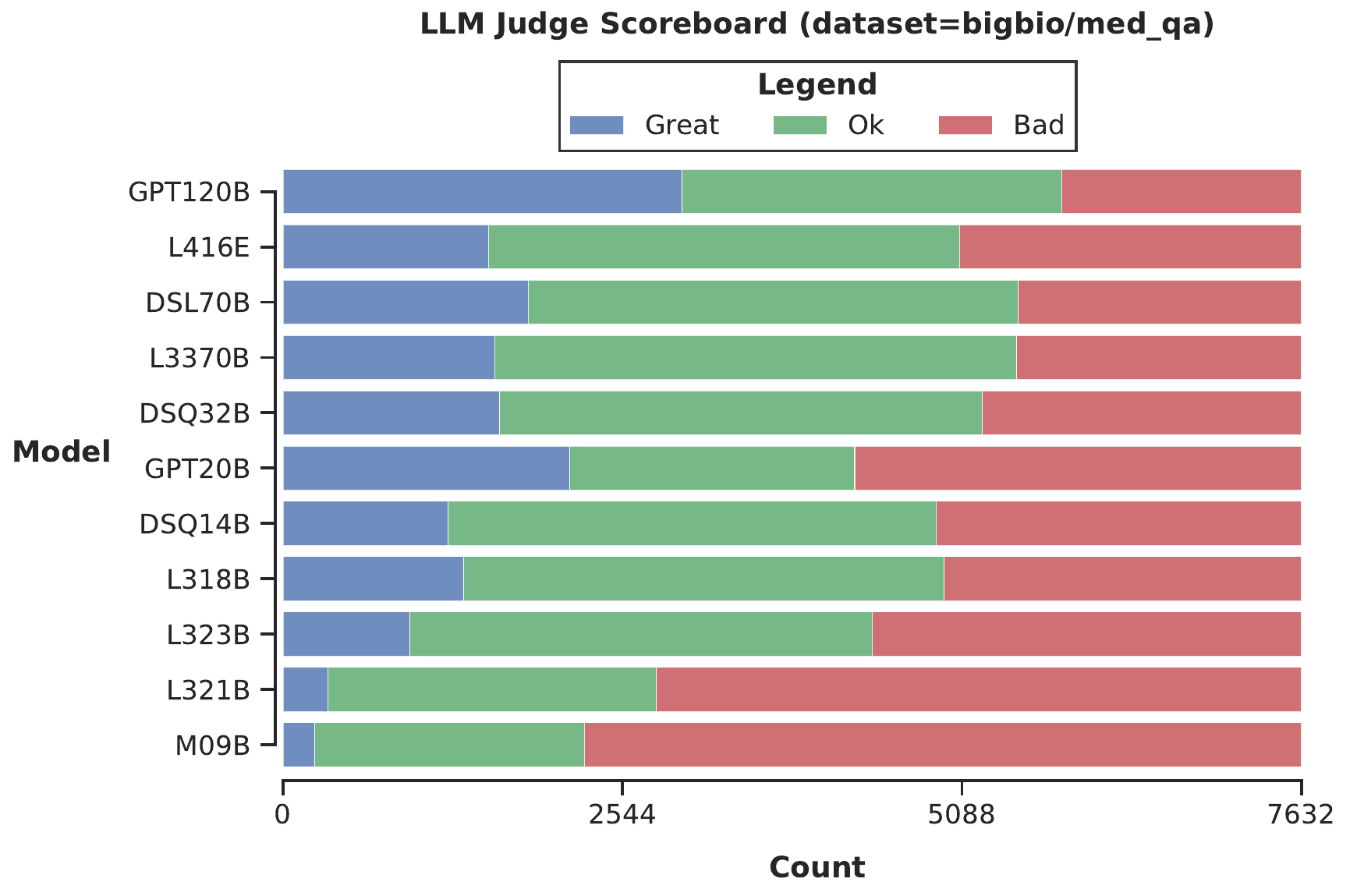}
    \caption{Distribution of LLM judge scores on MedQA dataset under independent evaluation.}
    \label{fig:independent_evaluation_scoreboard_med_qa}
\end{figure}

\begin{figure}[hp]
    \centering
    \includegraphics[width=0.9\textwidth]{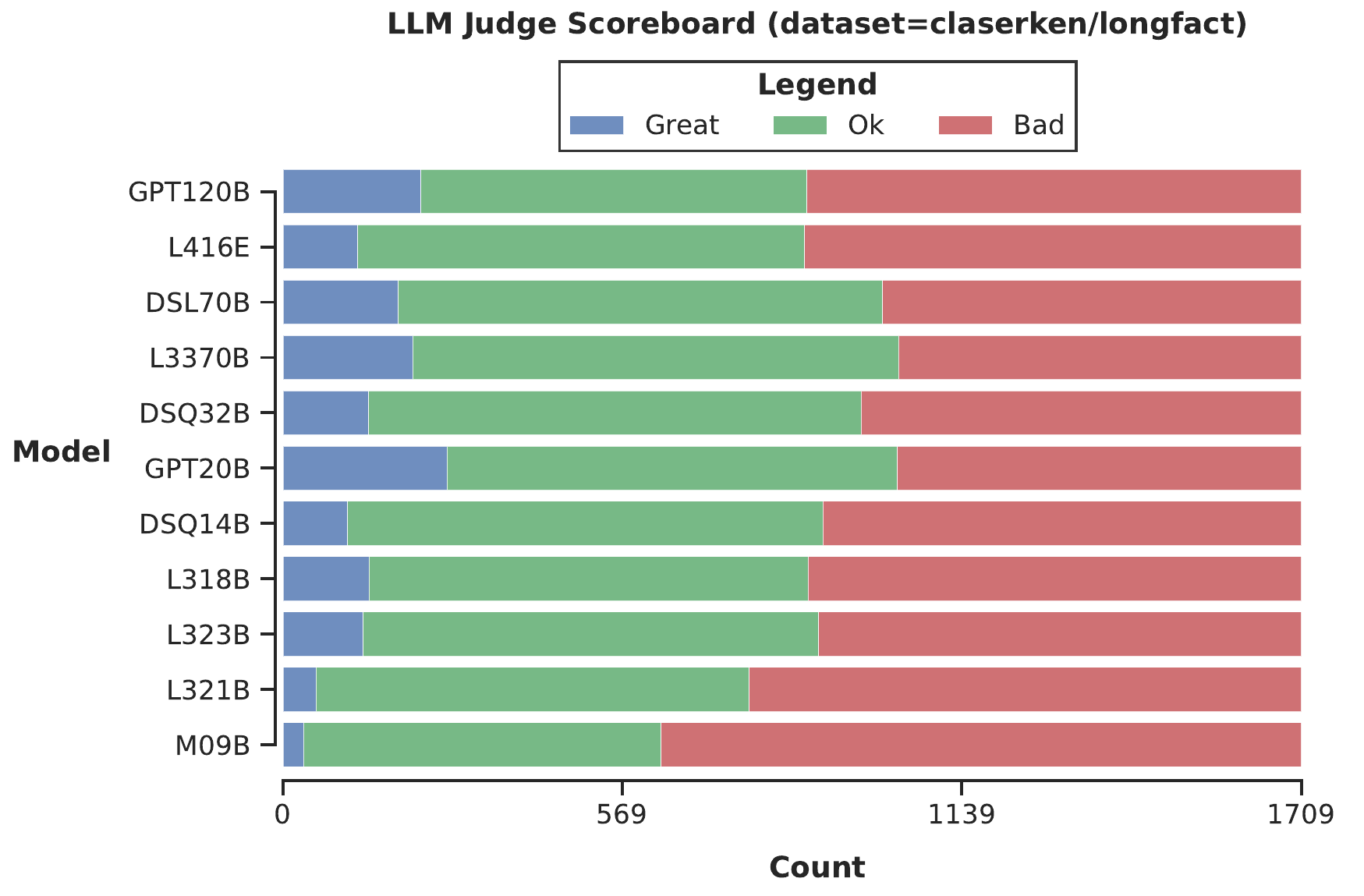}
    \caption{Distribution of LLM judge scores on LongFact dataset under independent evaluation.}
    \label{fig:independent_evaluation_scoreboard_longfact}
\end{figure}

\begin{figure}[hp]
    \centering
    \includegraphics[width=0.9\textwidth]{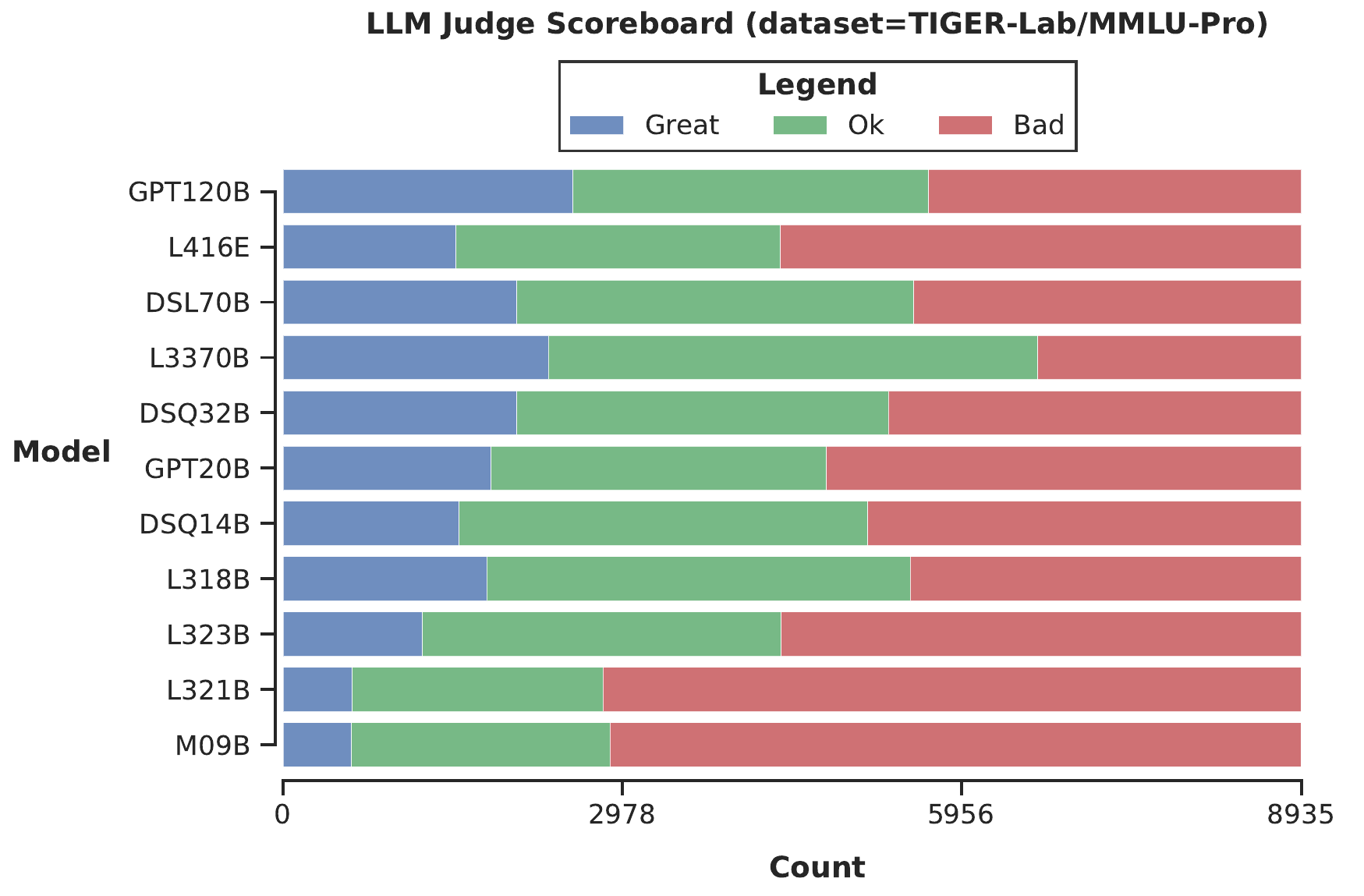}
    \caption{Distribution of LLM judge scores on MMLU-Pro dataset under independent evaluation.}
    \label{fig:independent_evaluation_scoreboard_mmlu_pro}
\end{figure}

\begin{figure}[hp]
    \centering
    \includegraphics[width=0.9\textwidth]{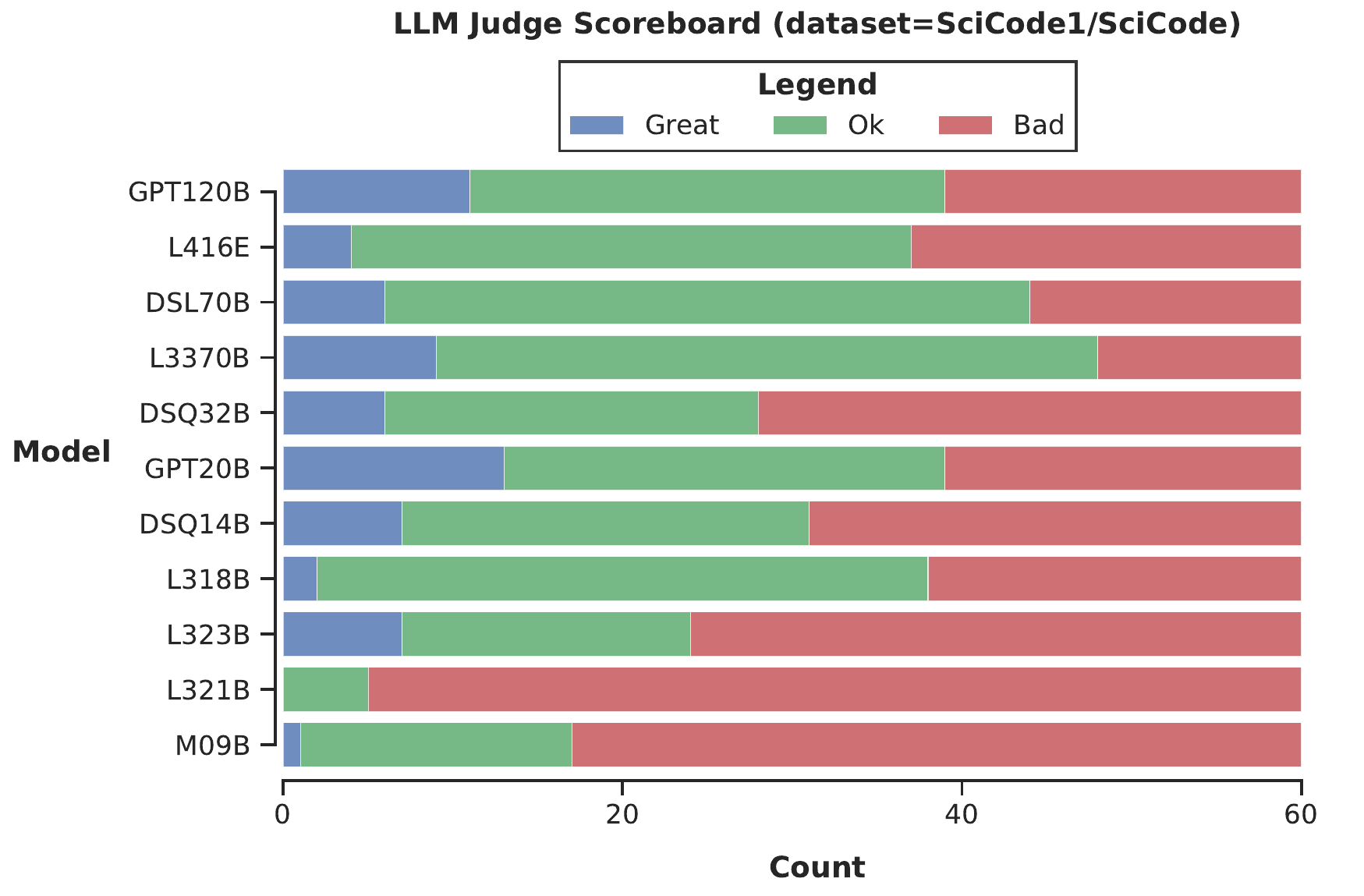}
    \caption{Distribution of LLM judge scores on SciCode dataset under independent evaluation.}
    \label{fig:independent_evaluation_scoreboard_scicode}
\end{figure}

LLMs are known to be overconfident~\citep{huang2025confqa,ren2023self}.
Thus, a relative assessment of model outputs is likely to have a higher variation than an independent assessment.
To verify this, we conducted an ablation in which the model outputs are assessed separately.
The results of this are shown in \cref{fig:independent_evaluation_scoreboard_aime_2024,fig:independent_evaluation_scoreboard_med_qa,fig:independent_evaluation_scoreboard_longfact,fig:independent_evaluation_scoreboard_mmlu_pro,fig:independent_evaluation_scoreboard_scicode}.
Intuitively, the combination of this and a mixture of small and large models renders a relative assessment more meaningful here.

\clearpage

\section{Prediction Distributions}\label{app:prediction_distributions}

\begin{figure}[hp]
    \centering
    \includegraphics[width=.9\textwidth]{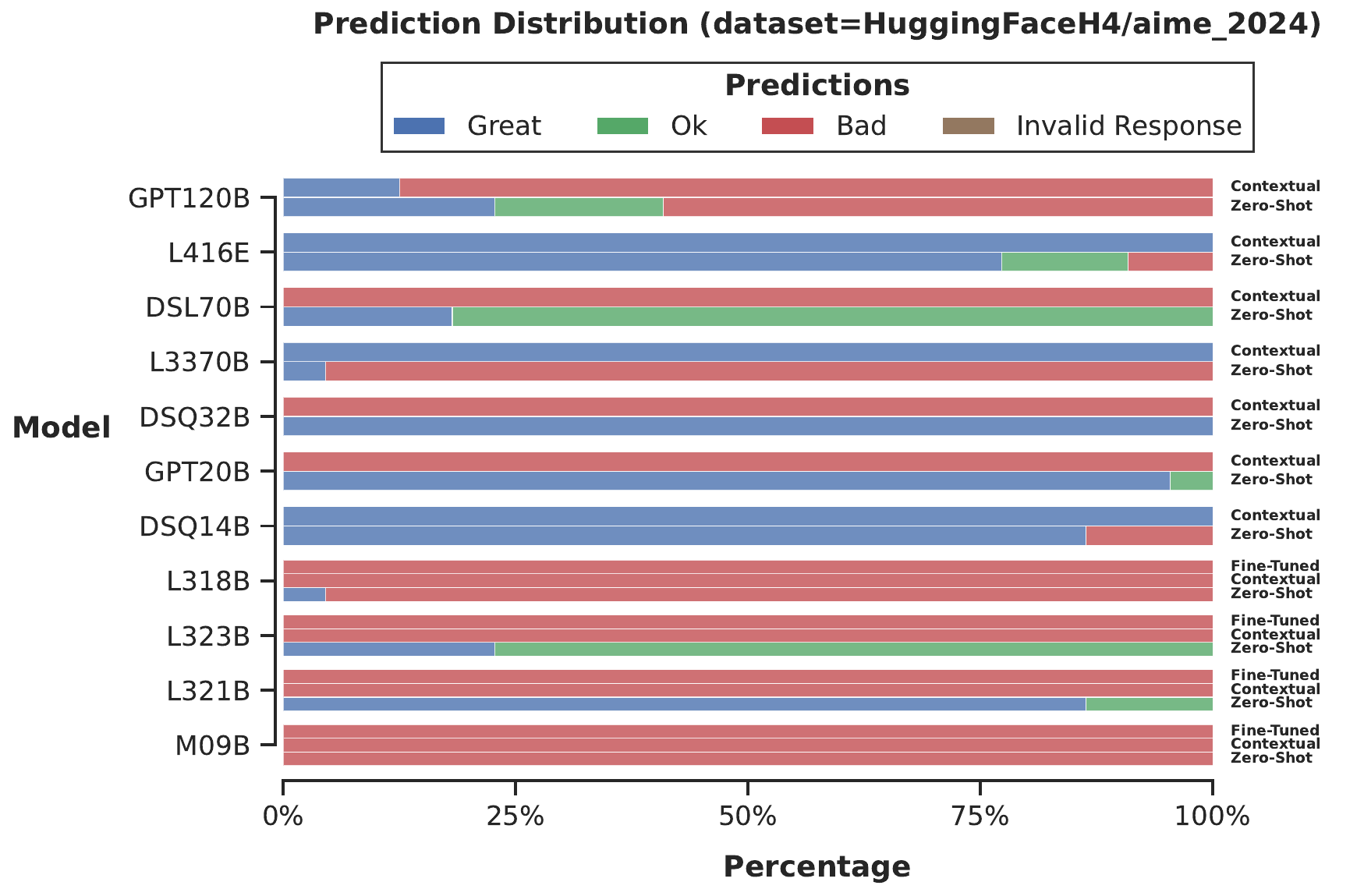}
    \caption{Zero-Shot and Contextual Prediction Distributions on AIME 2024}
    \label{fig:zero_shot_and_contextual_predictions_aime_2024}
\end{figure}

\begin{figure}[hp]
    \centering
    \includegraphics[width=.9\textwidth]{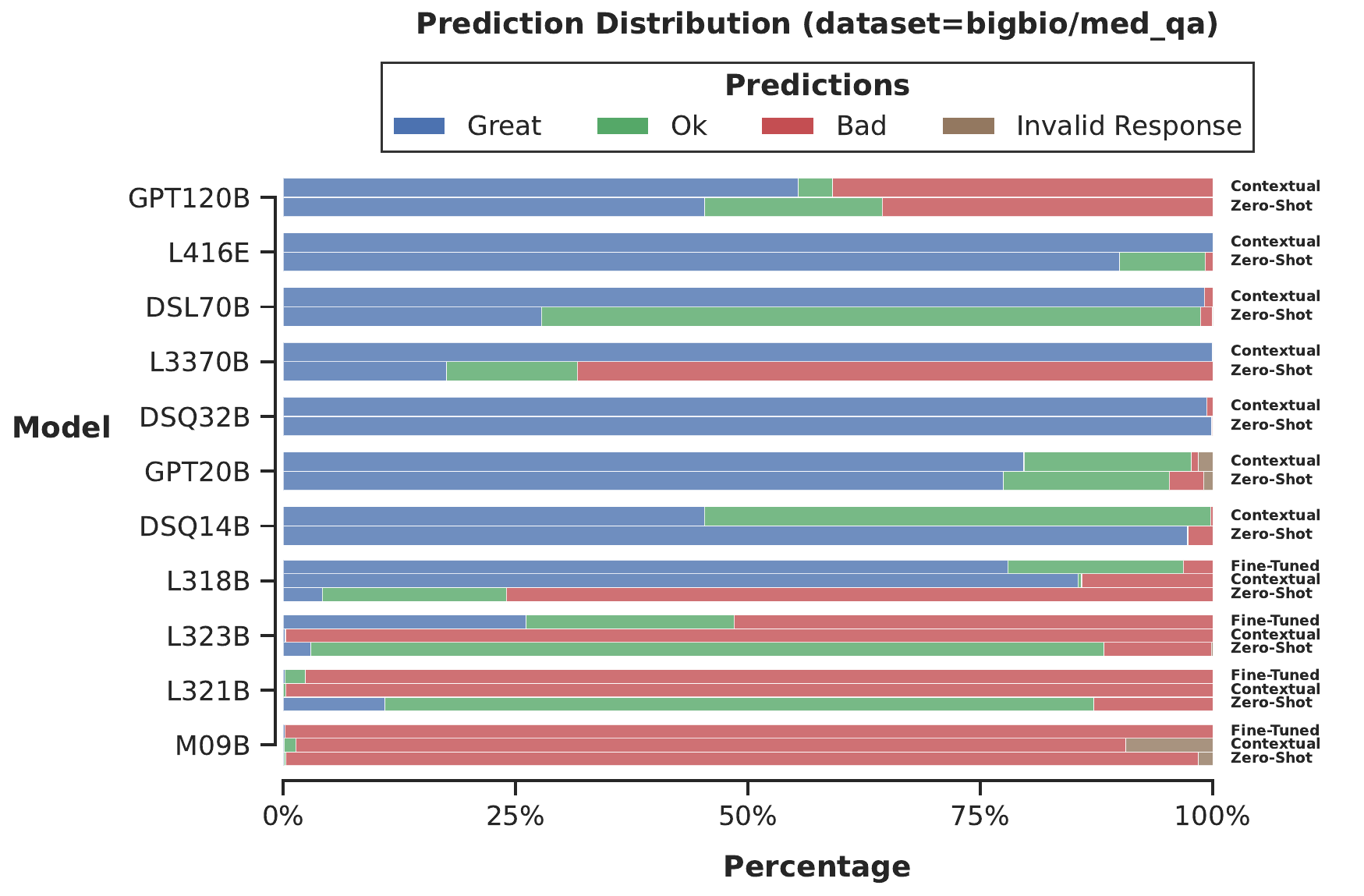}
    \caption{Zero-Shot and Contextual Prediction Distributions on MedQA}
    \label{fig:zero_shot_and_contextual_predictions_med_qa}
\end{figure}

\begin{figure}[hp]
    \centering
    \includegraphics[width=.9\textwidth]{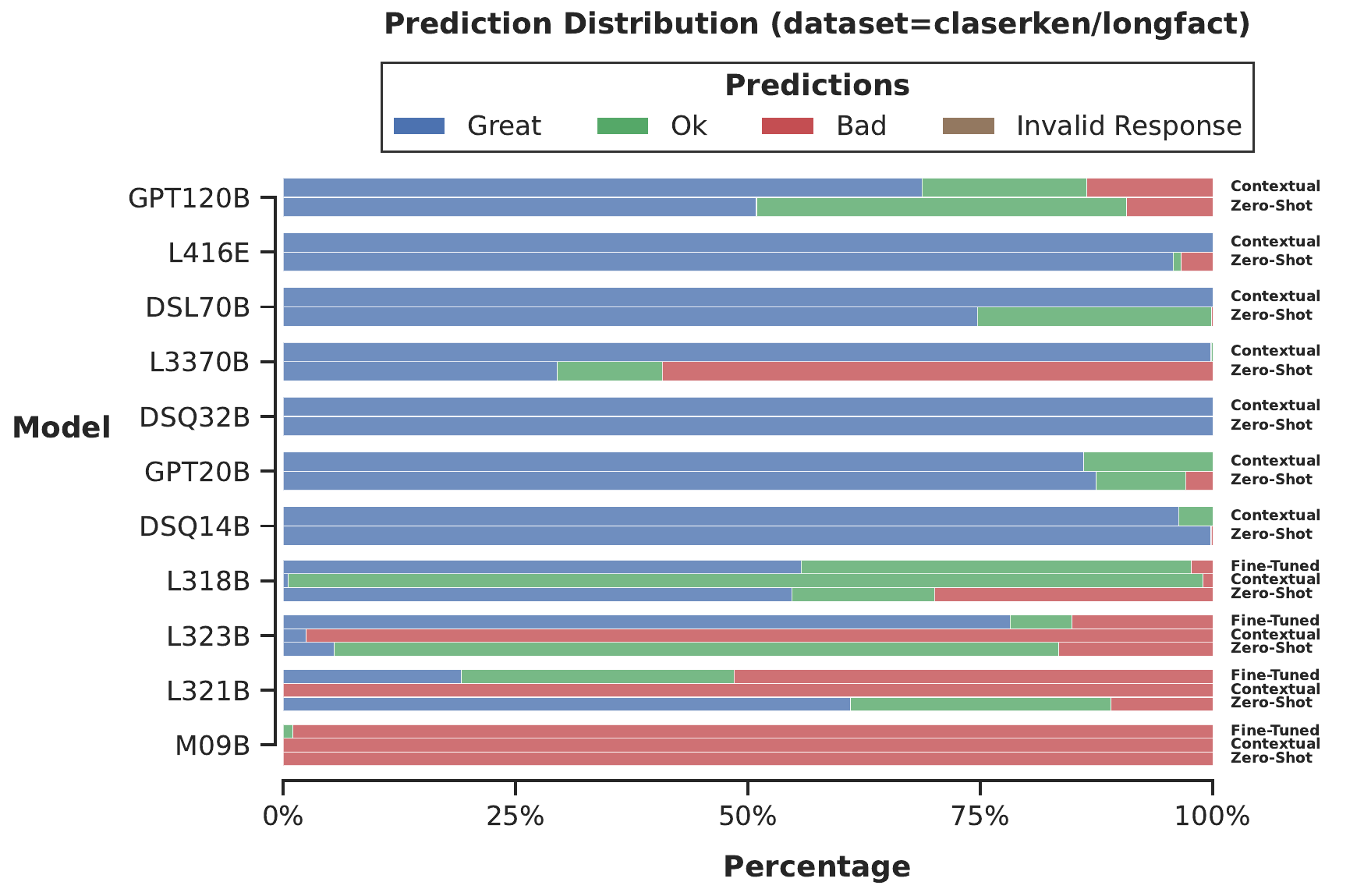}
    \caption{Zero-Shot and Contextual Prediction Distributions on LongFact}
    \label{fig:zero_shot_and_contextual_predictions_longfact}
\end{figure}

\begin{figure}[hp]
    \centering
    \includegraphics[width=.9\textwidth]{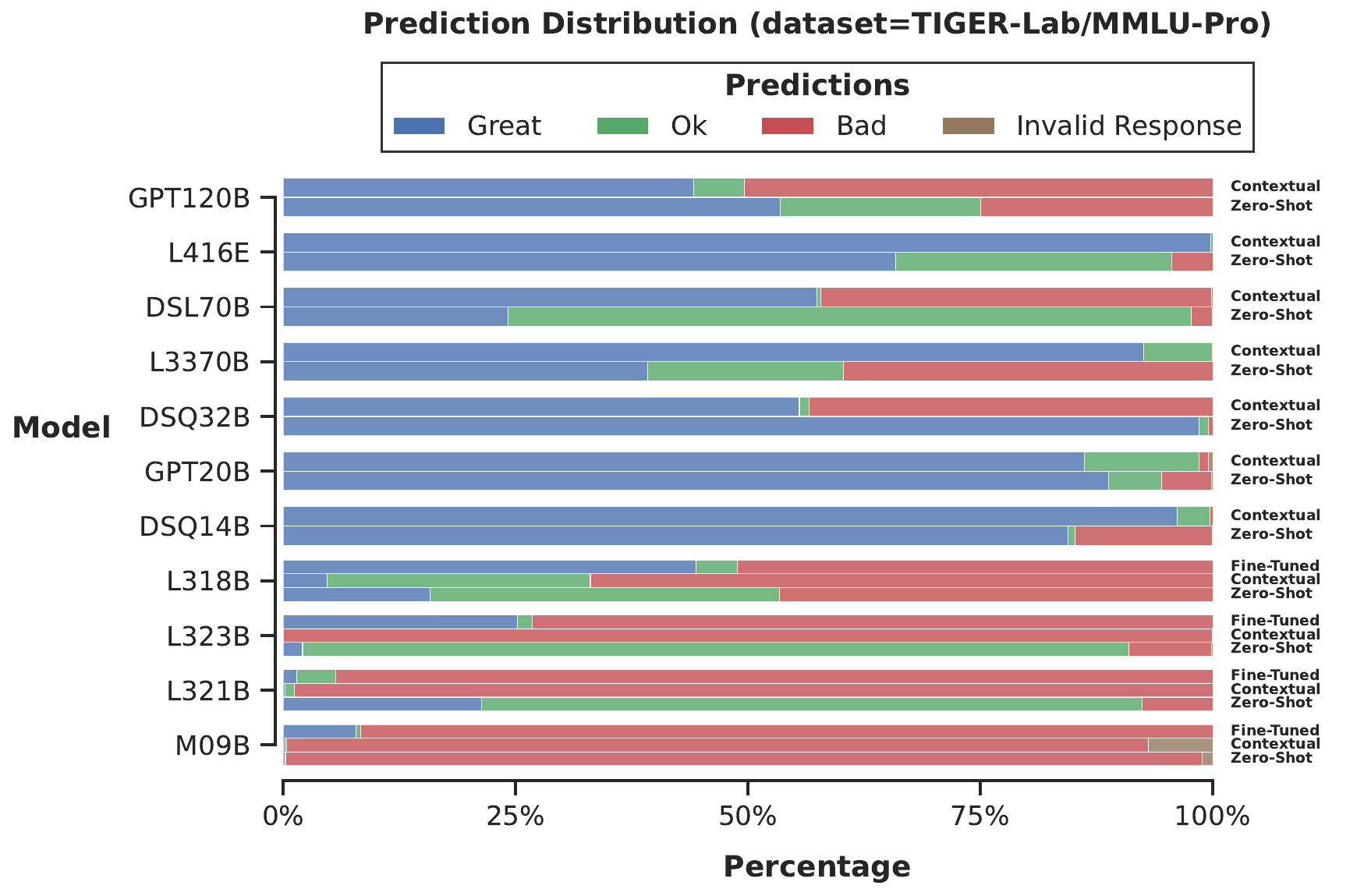}
    \caption{Zero-Shot and Contextual Prediction Distributions on MMLU-Pro}
    \label{fig:zero_shot_and_contextual_predictions_mmlu_pro}
\end{figure}

\begin{figure}[hp]
    \centering
    \includegraphics[width=.9\textwidth]{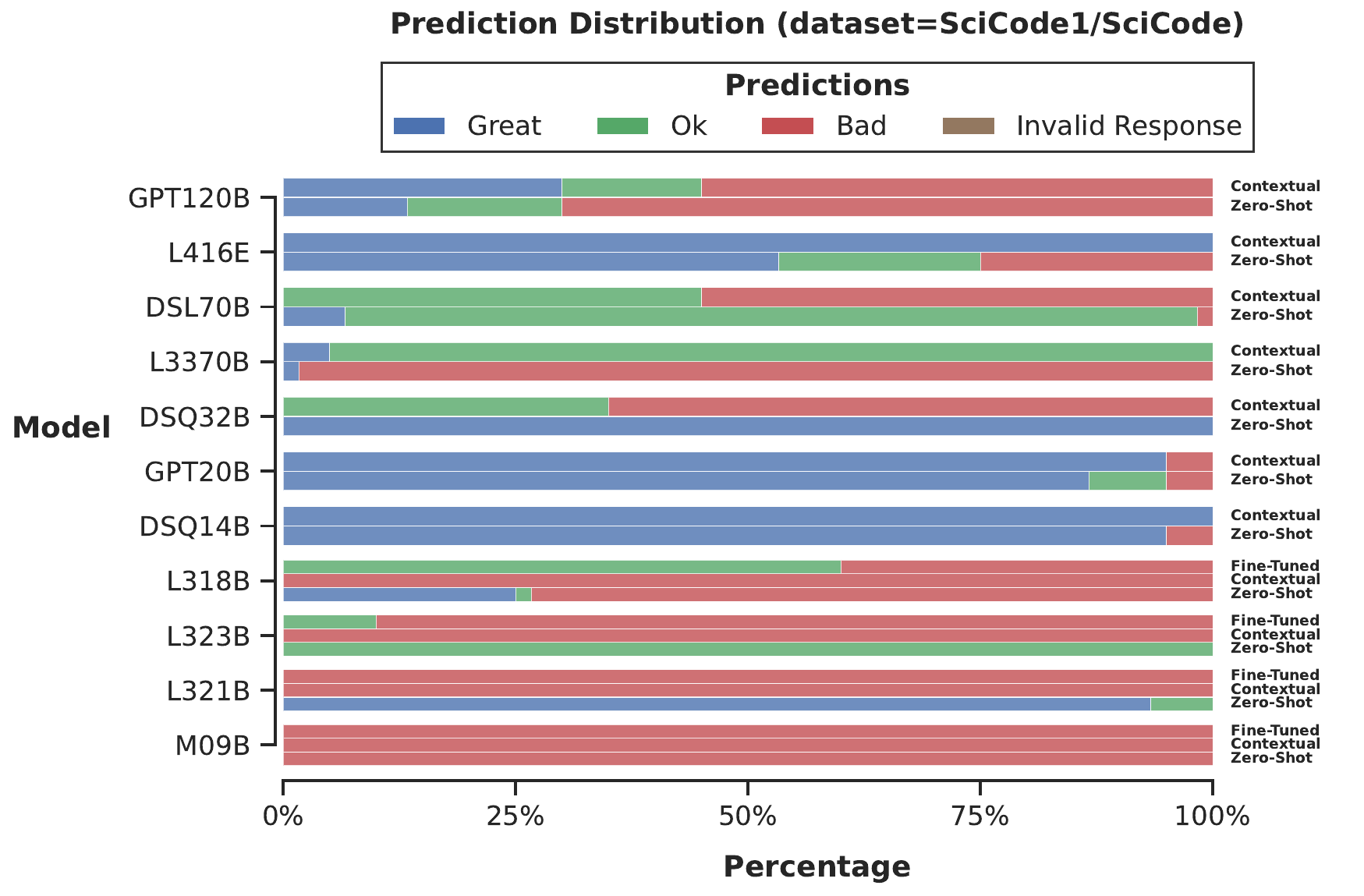}
    \caption{Zero-Shot and Contextual Prediction Distributions on SciCode}
    \label{fig:zero_shot_and_contextual_predictions_scicode}
\end{figure}

\cref{fig:zero_shot_and_contextual_predictions_aime_2024,fig:zero_shot_and_contextual_predictions_med_qa,fig:zero_shot_and_contextual_predictions_longfact,fig:zero_shot_and_contextual_predictions_mmlu_pro,fig:zero_shot_and_contextual_predictions_scicode} show the zero-shot and contextual prediction distributions for each of the models on each of the datasets.
As expected, the distributions tend to have lower variance than the actual score distribution (as queries are being categorized coarsely).

\clearpage

\section{Alternate Judge Results}\label{app:alternate_judge_results}

The use of Llama 3.3 70B as a judge was done as it's a well-studied and reliable large model.
As such, the evaluations it gives to the output of different models can be safely assumed to be meaningful.
This is sufficient for the goals of this work.
However, to validate that this continues to hold if the judge model is trained, we repeated the experiments with GPT OSS 120B as the judge model, replacing it as a model to be evaluated by MobileLLM-R1~\citep{zhao2025mobilellm} in order to keep the number of models being evaluated constant.
The results are shown in Figures 1 through 10.
While these results bear some similarities to the results in the main text, several differences are present.
Most notably, GPT OSS 120B is a harsher judge than Llama 3.3 70B.
Despite this, the report card strategy is still able to generally improve the predictive performance of the system.

We noted that when running these experiments, our experiences suggest that GPT OSS 120B is much less suitable as a judge, being prone to (1)~low-variety blanket evaluations of responses, and (2)~producing corrupted outputs where arbitrary thinking tokens not included in the specification for GPT OSS 120B are present.
These observations are consistent with the fact that this task is most certainly outside the training regime of these models, and the trend of the newest generation of models (from around the time of Llama 4 onwards) is that they are heavily overfitted to benchmarks.
A full analysis of this behaviour is outside the scope of this work.

\begin{figure}[hp]
    \centering
    \includegraphics[width=0.9\textwidth]{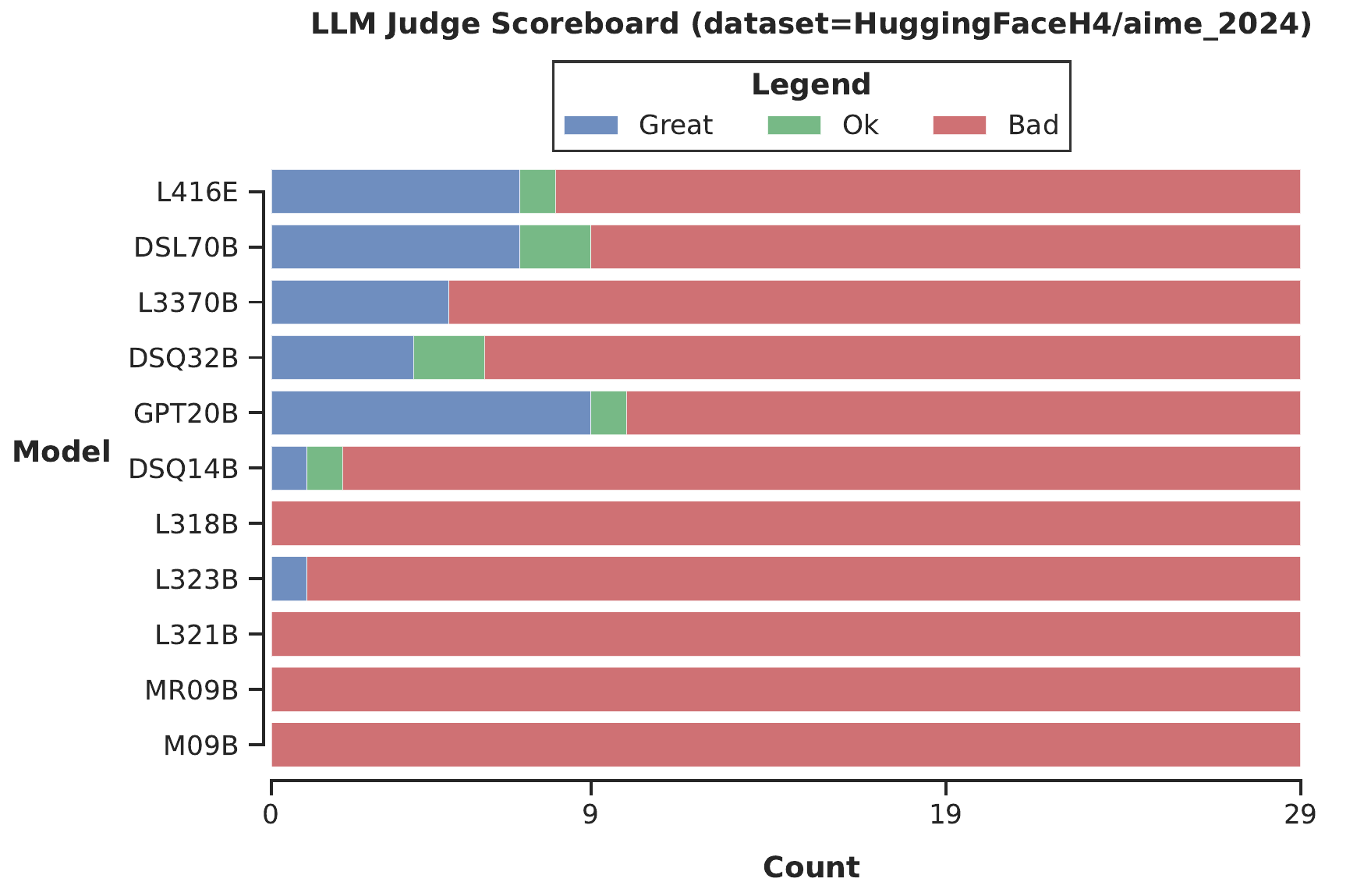}
    \caption{Distribution of LLM judge scores on AIME 2024 dataset with GPT OSS 120B as the judge.}
    \label{fig:alternate_judge_scoreboard_aime_2024}
\end{figure}

\begin{figure}[hp]
    \centering
    \includegraphics[width=0.9\textwidth]{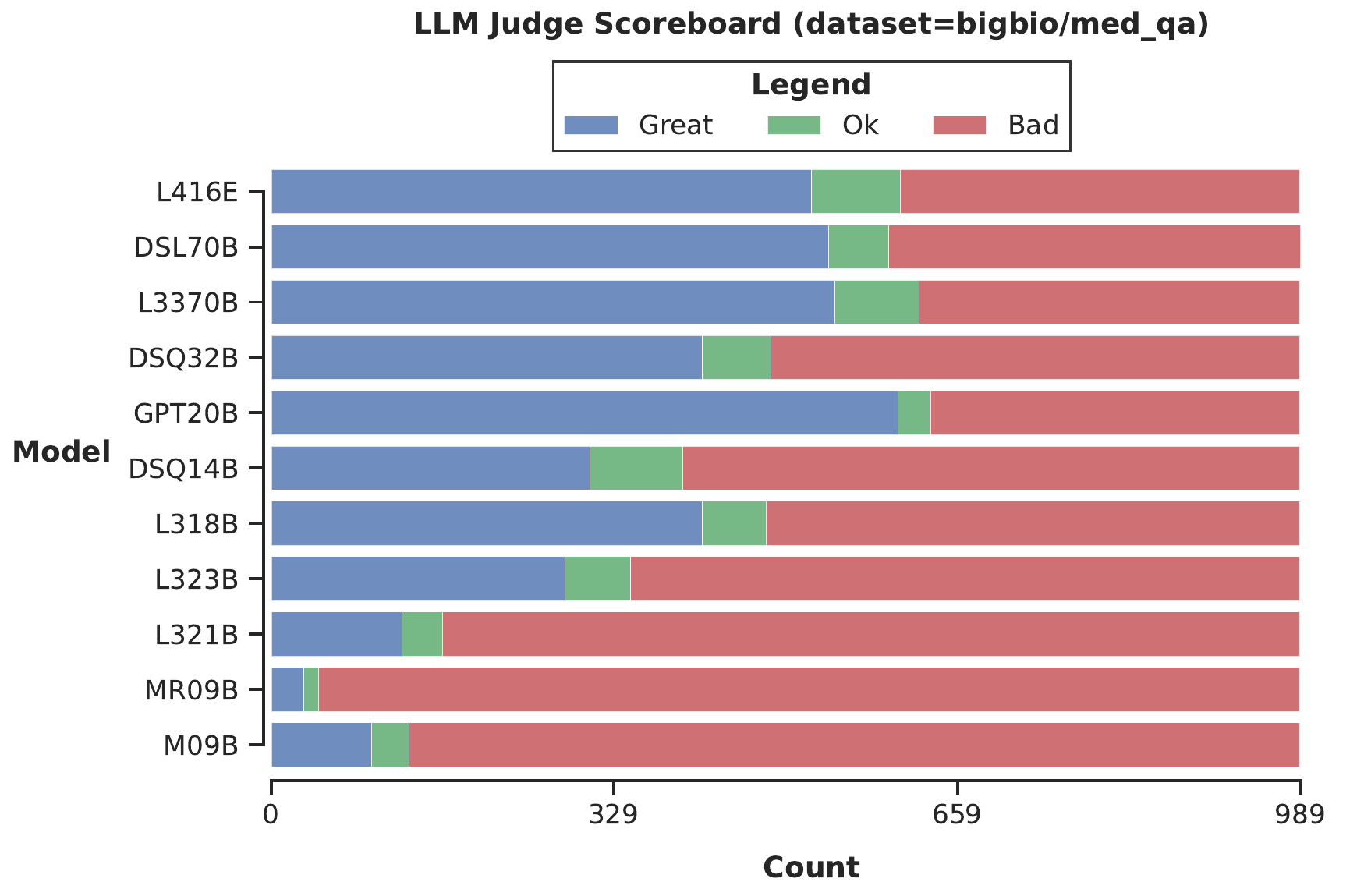}
    \caption{Distribution of LLM judge scores on MedQA dataset with GPT OSS 120B as the judge.}
    \label{fig:alternate_judge_scoreboard_med_qa}
\end{figure}

\begin{figure}[hp]
    \centering
    \includegraphics[width=0.9\textwidth]{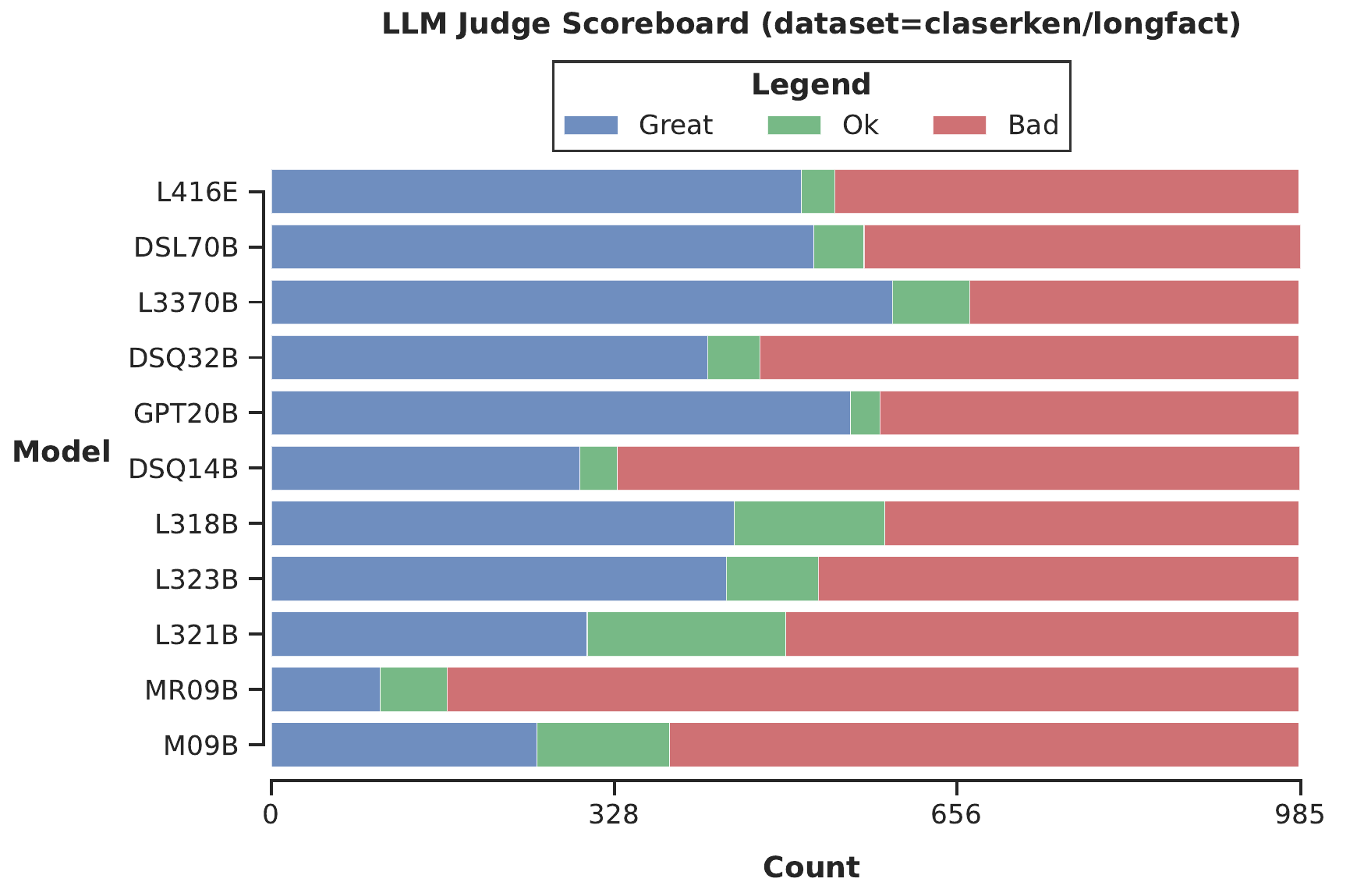}
    \caption{Distribution of LLM judge scores on LongFact dataset with GPT OSS 120B as the judge.}
    \label{fig:alternate_judge_scoreboard_longfact}
\end{figure}

\begin{figure}[hp]
    \centering
    \includegraphics[width=0.9\textwidth]{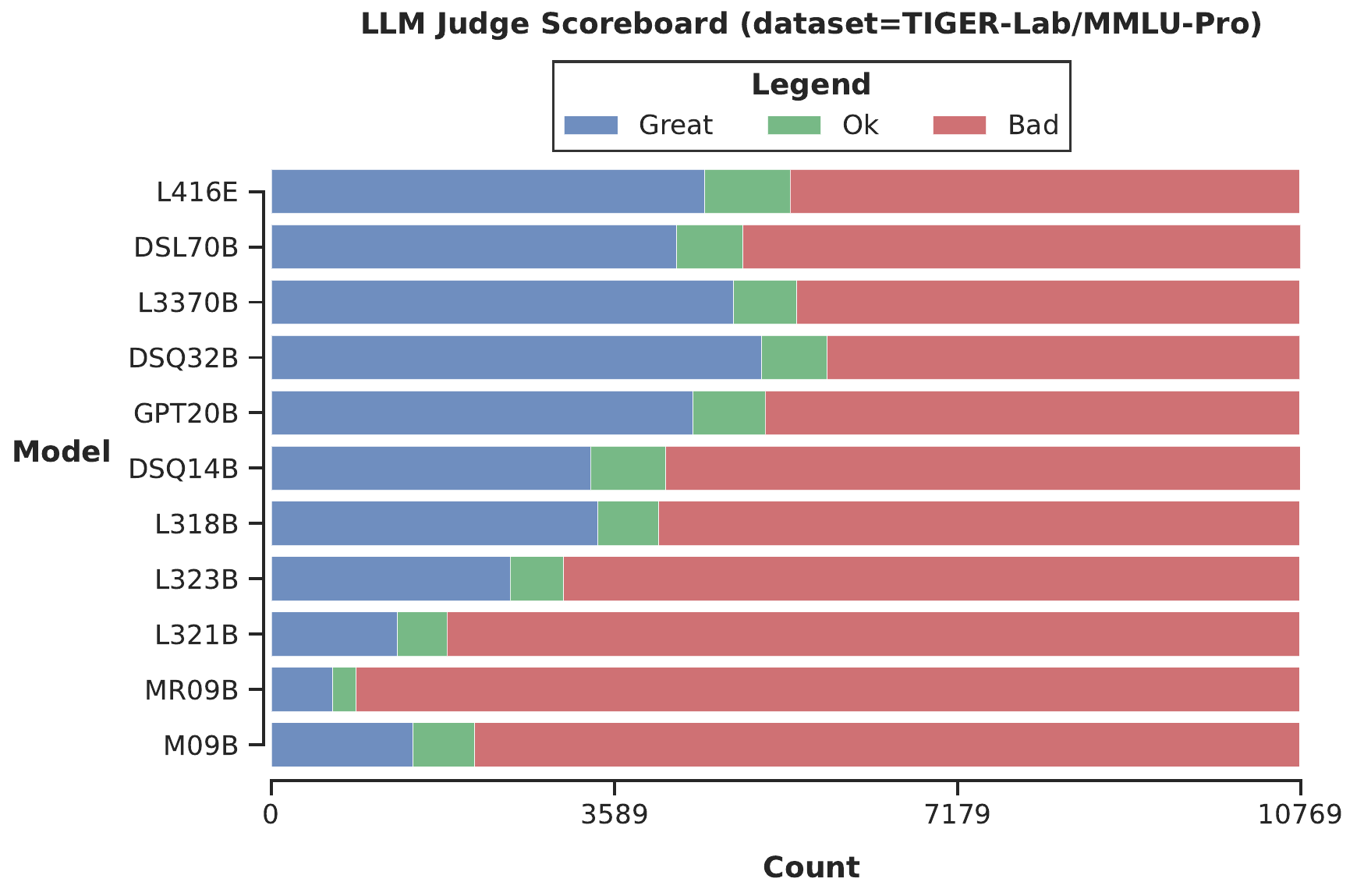}
    \caption{Distribution of LLM judge scores on MMLU-Pro dataset with GPT OSS 120B as the judge.}
    \label{fig:alternate_judge_scoreboard_mmlu_pro}
\end{figure}

\begin{figure}[hp]
    \centering
    \includegraphics[width=0.9\textwidth]{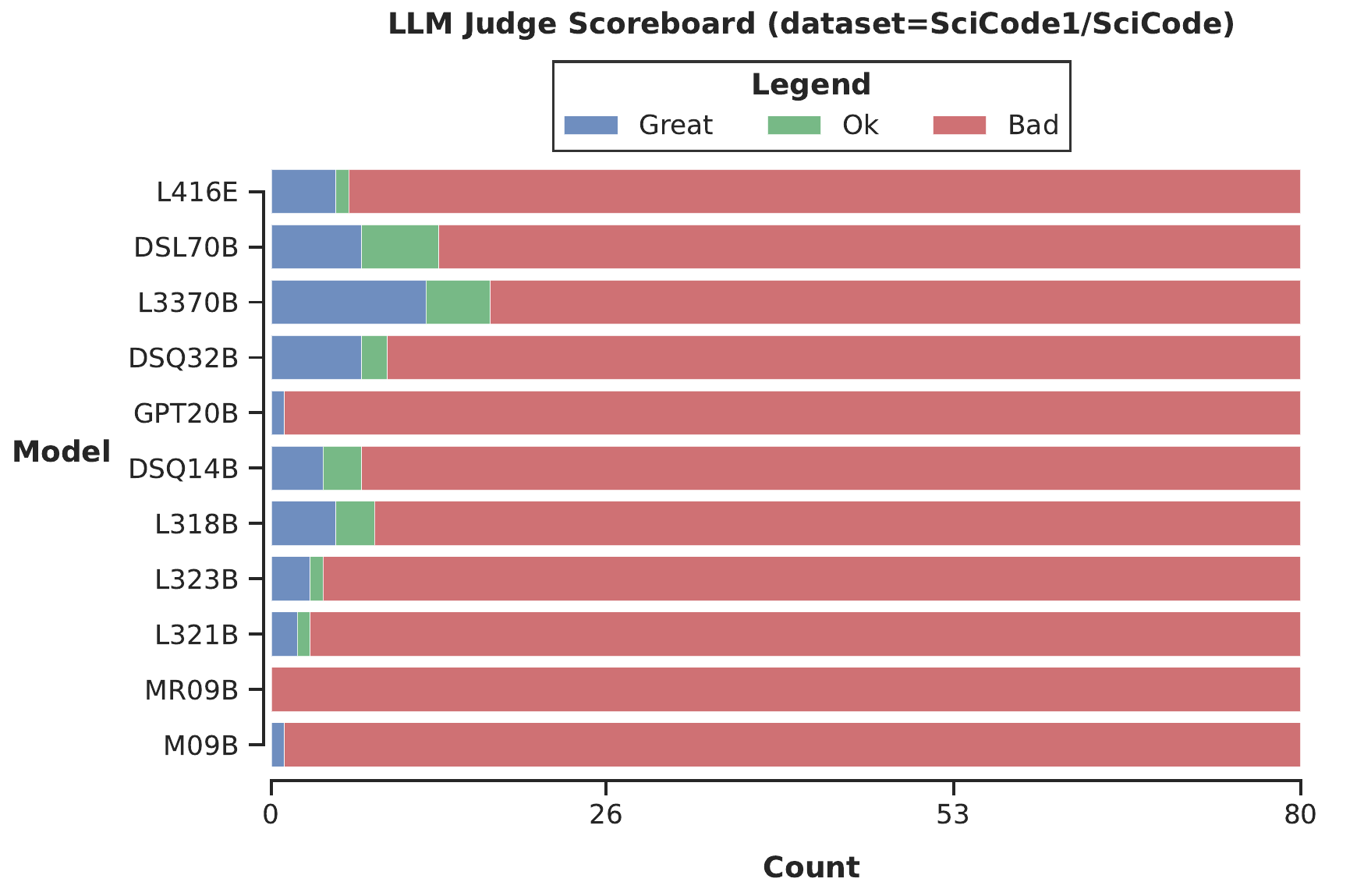}
    \caption{Distribution of LLM judge scores on SciCode dataset with GPT OSS 120B as the judge.}
    \label{fig:alternate_judge_scoreboard_scicode}
\end{figure}

\begin{figure}
    \centering
    \includegraphics[width=.48\textwidth]{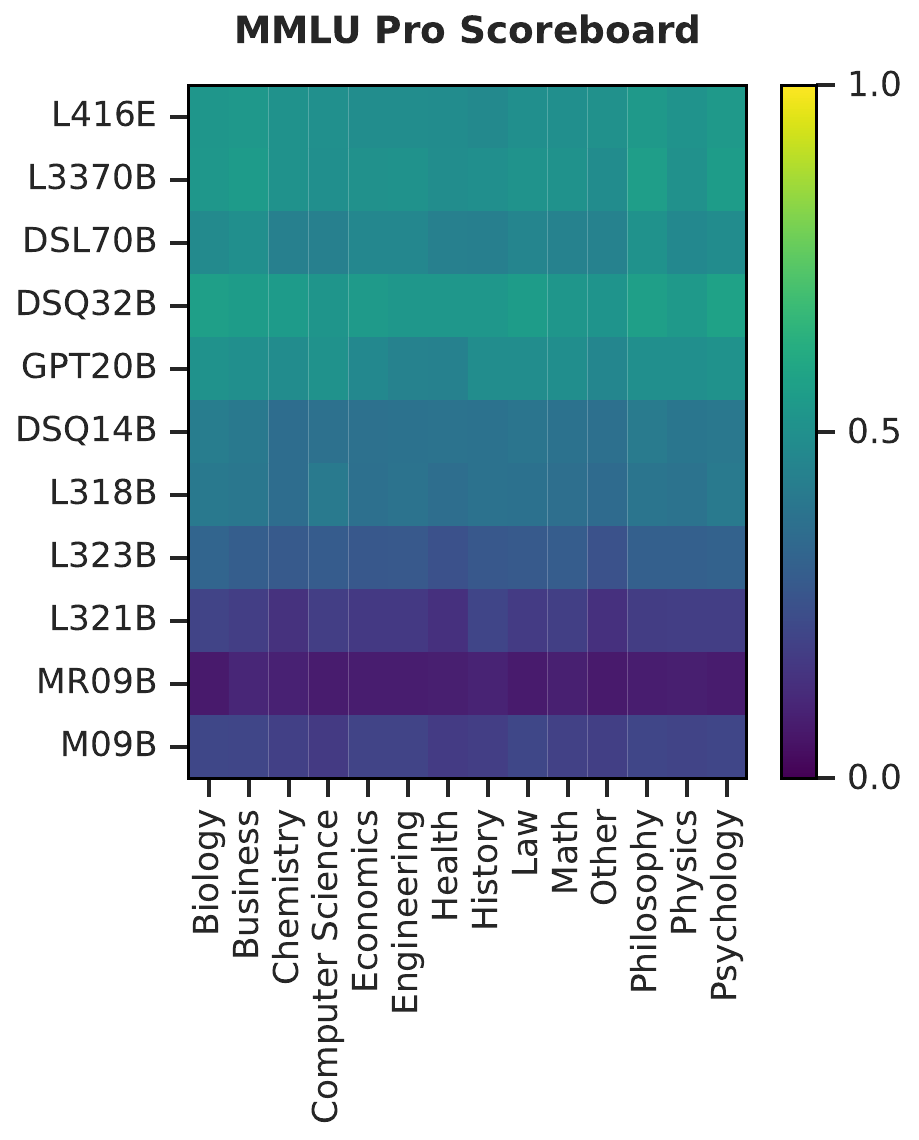}
    \caption{The probability of a model receiving a grade of either "great" or "ok" as a function of the query's category on the MMLU-Pro dataset when using GPT OSS 120B as the judge.}
    \label{fig:alternate_judge_mmlu_pro_scoreboard_by_category}
\end{figure}

\begin{figure}
    \centering
    \includegraphics[width=.48\textwidth]{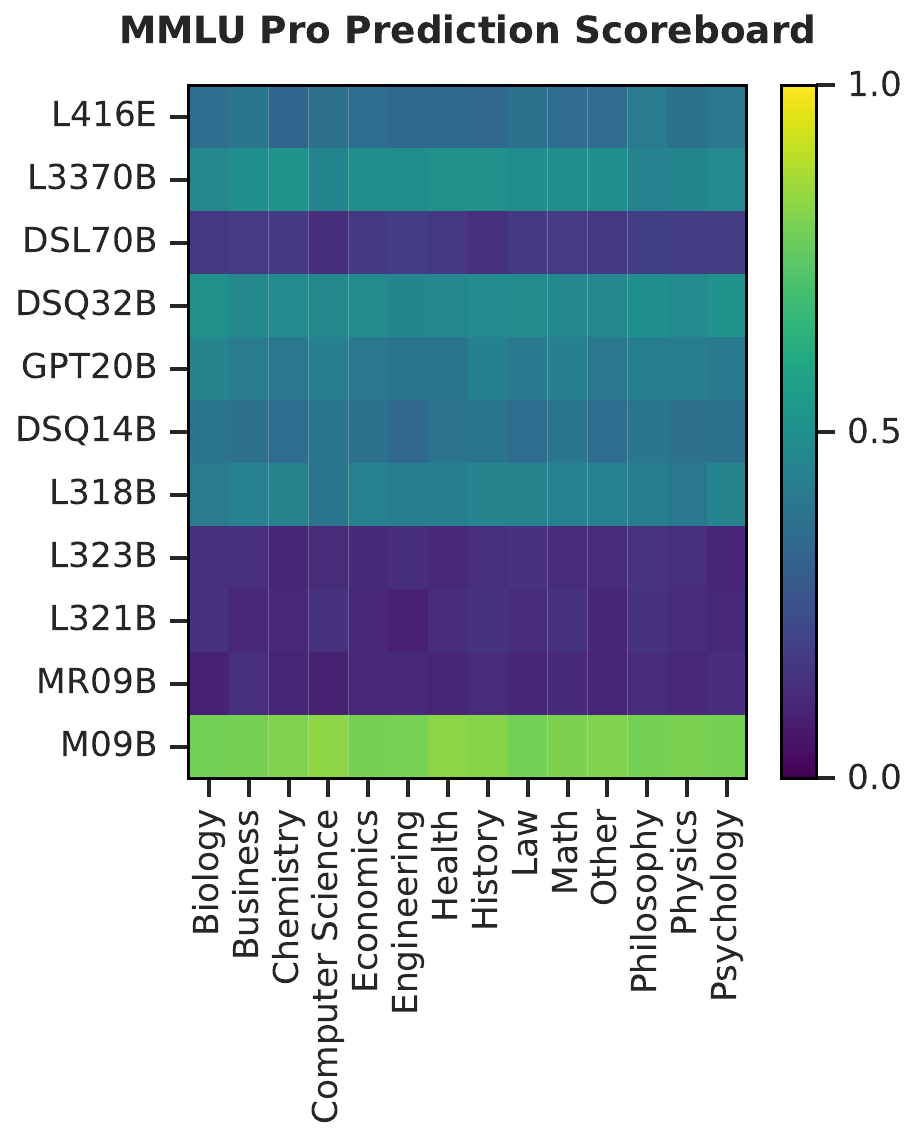}
    \caption{The probability of a model receiving a grade of either "great" or "ok" as a function of the query's category on the MMLU-Pro dataset when using GPT OSS 120B as the judge.}
    \label{fig:alternate_judge_mmlu_pro_zero_shot_prediction_accuracy_by_category}
\end{figure}

\begin{figure}[hp]
    \centering
    \includegraphics[width=0.9\textwidth]{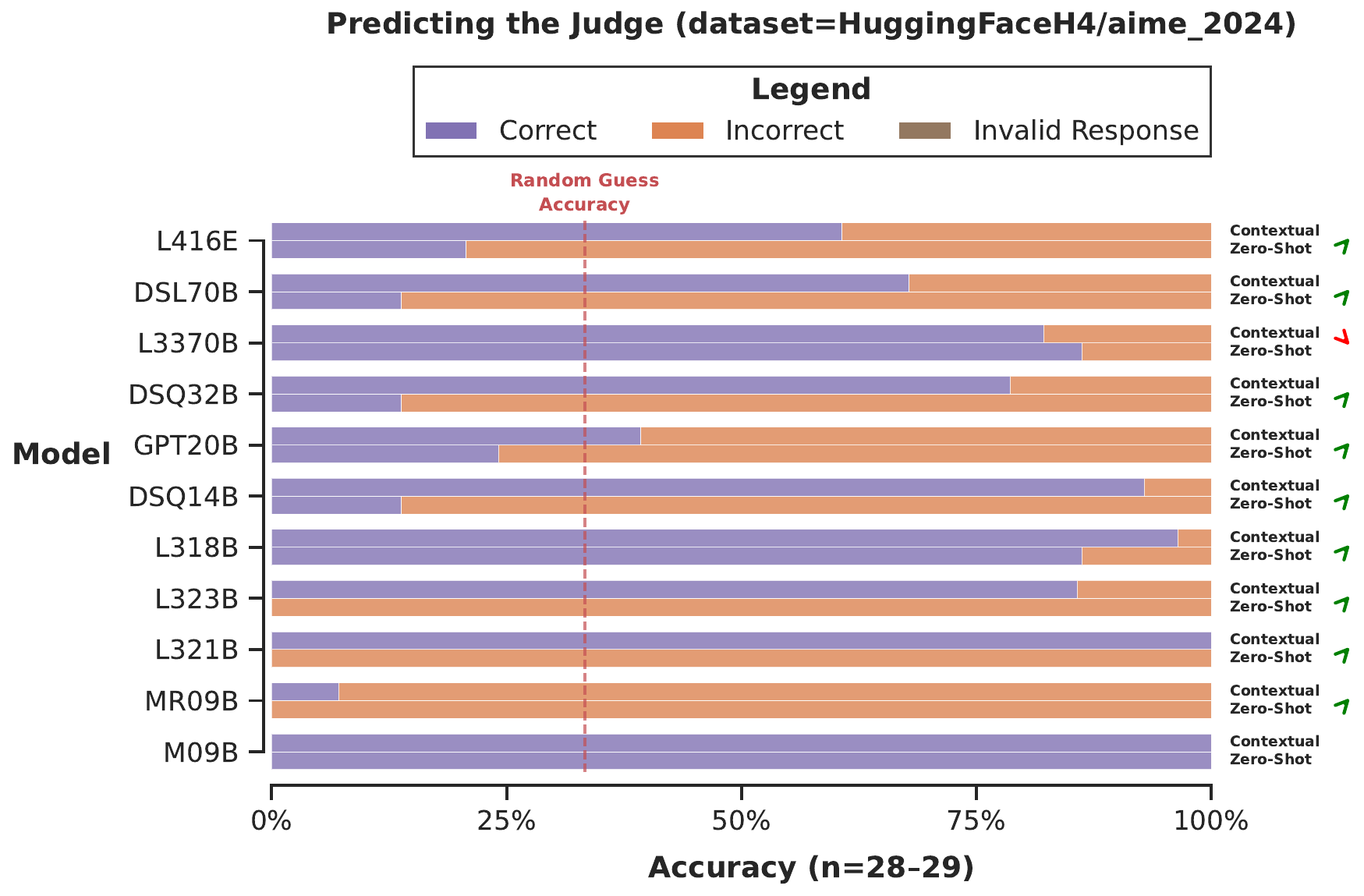}
    \caption{Contextual prediction accuracy of models on AIME 2024 dataset when using GPT OSS 120B as the judge.}
    \label{fig:alternate_judge_zero_shot_and_contextual_prediction_accuracy_aime_2024}
\end{figure}

\begin{figure}[hp]
    \centering
    \includegraphics[width=0.9\textwidth]{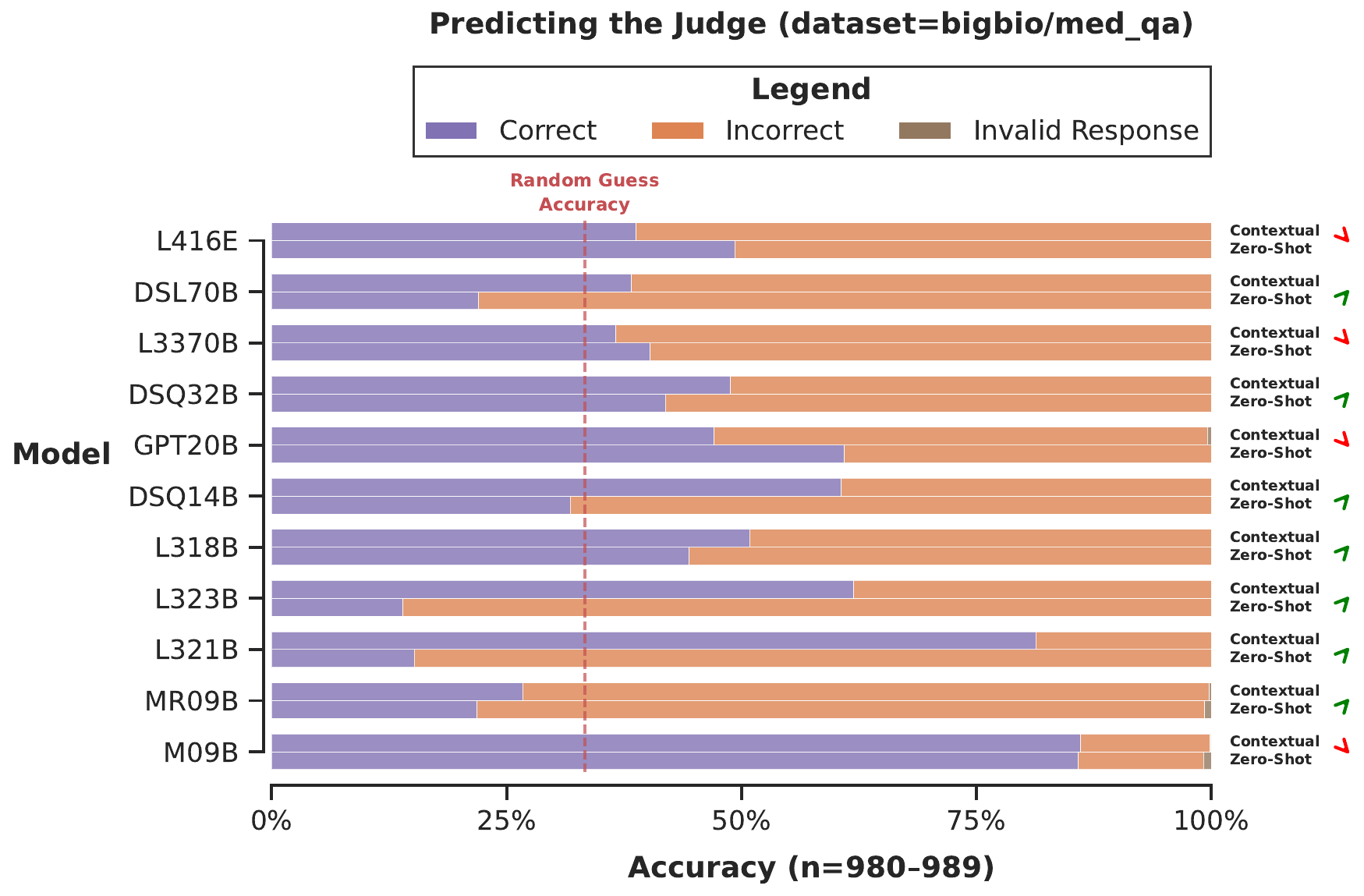}
    \caption{Contextual prediction accuracy of models on MedQA dataset when using GPT OSS 120B as the judge.}
    \label{fig:alternate_judge_zero_shot_and_contextual_prediction_accuracy_med_qa}
\end{figure}

\begin{figure}[hp]
    \centering
    \includegraphics[width=0.9\textwidth]{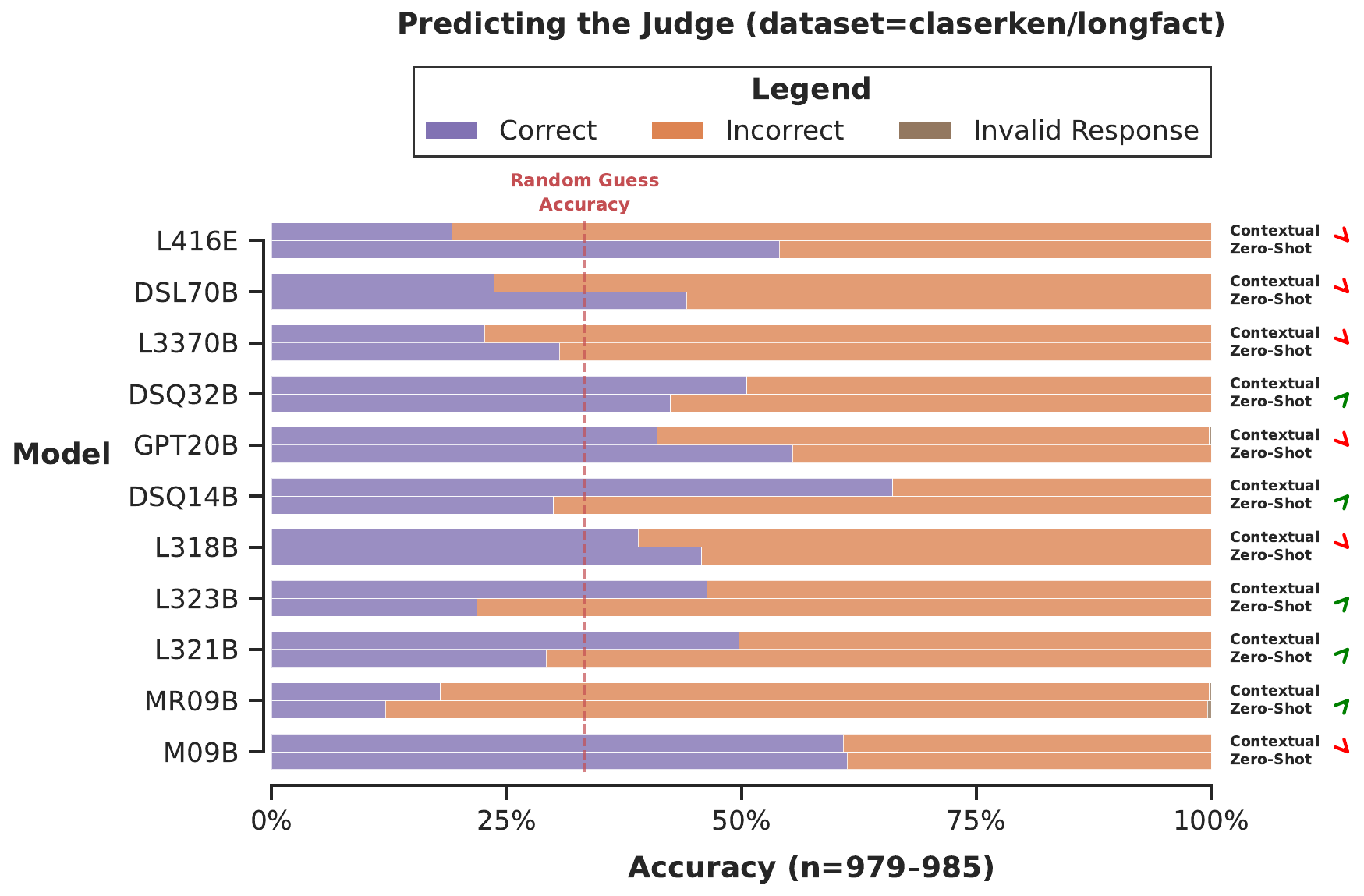}
    \caption{Contextual prediction accuracy of models on LongFact dataset when using GPT OSS 120B as the judge.}
    \label{fig:alternate_judge_zero_shot_and_contextual_prediction_accuracy_longfact}
\end{figure}

\begin{figure}[hp]
    \centering
    \includegraphics[width=0.9\textwidth]{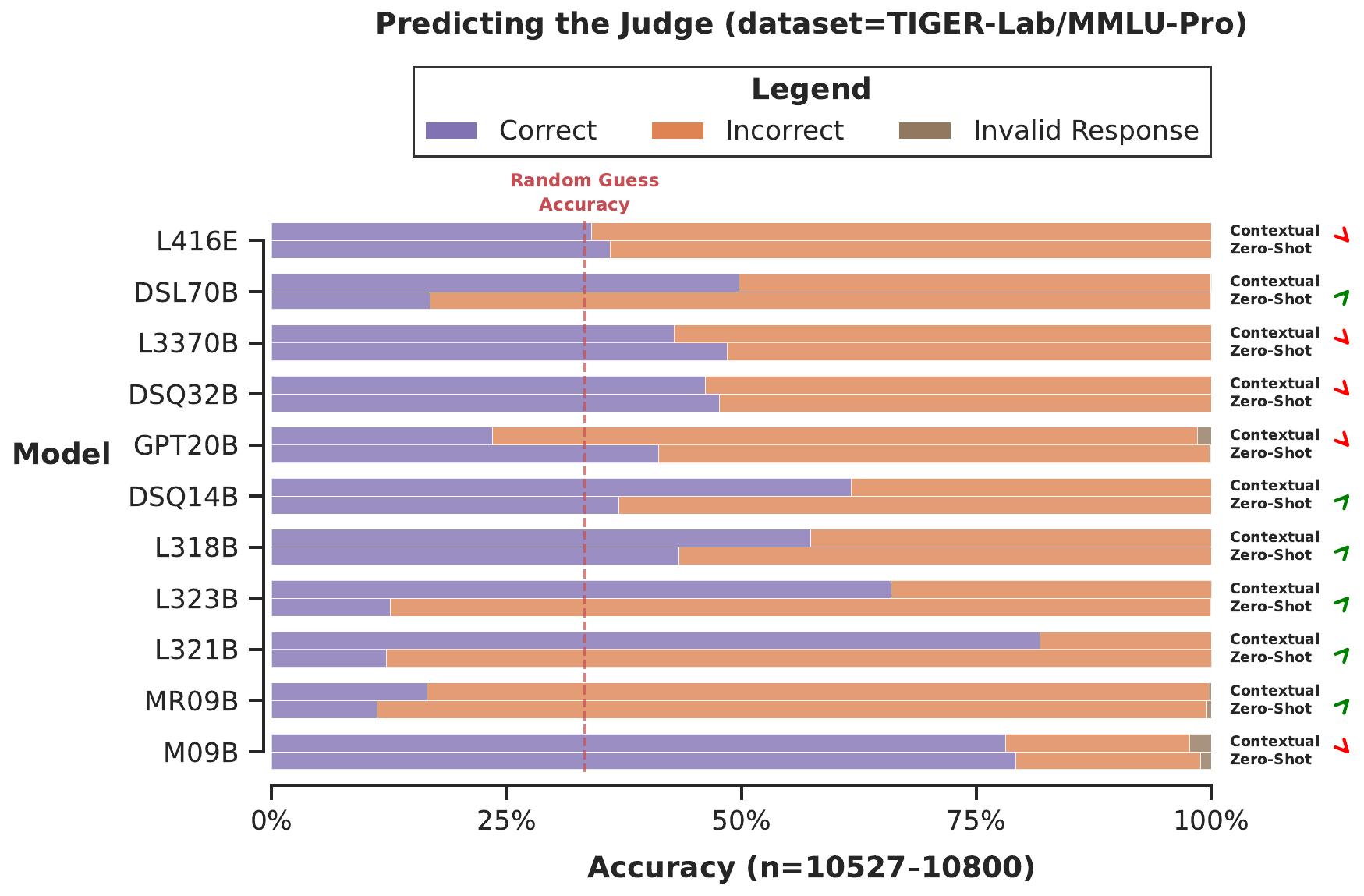}
    \caption{Contextual prediction accuracy of models on MMLU-Pro dataset when using GPT OSS 120B as the judge.}
    \label{fig:alternate_judge_zero_shot_and_contextual_prediction_accuracy_mmlu_pro}
\end{figure}

\begin{figure}[hp]
    \centering
    \includegraphics[width=0.9\textwidth]{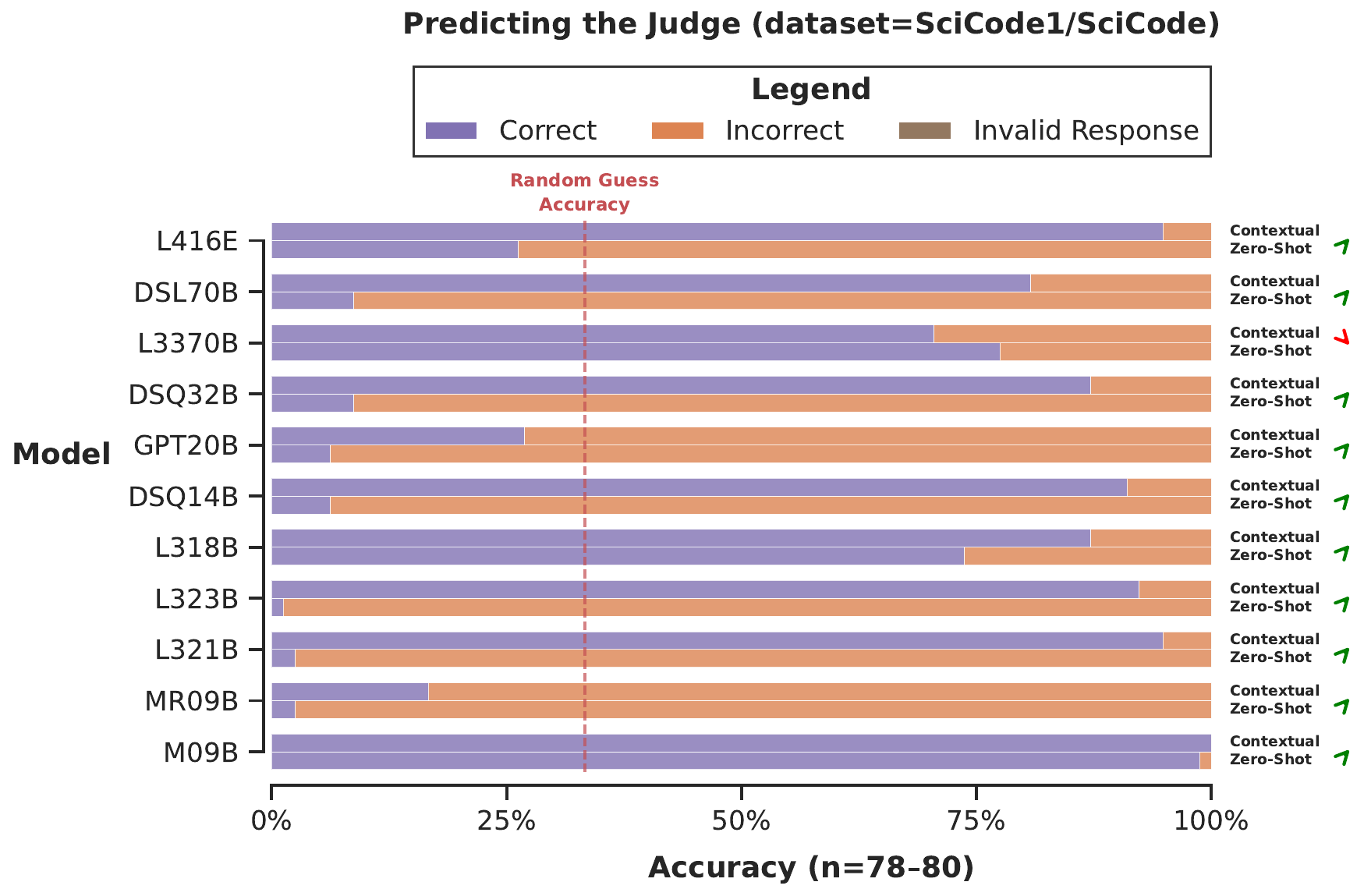}
    \caption{Contextual prediction accuracy of models on SciCode dataset when using GPT OSS 120B as the judge.}
    \label{fig:alternate_judge_zero_shot_and_contextual_prediction_accuracy_scicode}
\end{figure}

\clearpage

\section{Short Feedback Template Results}\label{app:short_feedback_template_results}

\begin{figure}[hp]
    \centering
    \includegraphics[width=0.9\textwidth]{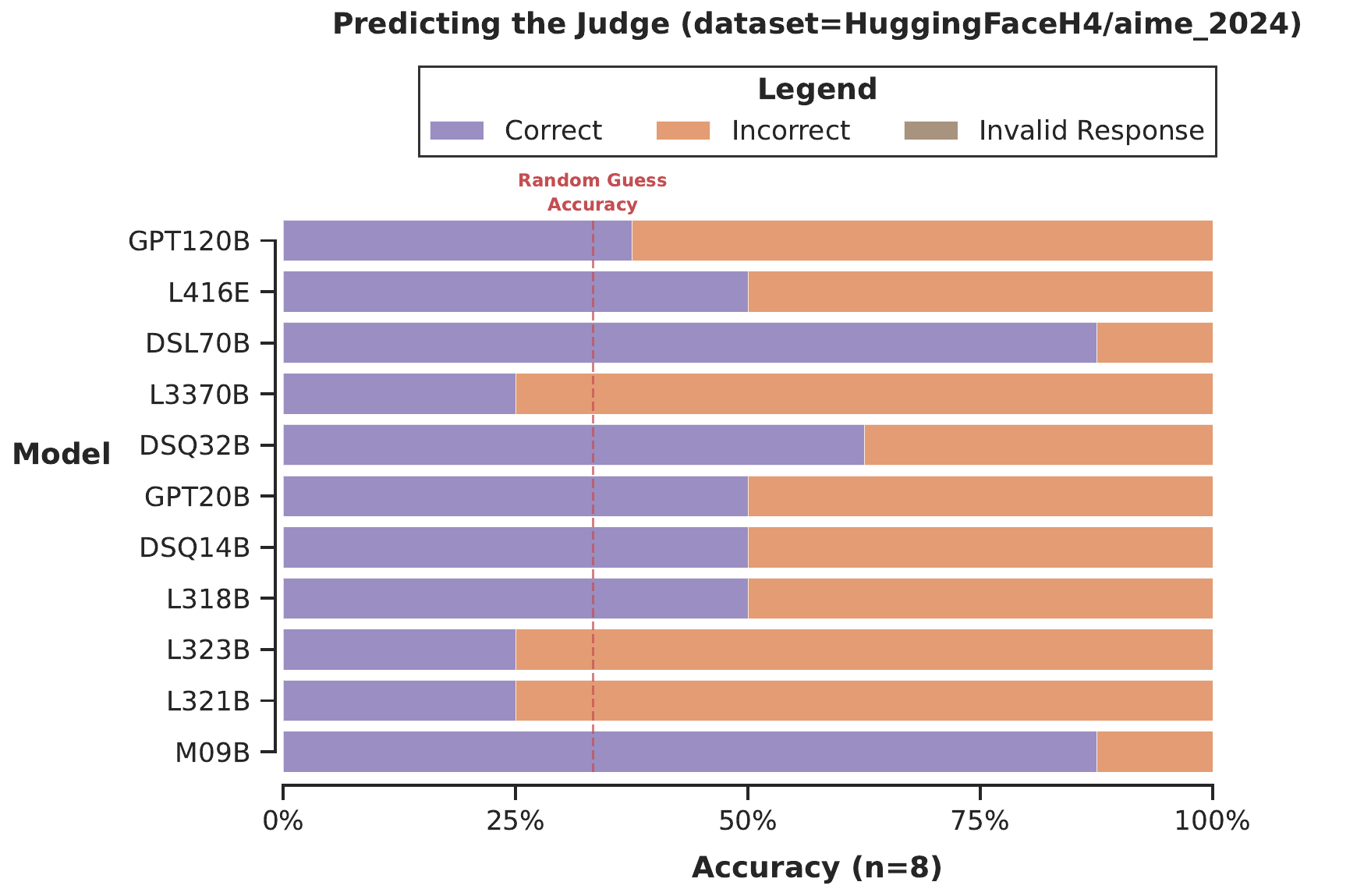}
    \caption{Contextual prediction accuracy of models on AIME 2024 dataset when using the short feedback template given in \cref{ppt:short_feedback_template}.}
    \label{fig:short_template_prediction_accuracy_aime_2024}
\end{figure}

\begin{figure}[hp]
    \centering
    \includegraphics[width=0.9\textwidth]{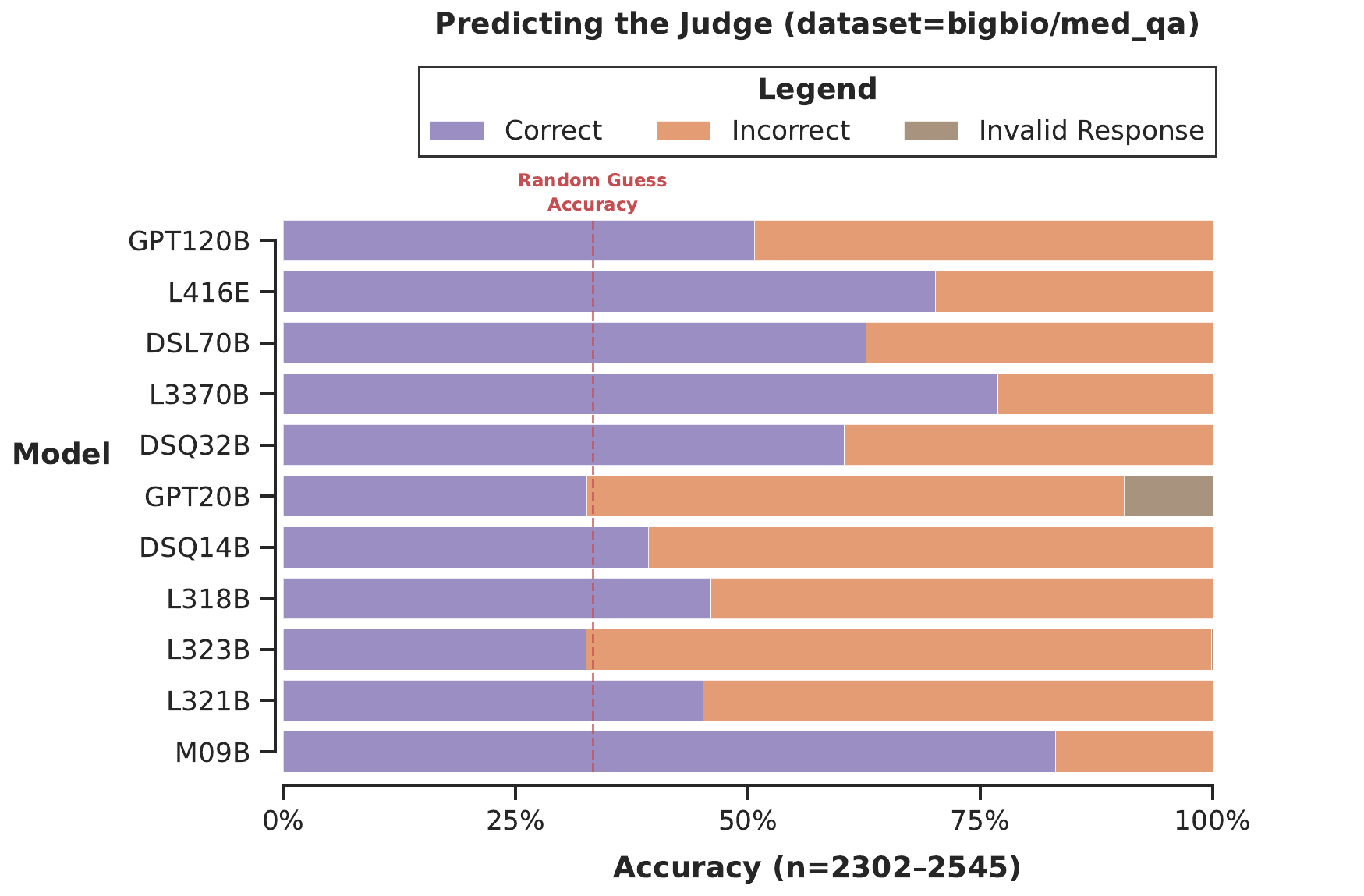}
    \caption{Contextual prediction accuracy of models on MedQA dataset when using the short feedback template given in \cref{ppt:short_feedback_template}.}
    \label{fig:short_template_prediction_accuracy_med_qa}
\end{figure}

\begin{figure}[hp]
    \centering
    \includegraphics[width=0.9\textwidth]{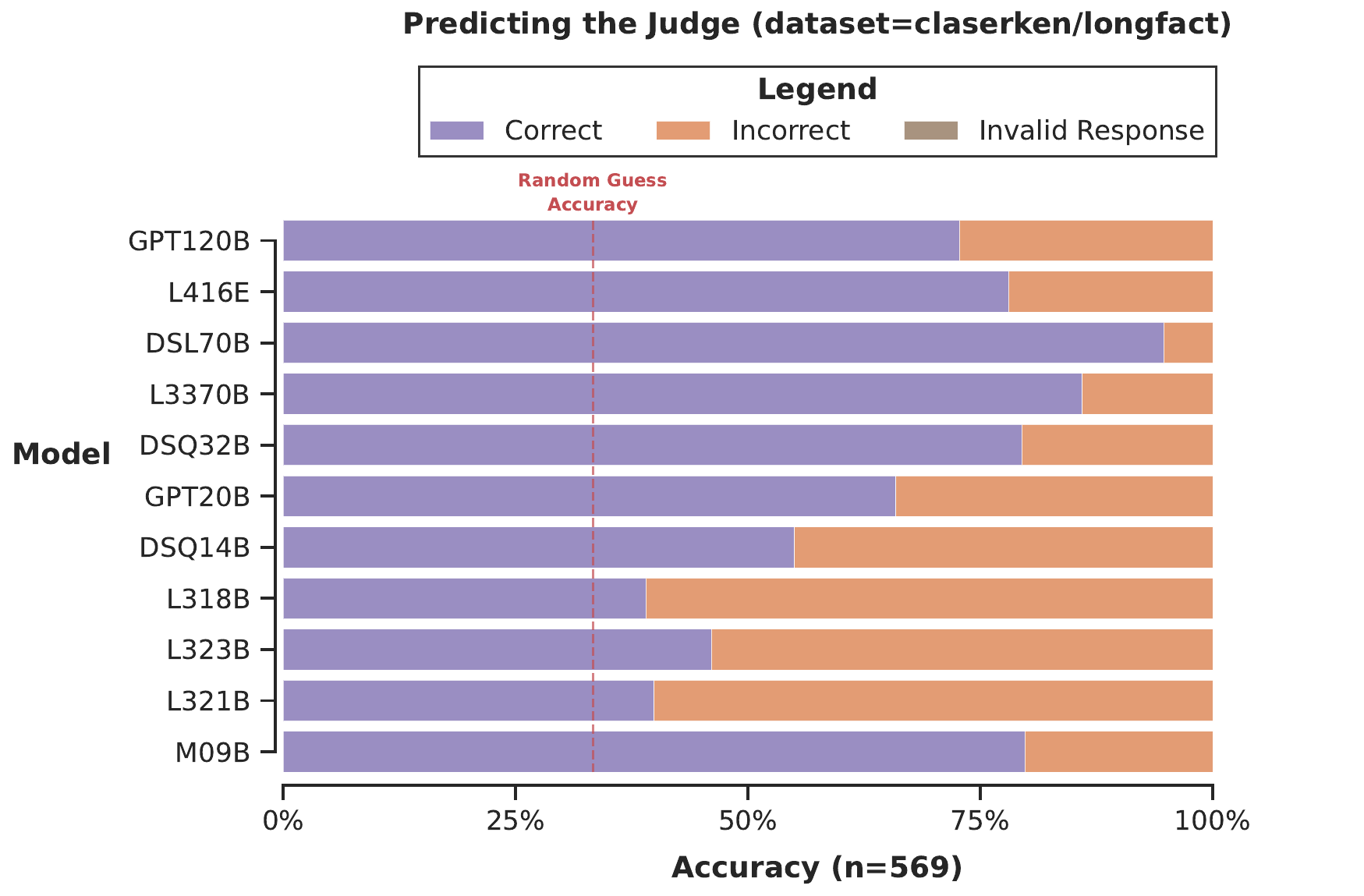}
    \caption{Contextual prediction accuracy of models on LongFact dataset when using the short feedback template given in \cref{ppt:short_feedback_template}.}
    \label{fig:short_template_prediction_accuracy_longfact}
\end{figure}

\begin{figure}[hp]
    \centering
    \includegraphics[width=0.9\textwidth]{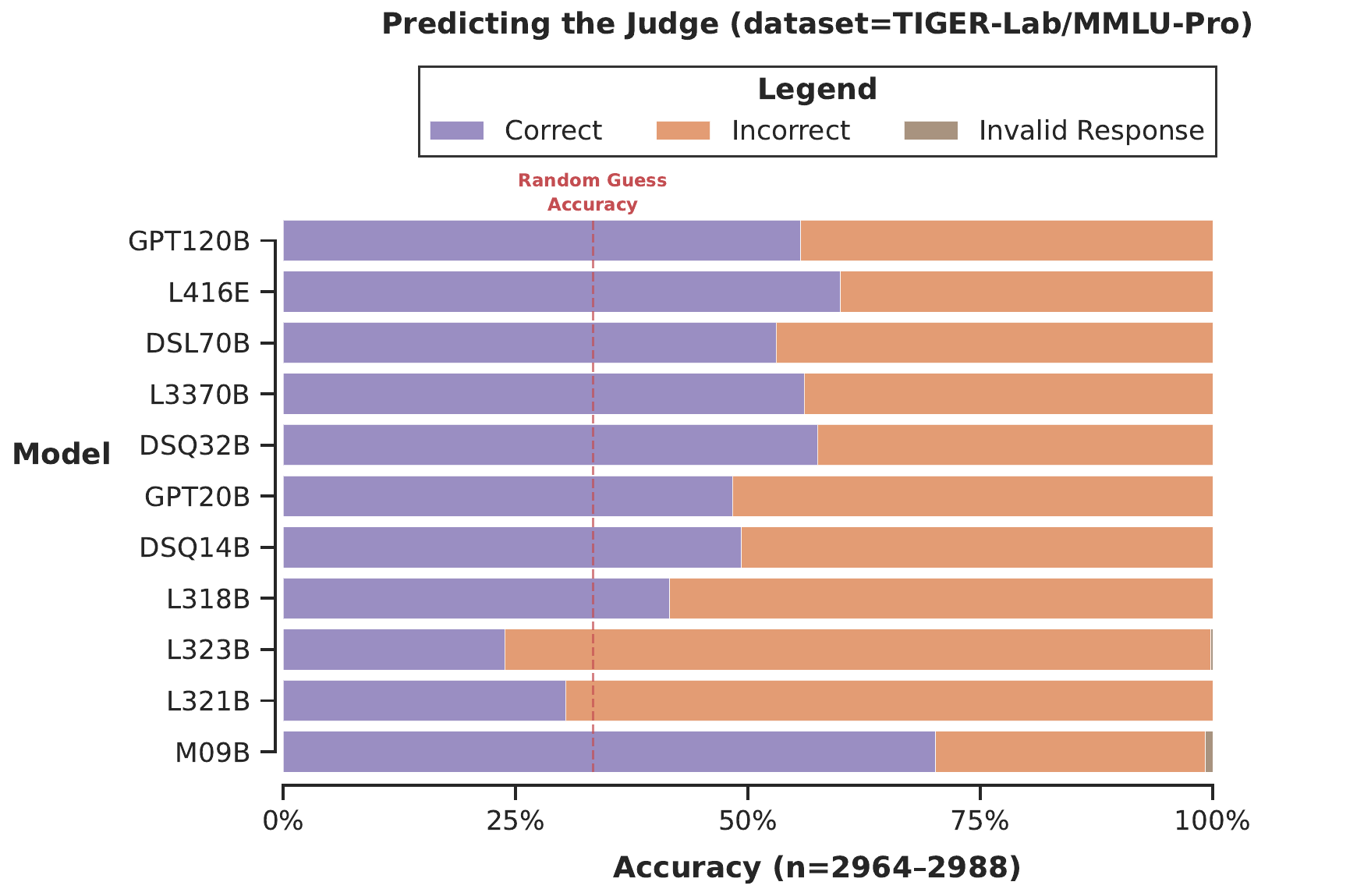}
    \caption{Contextual prediction accuracy of models on MMLU-Pro dataset when using the short feedback template given in \cref{ppt:short_feedback_template}.}
    \label{fig:short_template_prediction_accuracy_mmlu_pro}
\end{figure}

\begin{figure}[hp]
    \centering
    \includegraphics[width=0.9\textwidth]{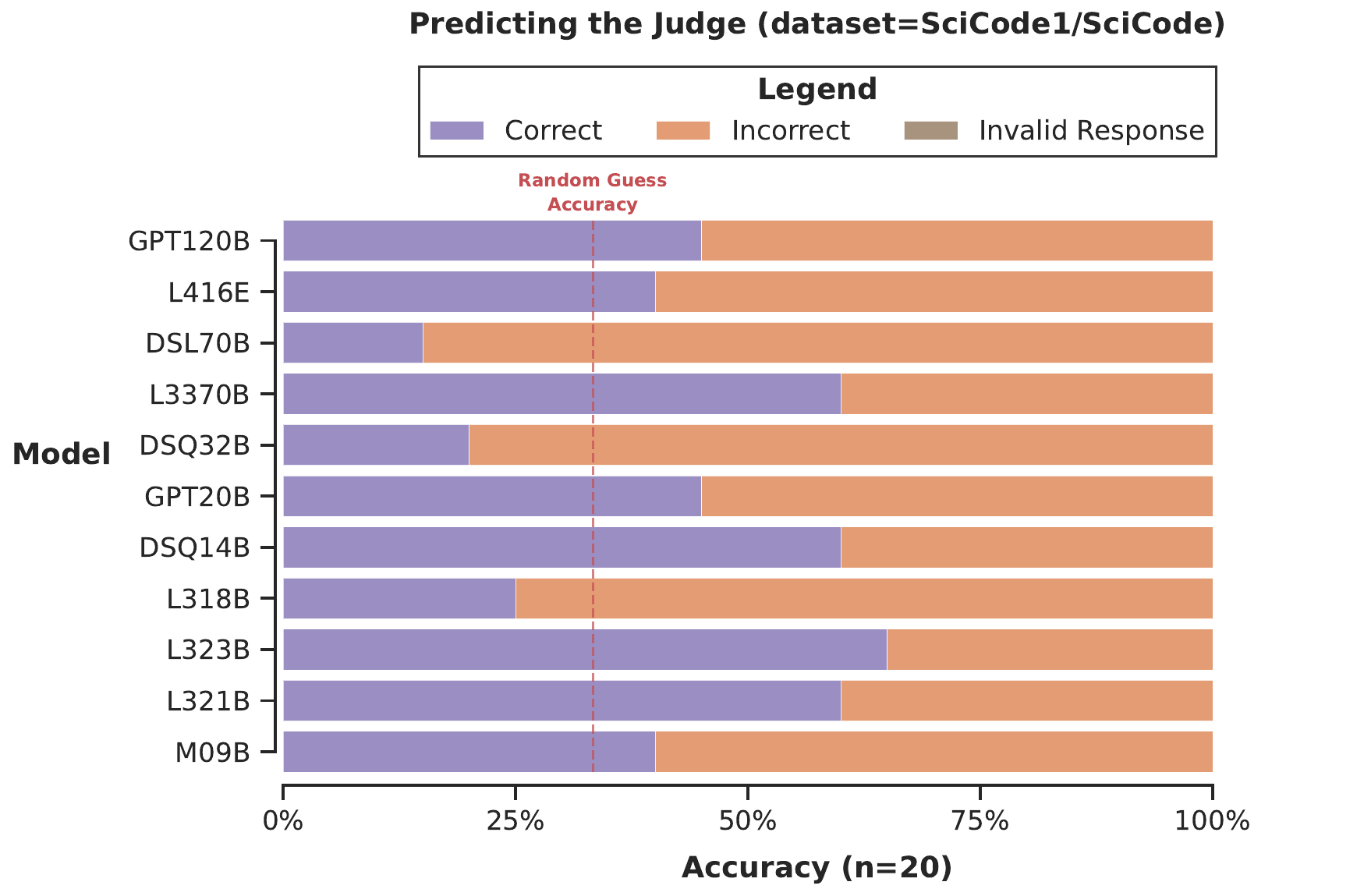}
    \caption{Contextual prediction accuracy of models on SciCode dataset when using the short feedback template given in \cref{ppt:short_feedback_template}.}
    \label{fig:short_template_prediction_accuracy_scicode}
\end{figure}

The feedback template shown in \cref{ppt:long_feedback_template} and used in our primary experiments is exceedingly long.
To determine whether this is important or not, we repeated our zero-shot experiments with the shorter feedback template shown in \cref{ppt:short_feedback_template}.
The results are shown in \cref{fig:short_template_prediction_accuracy_aime_2024,fig:short_template_prediction_accuracy_med_qa,fig:short_template_prediction_accuracy_longfact,fig:short_template_prediction_accuracy_mmlu_pro,fig:short_template_prediction_accuracy_scicode}.
The principal takeaway from this ablation is that the longer and more comprehensive template led to a much stronger performance than the short template.

\clearpage

\section{Mischievous Rubric Results}\label{app:mischievous_rubric_results}

\begin{figure}[hp]
    \centering
    \includegraphics[width=0.9\textwidth]{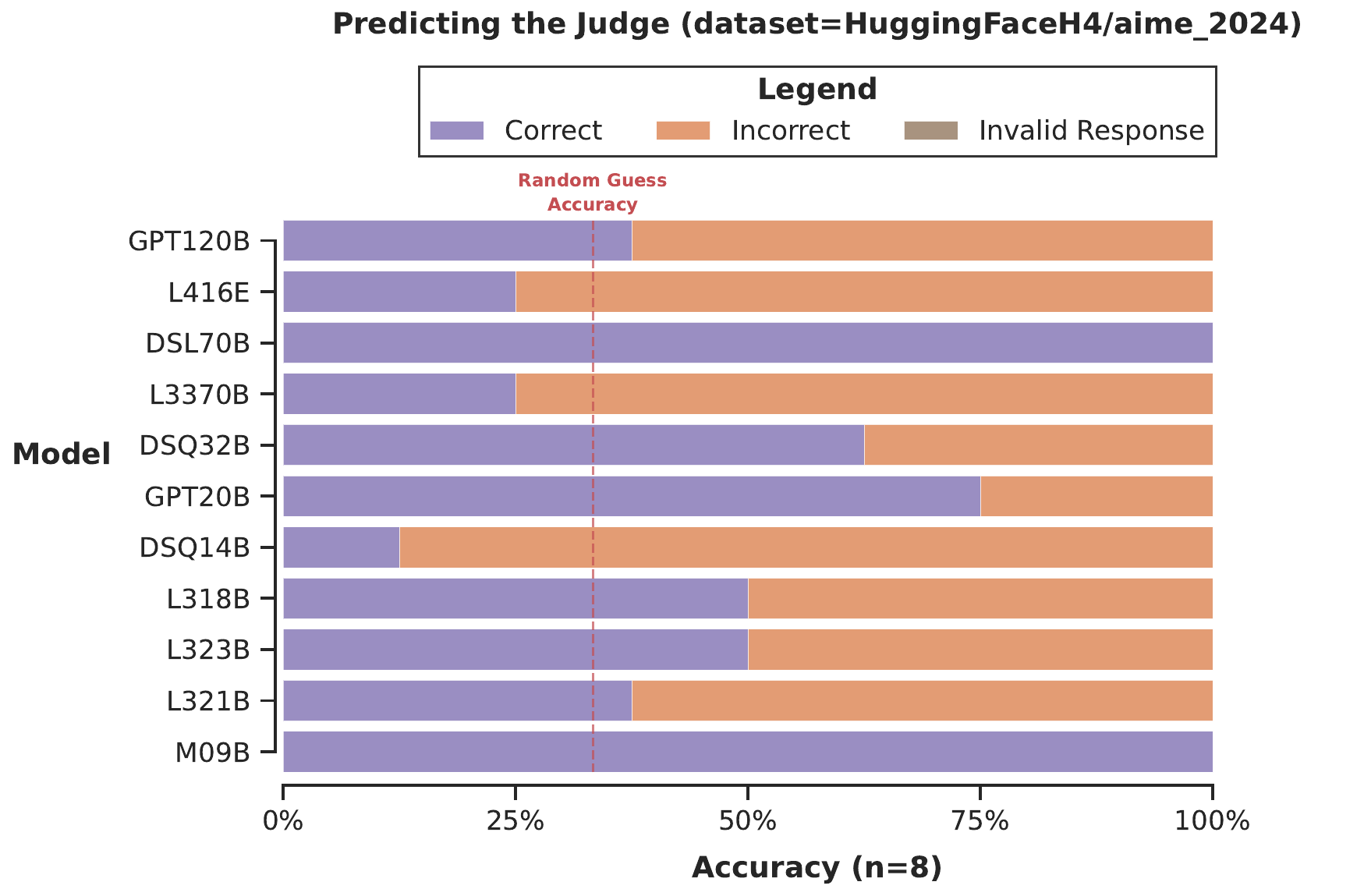}
    \caption{Contextual prediction accuracy of models on AIME 2024 dataset when evaluation is performed according to the mischievous rubric given in \cref{ppt:mischievous_rubric}.}
    \label{fig:mischievous_rubric_prediction_accuracy_aime_2024}
\end{figure}

\begin{figure}[hp]
    \centering
    \includegraphics[width=0.9\textwidth]{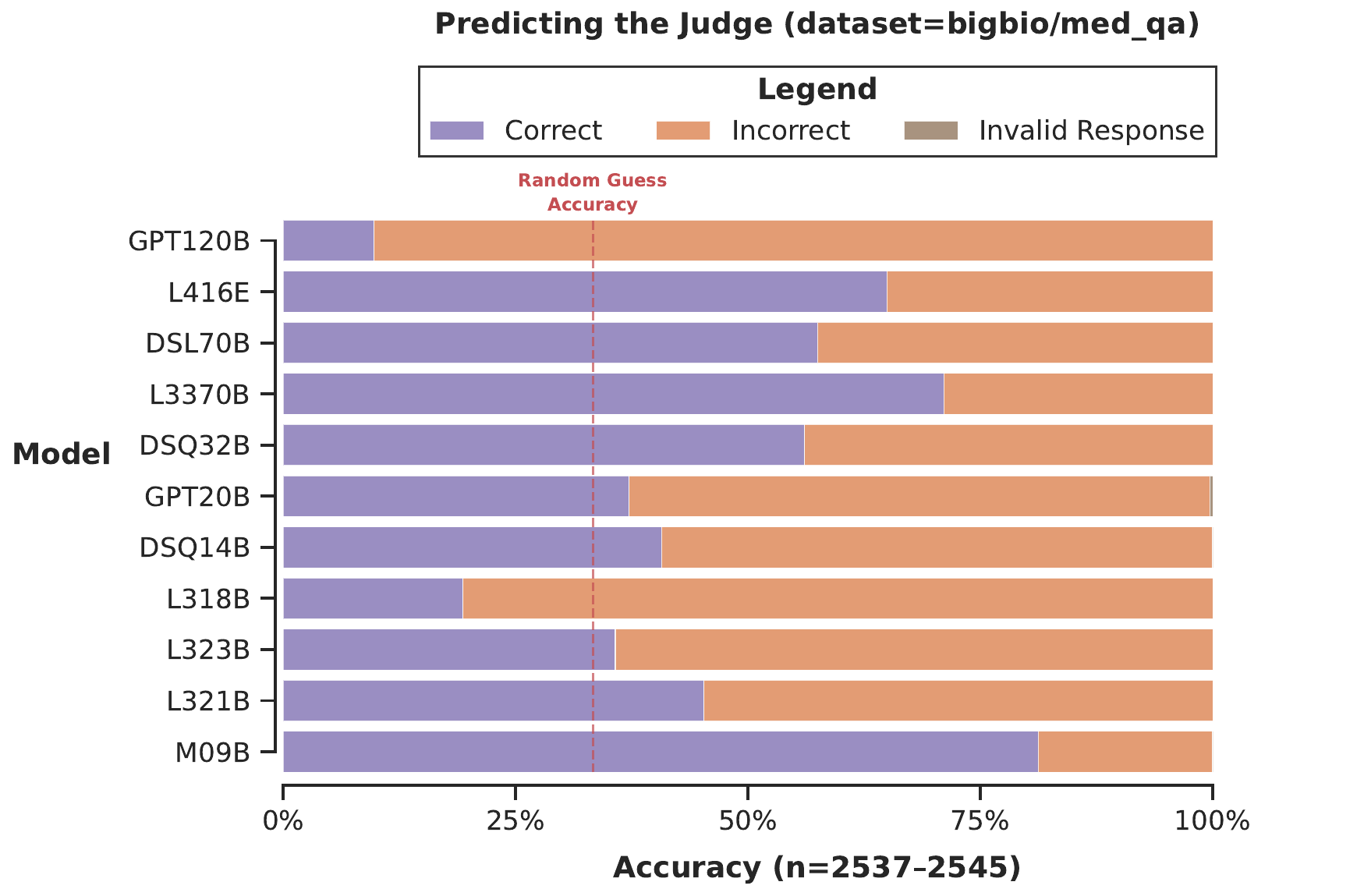}
    \caption{Contextual prediction accuracy of models on MedQA dataset when evaluation is performed according to the mischievous rubric given in \cref{ppt:mischievous_rubric}.}
    \label{fig:mischievous_rubric_prediction_accuracy_med_qa}
\end{figure}

\begin{figure}[hp]
    \centering
    \includegraphics[width=0.9\textwidth]{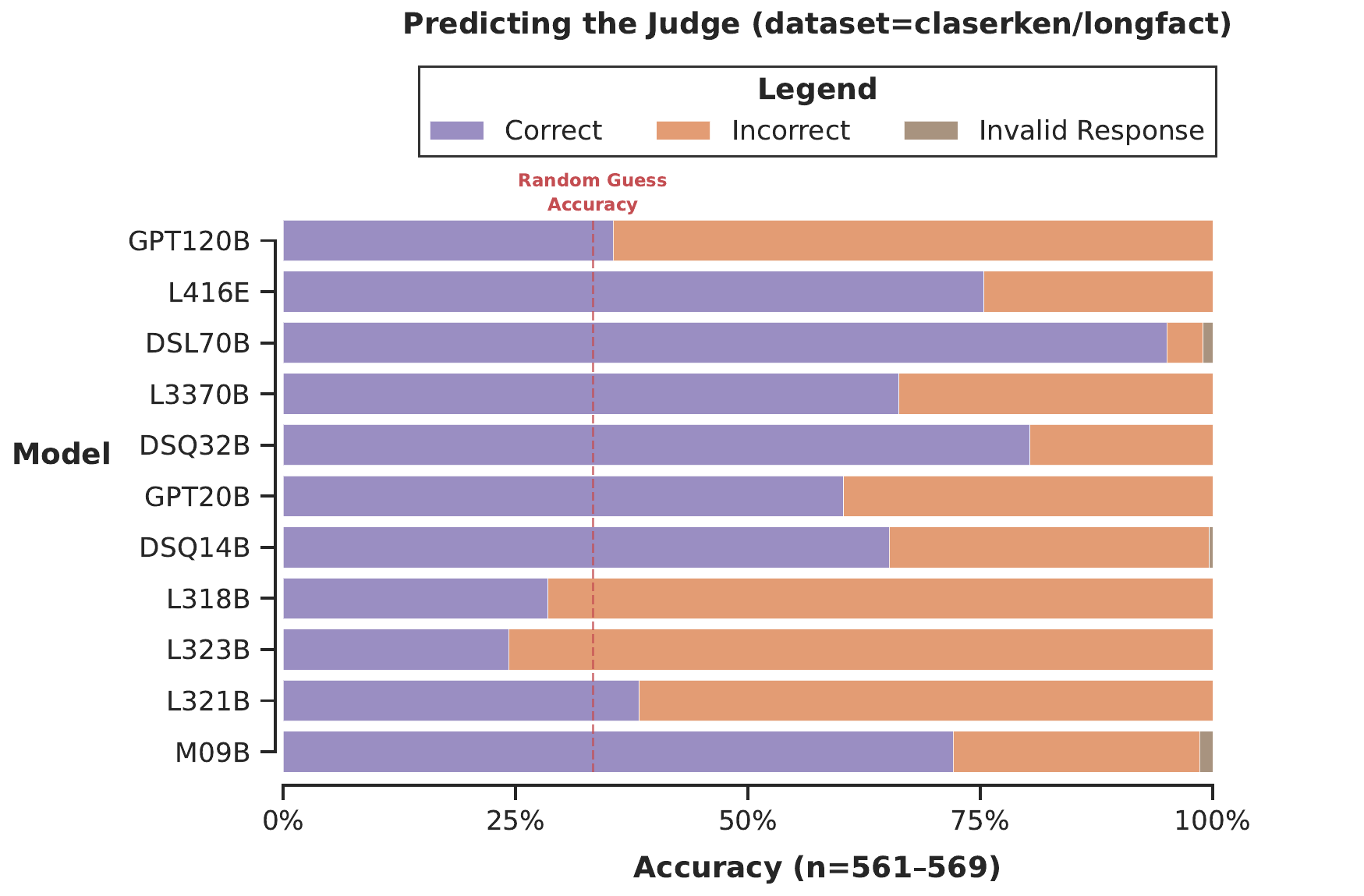}
    \caption{Contextual prediction accuracy of models on LongFact dataset when evaluation is performed according to the mischievous rubric given in \cref{ppt:mischievous_rubric}.}
    \label{fig:mischievous_rubric_prediction_accuracy_longfact}
\end{figure}

\begin{figure}[hp]
    \centering
    \includegraphics[width=0.9\textwidth]{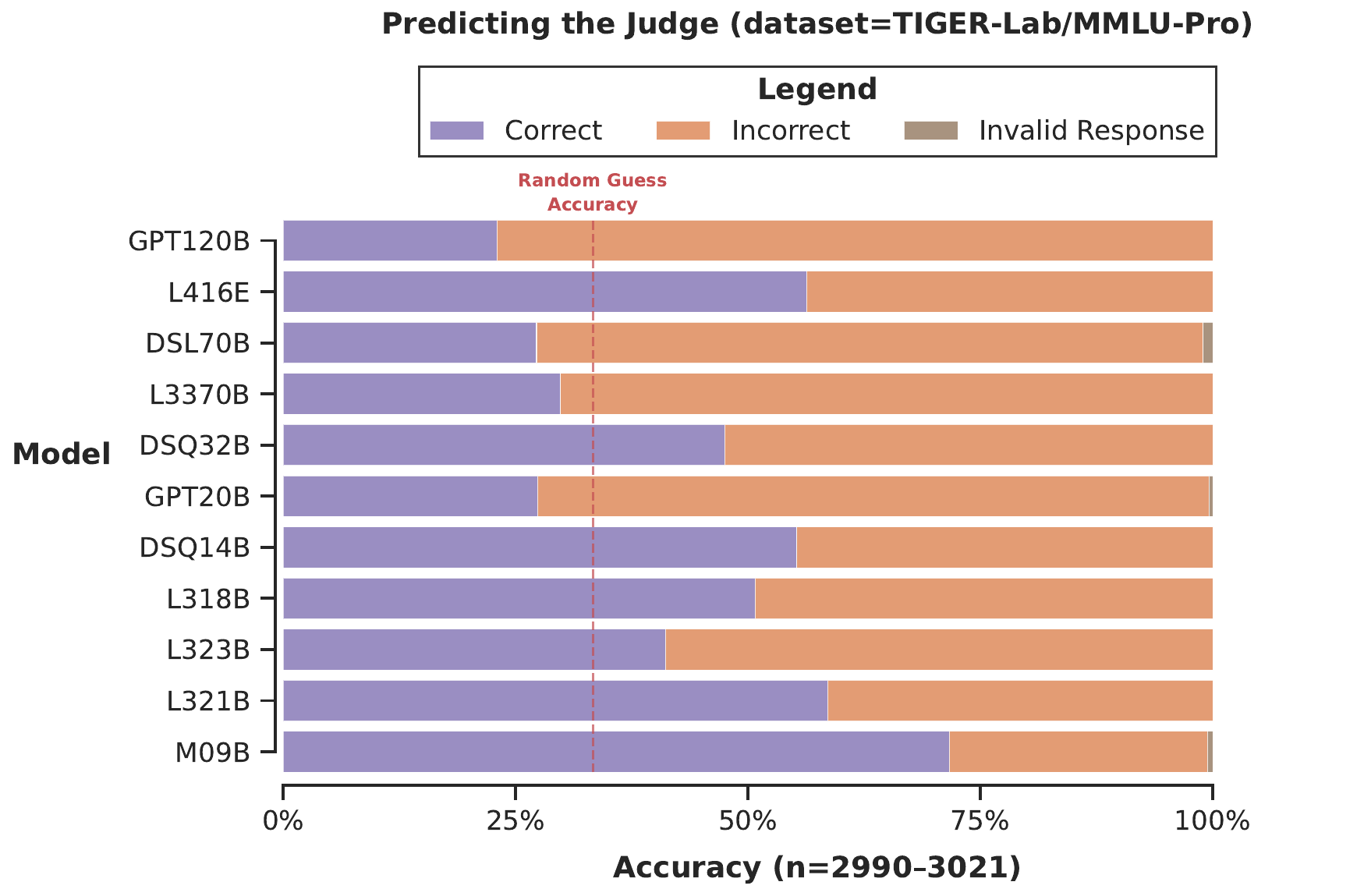}
    \caption{Contextual prediction accuracy of models on MMLU-Pro dataset when evaluation is performed according to the mischievous rubric given in \cref{ppt:mischievous_rubric}.}
    \label{fig:mischievous_rubric_prediction_accuracy_mmlu_pro}
\end{figure}

\begin{figure}[hp]
    \centering
    \includegraphics[width=0.9\textwidth]{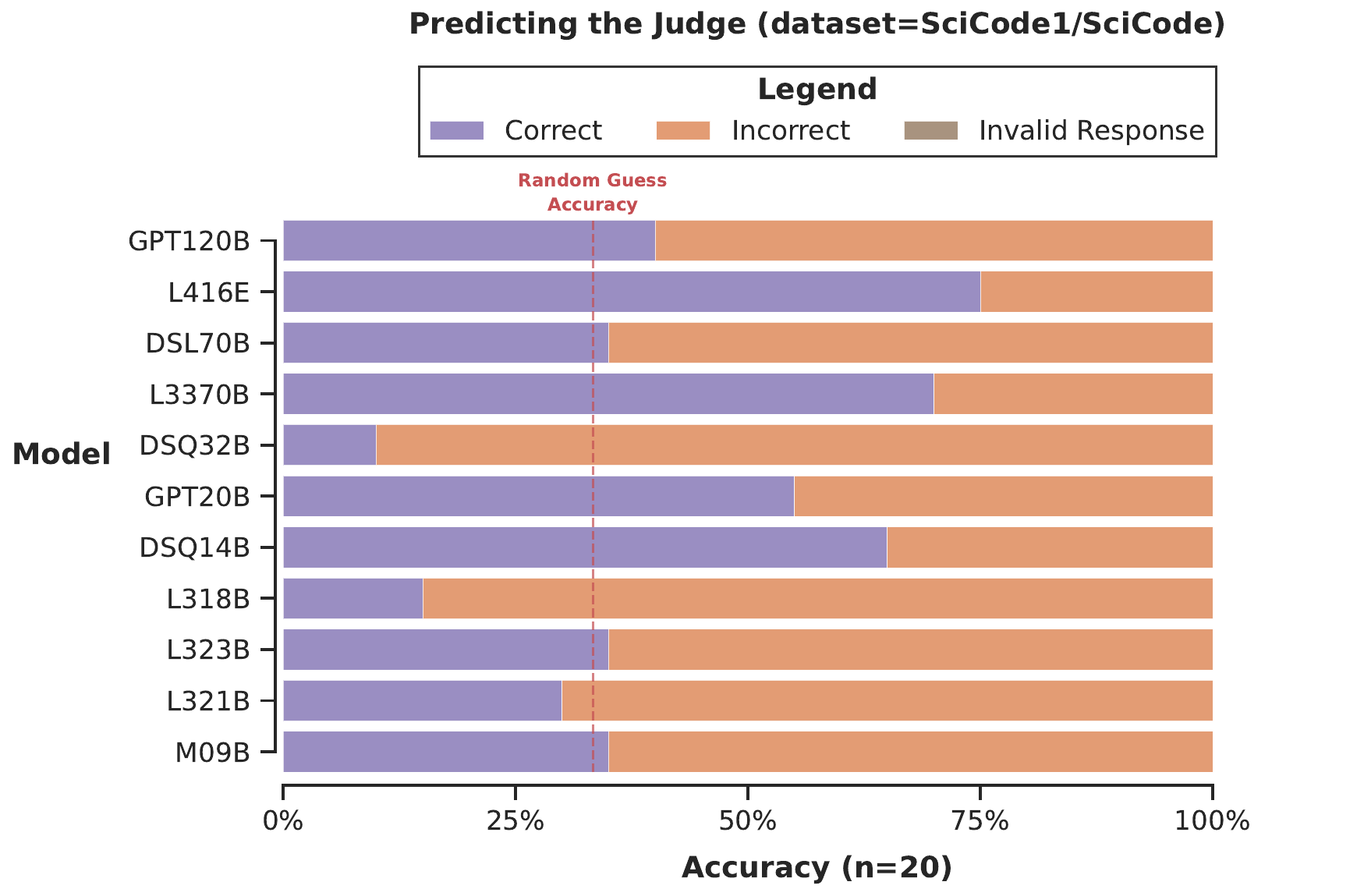}
    \caption{Contextual prediction accuracy of models on SciCode dataset when evaluation is performed according to the mischievous rubric given in \cref{ppt:mischievous_rubric}.}
    \label{fig:mischievous_rubric_prediction_accuracy_scicode}
\end{figure}

The rubric used in our primary experiments (and shown in \cref{ppt:rubric}) scores models according to a naive interpretation of performance.
Part of the benefit of using a LLM or agentic judge is that you can define the evaluation criteria in a freeform manner.
To demonstrate that this flexibility carries over to the prediction task, we repeated our experiment using an alternative, reversed rubric for evaluating the models.
This rubric is shown in \cref{ppt:mischievous_rubric} with the rest of the prompt template used in \cref{ppt:contextual_other_model_score_prediction_template}.
The results of this experiment are shown in \cref{fig:mischievous_rubric_prediction_accuracy_aime_2024,fig:mischievous_rubric_prediction_accuracy_med_qa,fig:mischievous_rubric_prediction_accuracy_longfact,fig:mischievous_rubric_prediction_accuracy_mmlu_pro,fig:mischievous_rubric_prediction_accuracy_scicode}.
While the predictive ability of the models given in \cref{fig:mischievous_rubric_prediction_accuracy_aime_2024,fig:mischievous_rubric_prediction_accuracy_med_qa,fig:mischievous_rubric_prediction_accuracy_longfact,fig:mischievous_rubric_prediction_accuracy_mmlu_pro,fig:mischievous_rubric_prediction_accuracy_scicode} is clearly less than than those given in \cref{fig:zero_shot_and_contextual_prediction_accuracy_with_finetuned_aime_2024,fig:zero_shot_and_contextual_prediction_accuracy_with_finetuned_med_qa,fig:zero_shot_and_contextual_prediction_accuracy_with_finetuned_longfact,fig:zero_shot_and_contextual_prediction_accuracy_with_finetuned_mmlu_pro,fig:zero_shot_and_contextual_prediction_accuracy_with_finetuned_scicode}, a clear (better than random) predictive ability is displayed by most of the models on most of the datasets.
This suggests that---in addition to using the report cards---the models are relying on their intrinsic understanding of themselves or the difficulty of the questions.

\clearpage

\section{Prediction Accuracy without Fine-Tuned Models}

\begin{figure}[ht]
    \centering
    \includegraphics[width=.9\textwidth]{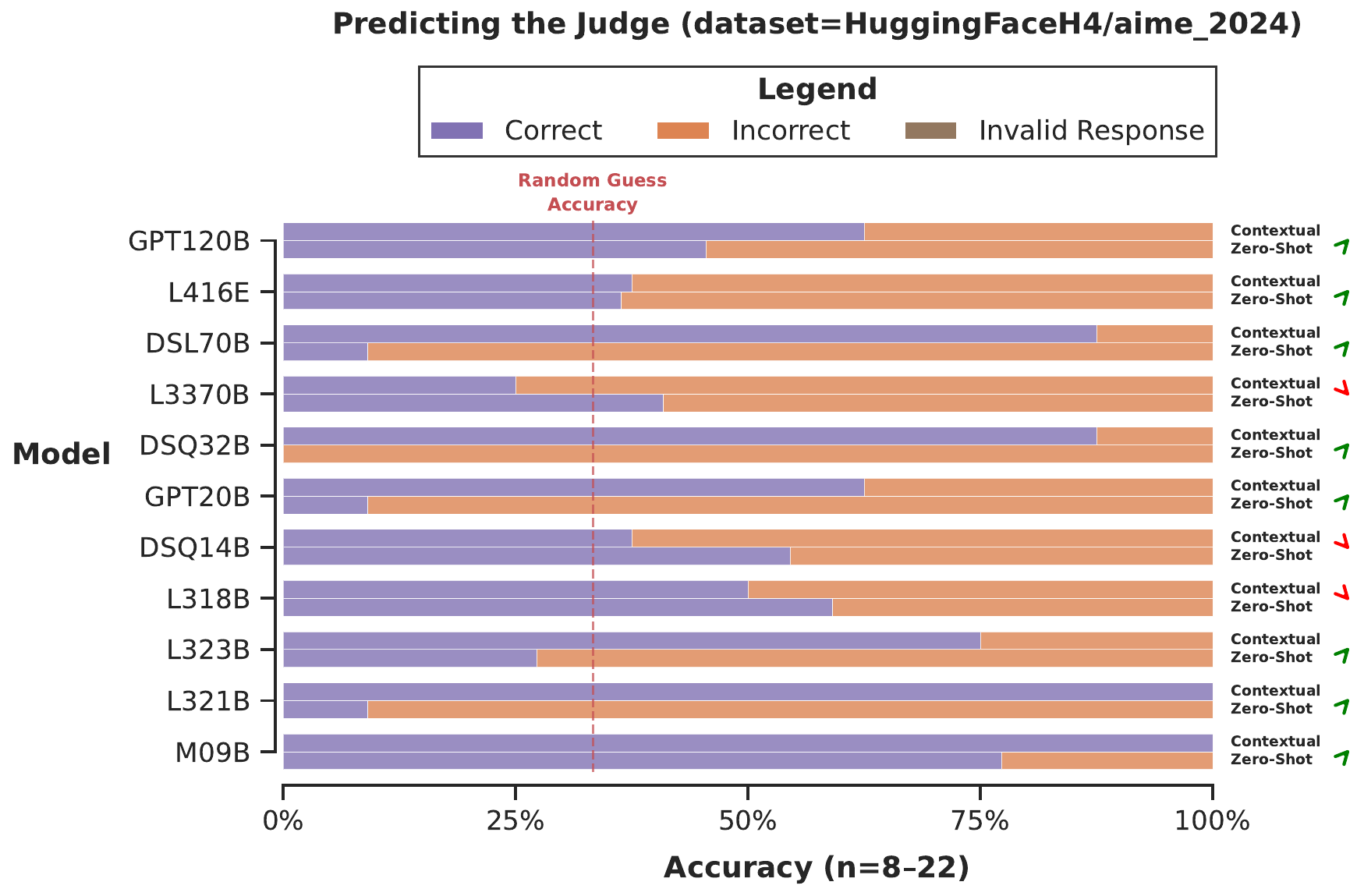}
    \caption{The zero-shot and contextual prediction accuracy of the models on the AIME 2024 dataset.}
    \label{fig:zero_shot_and_contextual_prediction_accuracy_aime_2024}
\end{figure}

\begin{figure}[ht]
    \centering
    \includegraphics[width=.9\textwidth]{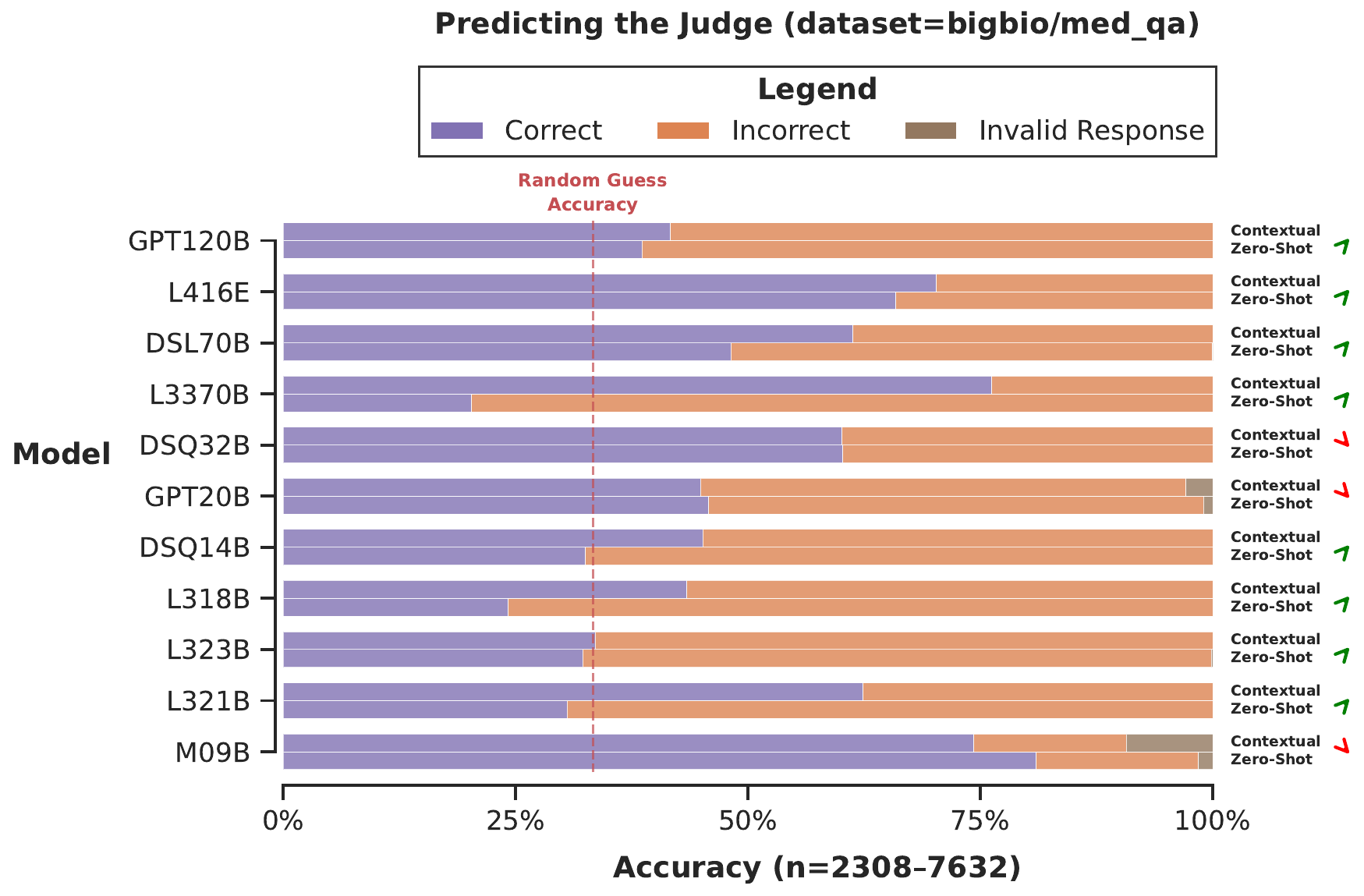}
    \caption{The zero-shot and contextual prediction accuracy of the models on the MedQA dataset.}
    \label{fig:zero_shot_and_contextual_prediction_accuracy_med_qa}
\end{figure}

\begin{figure}[ht]
    \centering
    \includegraphics[width=.9\textwidth]{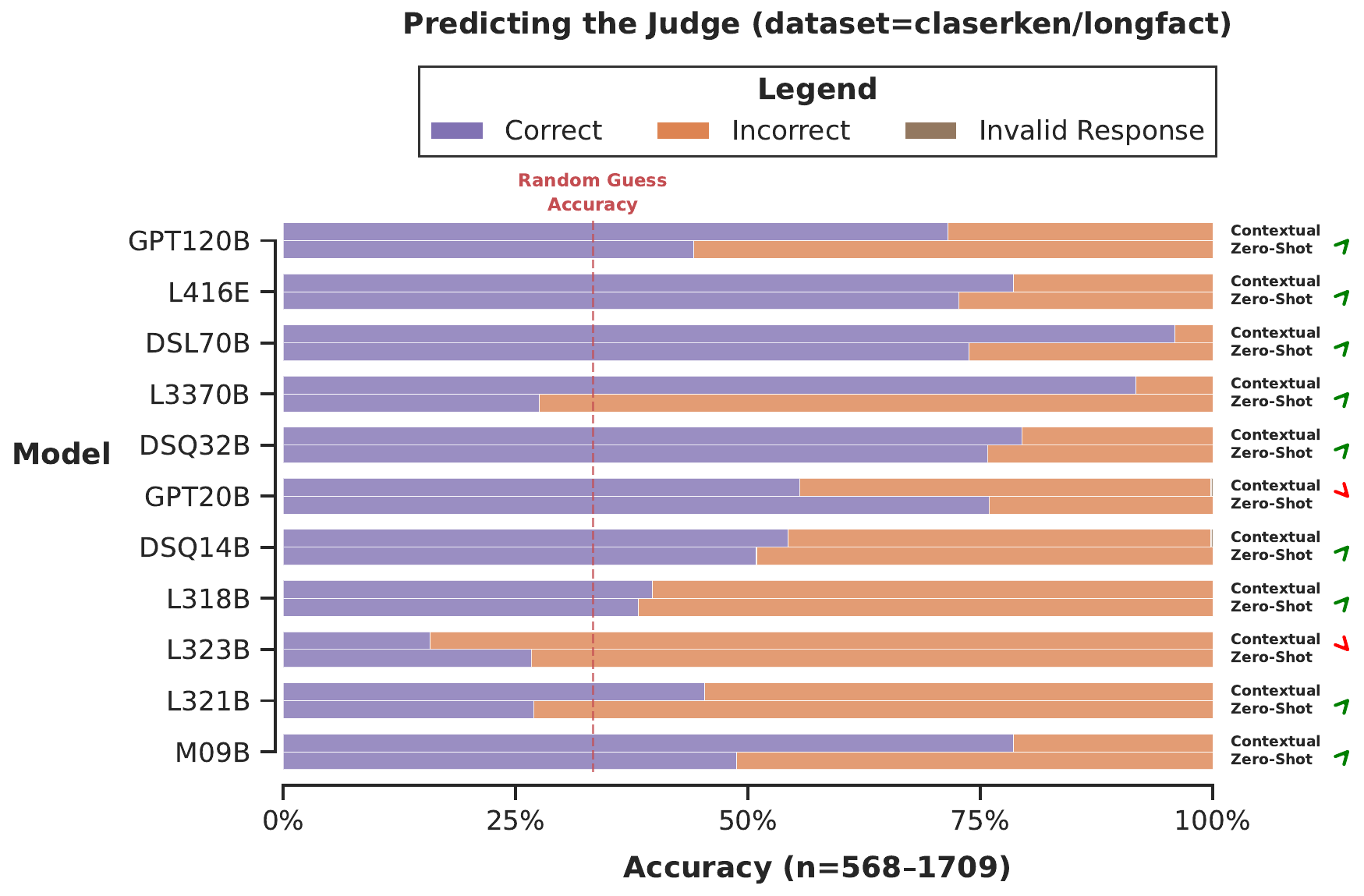}
    \caption{The zero-shot and contextual prediction accuracy of the models on the LongFact dataset.}
    \label{fig:zero_shot_and_contextual_prediction_accuracy_longfact}
\end{figure}

\begin{figure}[ht]
    \centering
    \includegraphics[width=.9\textwidth]{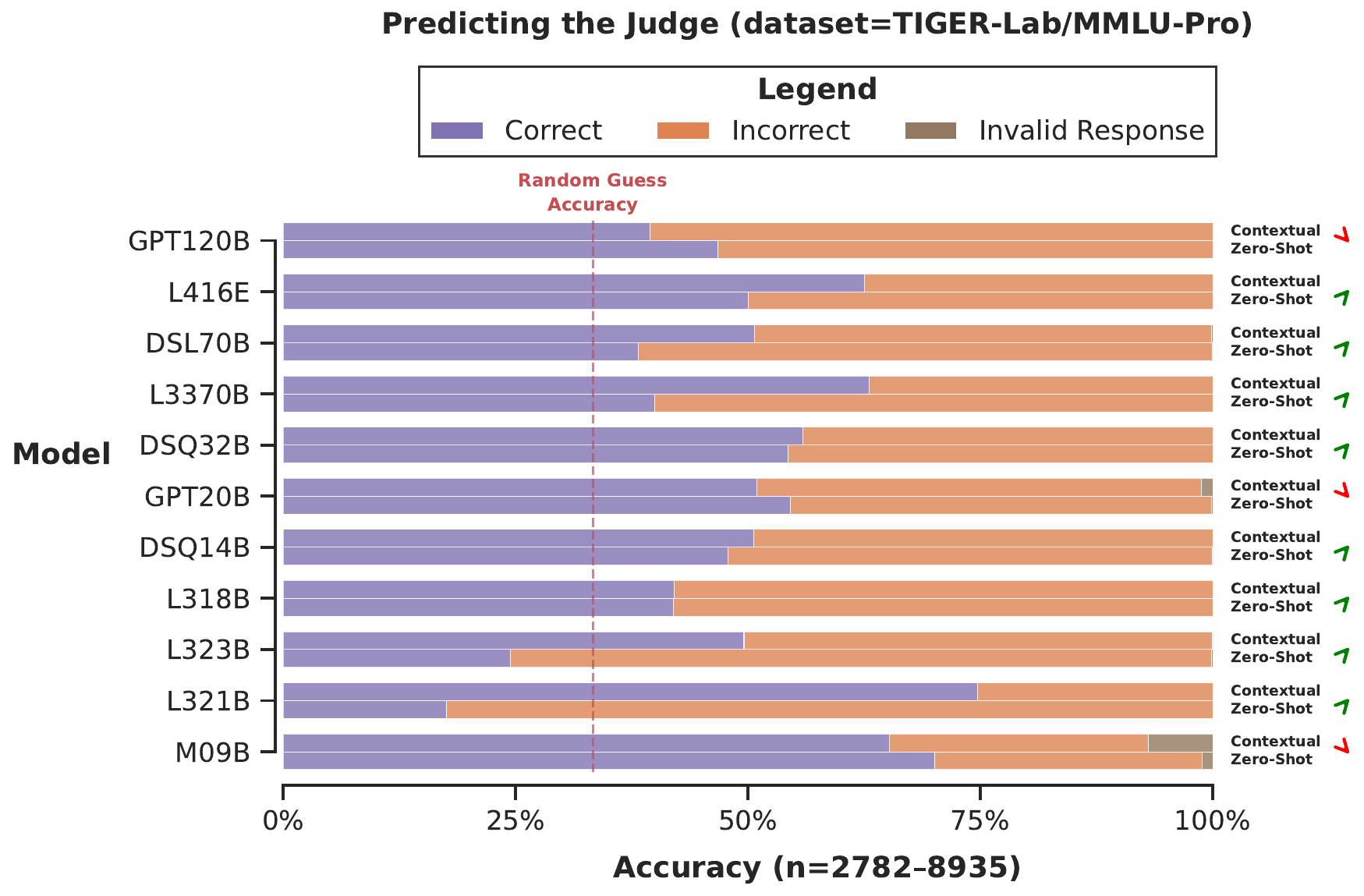}
    \caption{The zero-shot and contextual prediction accuracy of the models on the MMLU-Pro dataset.}
    \label{fig:zero_shot_and_contextual_prediction_accuracy_mmlu_pro}
\end{figure}

\begin{figure}[ht]
    \centering
    \includegraphics[width=.9\textwidth]{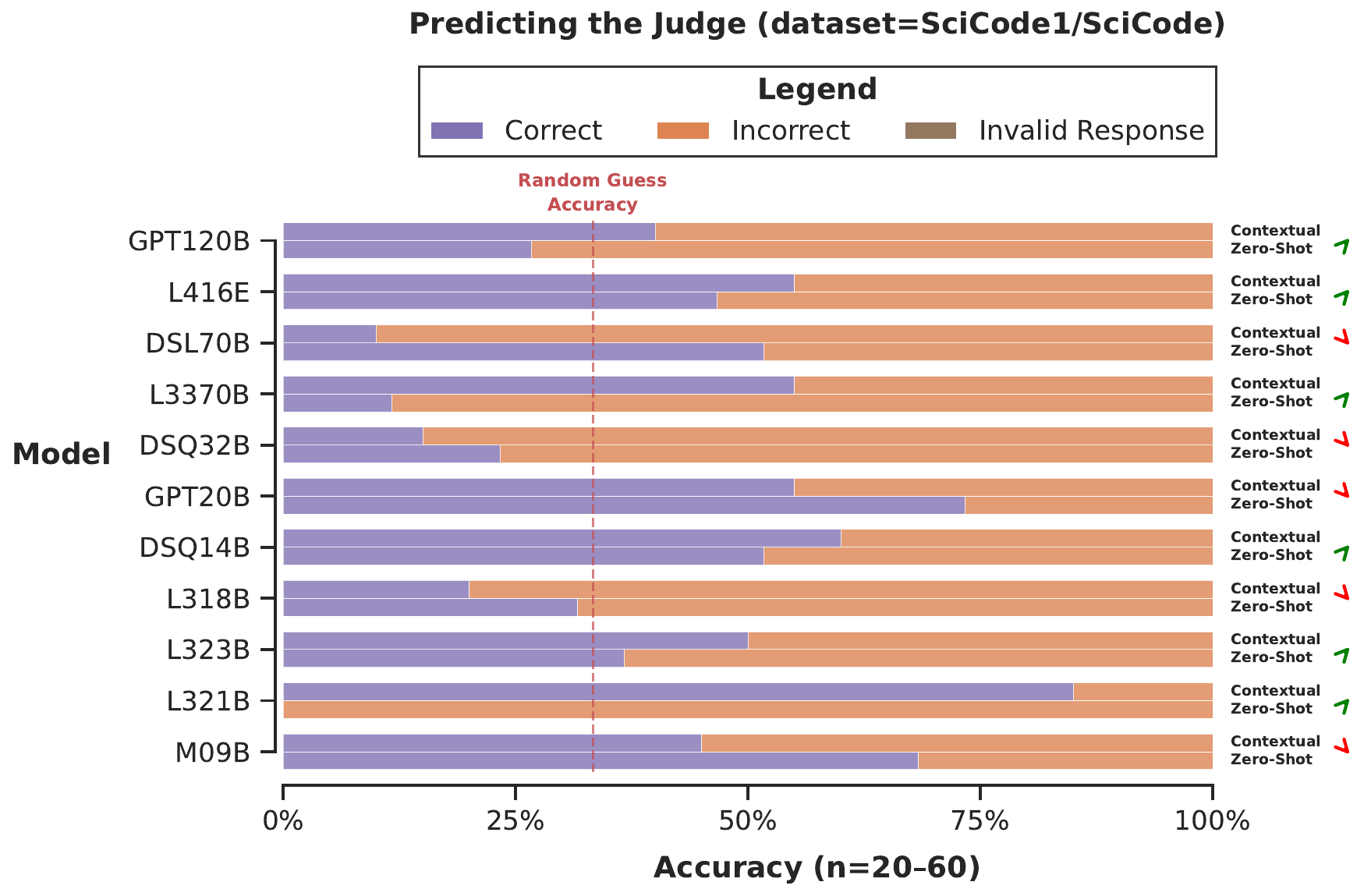}
    \caption{The zero-shot and contextual prediction accuracy of the models on the SciCode dataset.}
    \label{fig:zero_shot_and_contextual_prediction_accuracy_scicode}
\end{figure}

As the evaluation is contextual, the results shown in \cref{sec:zero_shot_and_in_context_prediction,sec:fine_tuning_approach} occur in the context of the fine-tuned models.
\cref{fig:zero_shot_and_contextual_prediction_accuracy_aime_2024,fig:zero_shot_and_contextual_prediction_accuracy_med_qa,fig:zero_shot_and_contextual_prediction_accuracy_longfact,fig:zero_shot_and_contextual_prediction_accuracy_mmlu_pro,fig:zero_shot_and_contextual_prediction_accuracy_scicode} show both the zero-shot and contextual prediction accuracy of the models when the fine-tuned models are excluded from the experiment.
Evidently, the effect of the fine-tuned models being added to the mix are relatively minimal here.

\clearpage

\section{Use of Large Language Models in Writing}\label{app:use_of_large_language_models}

LLMs were used throughout the writing of the paper as general-purpose assist tools.
In particular, they were used to draft, refine, and polish sections of the paper as well as to discover and compare some relevant works.
Only some Appendix~sections can be said to be exempt of the above, with the contribution of the LLMs in writing being less pronounced around precise factual areas of the text (i.e., where the exact choice of words was critical for correctness).

\end{document}